\documentclass{article}

\usepackage{microtype}
\usepackage{graphicx}
\usepackage{subfigure}
\usepackage{booktabs} 

\usepackage{hyperref}
\usepackage{rotating}

\usepackage{amsfonts}
\usepackage{listings}
\usepackage{cite}
\usepackage{multirow}
\usepackage{tabularx}
\usepackage{tcolorbox} 
\usepackage{array} 
\usepackage{pdflscape}
\usepackage{caption}

\usepackage[accepted]{icml2025}

\usepackage{amsmath}
\usepackage{amssymb}
\usepackage{mathtools}
\usepackage{amsthm}
\usepackage{xspace}

\usepackage[capitalize,noabbrev]{cleveref}

\theoremstyle{plain}

\theoremstyle{definition}

\theoremstyle{remark}

\usepackage[textsize=tiny]{todonotes}

\makeatletter
\let\scshape\relax 
\DeclareRobustCommand\scshape{%
  \not@math@alphabet\scshape\relax
  \ifnum\pdf@strcmp{\f@family}{\familydefault}=\z@
    \fontfamily{qbk}%
  \fi
  \fontshape\scdefault\selectfont}
\makeatother

\makeatletter
\patchcmd{\hyper@makecurrent}{%
    \ifx\Hy@param\Hy@chapterstring
        \let\Hy@param\Hy@chapapp
    \fi
}{%
    \iftoggle{inappendix}{
        \@checkappendixparam{chapter}%
        \@checkappendixparam{section}%
        \@checkappendixparam{subsection}%
        \@checkappendixparam{subsubsection}%
        \@checkappendixparam{paragraph}%
        \@checkappendixparam{subparagraph}%
    }{}%
}{}{\errmessage{failed to patch}}

\newcommand*{\@checkappendixparam}[1]{%
    \def\@checkappendixparamtmp{#1}%
    \ifx\Hy@param\@checkappendixparamtmp
        \let\Hy@param\Hy@appendixstring
    \fi
}
\makeatletter

\newtoggle{inappendix}
\togglefalse{inappendix}

\apptocmd{\appendix}{\toggletrue{inappendix}}{}{\errmessage{failed to patch}}

\newcommand{\lisat}{\textsc{LISAt}\xspace}
\newcommand{\lisatpre}{\textsc{LISAt\textsubscript{pre}}\xspace}

\icmltitlerunning{\lisat: Language-Instructed Segmentation Assistant for Satellite Imagery}

\begin{document}

\twocolumn[
\icmltitle{\lisat: Language-Instructed Segmentation Assistant for Satellite Imagery}



\icmlsetsymbol{equal}{*}

\begin{icmlauthorlist}
\icmlauthor{Jerome Quenum}{equal,ucb}
\icmlauthor{Wen-Han Hsieh}{equal,ucb}
\icmlauthor{Tsung-Han Wu}{ucb}
\icmlauthor{Ritwik Gupta}{ucb}
\icmlauthor{Trevor Darrell}{ucb}
\icmlauthor{David M. Chan}{ucb}
\end{icmlauthorlist}

\icmlaffiliation{ucb}{Department of Electrical Engineering and Computer Sciences, University of California-Berkeley, Berkeley, CA, USA}

\icmlcorrespondingauthor{Jerome Quenum}{jquenum@berkeley.edu}
\icmlcorrespondingauthor{Wen-Han Hsieh}{hense1219@berkeley.edu}

\icmlkeywords{Geospatial Artificial Intelligence, Multi-Modal Artificial Intelligence, Reasoning Segmentation with Satellite Images}

\vskip 0.3in
]



\printAffiliationsAndNotice{\icmlEqualContribution} 

\begin{abstract}

Segmentation models can recognize a pre-defined set of objects in images. However, segmentation models that can ``reason’’ over complex user queries that implicitly refer to multiple objects of interest are still in their infancy. Recent advances in ``reasoning segmentation’’---generating segmentation masks from complex, implicit query text---show that vision-language models can reason across an open domain of objects and produce reasonable segmentation outputs. However, our experiments show that such models struggle when operating on complicated remote-sensing images. In this work, we introduce \lisat, a vision-language model (VLM) designed to describe complex remote-sensing images, answer questions about those images, and also identify and segment objects within the scenes. We trained \lisat on a new curated geospatial reasoning-segmentation dataset, GRES, comprising 27,615 annotations across 9,205 images, and a multimodal geospatial pre-training dataset, PreGRES, containing $>$1M QA pairs. \lisat outperforms existing geospatial foundation models such as RS-GPT4V by over 10.04\% (BLEU-4) on remote-sensing visual description tasks and outperforms state-of-the-art open-domain models on remote-sensing reasoning segmentation tasks by 143.36\% (gIoU). Our model, datasets, and code are available on our \href{https://lisat-bair.github.io/LISAt/}{project page}. 



\end{abstract}

\section{Introduction}
\label{Introduction}

\begin{figure}
    \centering
    \includegraphics[width=\linewidth]{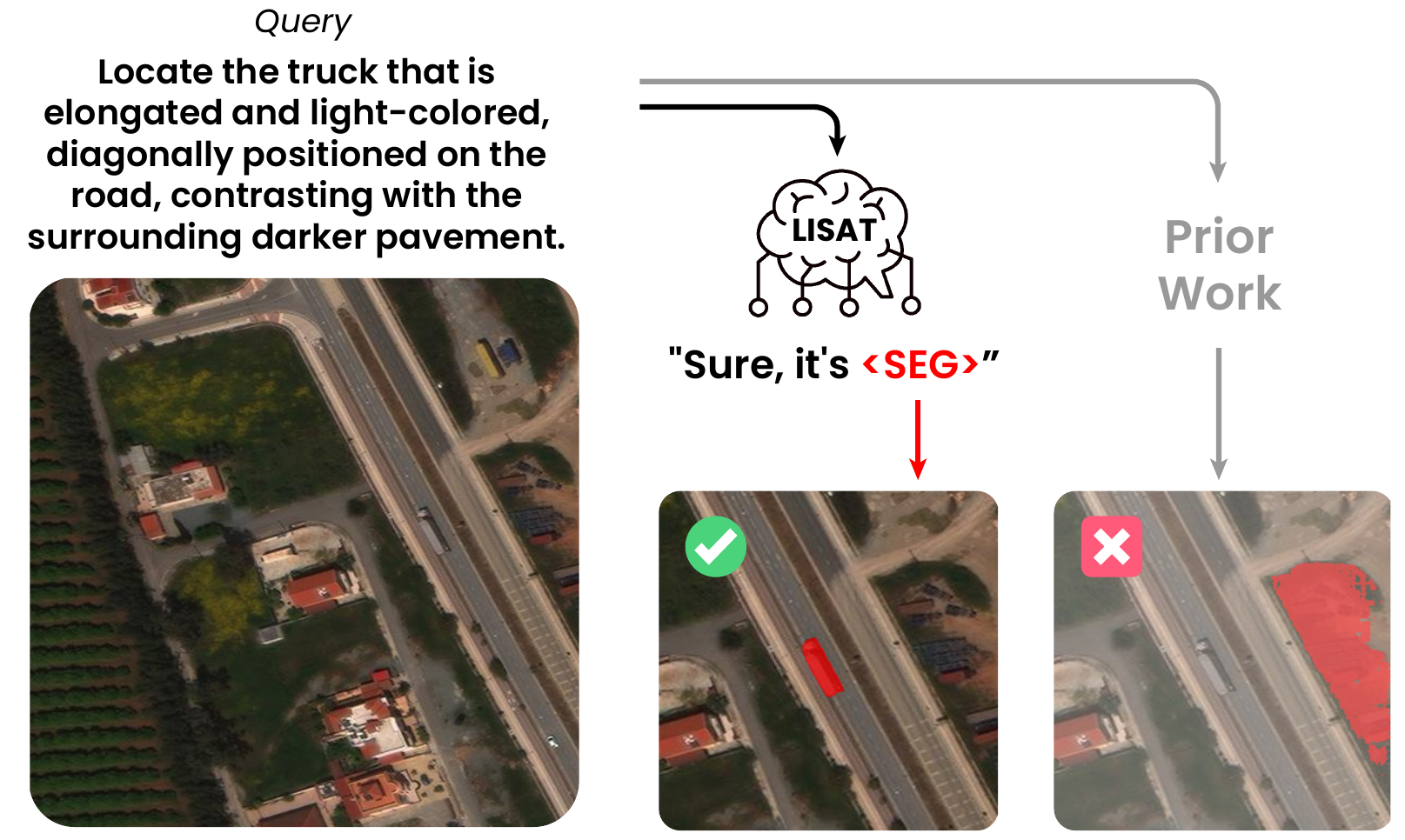}
    \caption{Existing models struggle to generate accurate segmentation masks for complex natural language queries in remote-sensing imagery. \lisat, our open-source, open-data, foundation model for geospatial reasoning segmentation trained on GRES, our new semi-synthetic dataset for remote-sensing reasoning segmentation, helps to bridge the gap between SOTA reasoning segmentation models and remote-sensing domains.}
    \label{fig:teaser}
\end{figure}

Segmentation models for remote-sensing have been a staple of geospatial analysis, supporting applications ranging from disaster response, environmental monitoring, and more \cite{weiss2020remote, subudhi2021survey}. These models typically operate within rigid boundaries but struggle to adapt to real-world scenarios in which the ability to segment regions based on flexible, user-defined queries---tasks often referred to as reasoning segmentation---is paramount \cite{lai2024lisa}. For instance, a query such as ``identify flood-prone urban areas'' or ``which regions have observed urban expansion'' demands that segmentation models move beyond static object recognition and into contextual, task-specific reasoning. Despite the clear need for such capabilities, reasoning segmentation remains largely underexplored in the remote-sensing domain, limiting the adaptability of segmentation systems in practice.


\textcolor{black}{Adapting existing vision models for the remote-sensing domain introduces a unique set of challenges differing fundamentally from those encountered in natural imagery (Rolf et al., 2024). Remote-sensing images can drastically vary in terms of clutter, objects of interest can be very small or very large, and implicit interactions between objects can span over long distances. Additionally, remote-sensing is characterized by both subtle visual differences between drastically different types of objects where it is challenging to distinguish between objects that look similar in satellite images but have vastly different semantic meanings (e.g., small cars vs. buildings), and extreme variations in scale along with object diversity where remote-sensing images contain objects of vastly different sizes (e.g., entire cities, forests vs. individual trees). Compounding these difficulties is the lack of high-quality annotated data consisting of natural language queries and remote-sensing imagery pairs. As a result, models trained on natural image datasets or designed for general-purpose reasoning struggle to achieve high performance when directly applied to geospatial tasks \cite{xu2024rs,zhang2024earthgpt}.}


Recent geospatial-specific foundation models, such as RS-GPT4V \cite{xu2024rs}, EarthGPT \cite{zhang2024earthgpt}, and others \cite{irvin2024teochat, kuckreja2024geochat} have demonstrated strong performance in related visual understanding and reasoning tasks such as visual captioning and visual question answering. Despite these advances, such models remain limited to textual outputs and cannot generate segmentation masks or localize objects within images. This lack of segmentation capability presents a significant barrier for applications that require spatially explicit reasoning. Some vision-language models can generate segmentations from text queries \cite{lai2024lisa,ren2024pixellm,hu2016segmentation}, but they struggle to adapt to the unique challenges of remote-sensing, \textcolor{black}{mentioned above} (see \autoref{lisa_models_performance}).

We address these challenges by introducing \lisat (\textbf{L}anguage \textbf{I}nstruction \textbf{S}egmentation \textbf{A}ssistant for Sa\textbf{t}ellite Images), an open-source and open-data vision-language model that bridges the gap between reasoning segmentation and remote-sensing foundation models. Central to \lisat’s development are two new datasets: the Geospatial Reasoning Segmentation dataset (GRES), comprising 27,615 pixel-level annotations paired with natural language reasoning-segmentation queries across 9,205 images, and PreGRES, a fine-tuning dataset aggregated from existing remote-sensing datasets consisting of over 1 million question-answer pairs. These datasets enable \lisat to handle scale, resolution, and complexity, in aligning textual queries with remote-sensing and top-down imagery. \lisat achieves significant performance gains over state-of-the-art geospatial and open-domain models. Specifically, \lisat outperforms existing geospatial foundation models, such as RS-GPT4V, by over 10.04\% on BLEU-4 on remote-sensing visual description tasks and outperforms state-of-the-art open-domain models on remote-sensing reasoning segmentation by 143.36\% on gIoU.

\section{Related Work}
\label{Related_Work}


Semantic segmentation has been an essential task in remote-sensing with applications ranging from urban planning \cite{gao2024enrich, hansen2024opentrench3d, chen2024edge}, economic assessment \cite{flotzinger2024dacl}, precision agriculture \cite{weiss2020remote}, resource management \cite{gao2024enrich, li2024cpseg} and environmental protection \cite{subudhi2021survey}. A key challenge facing such models, however, is that they are typically constrained to rigid, task-specific models that fail to generalize across applications without significant fine-tuning and adjustment despite using identical imagery and label ontologies. Recently, however, since the emergence of vision language models (VLMs) as a predominant paradigm \cite{radford2021learning, liu2024llava, liu2024visual}, there has been a renewed interest in multimodal foundation models that are capable of responding to/answering \textit{arbitrary} natural language (or multimodal) queries. Models such as GPT-4 \cite{achiam2023gpt} and LLaVA \cite{liu2024visual, liu2024llava} have expanded on this by allowing users to provide an image along with a natural language query, to solve tasks such as visual description and visual question answering.

\subsection{Geospatial Foundation Models}

Recent geospatial foundation models have adapted the foundation model paradigm to remote sensing, enabling multi-task capabilities for tasks like captioning, visual question answering, and object detection. EarthGPT \cite{zhang2024earthgpt} unifies a wide range of multi-sensor RS tasks, including scene classification, image captioning, and object detection, using a large-scale multimodal dataset derived from several task-specific datasets (see \autoref{sec:pregres}). TEOChat \cite{irvin2024teochat} introduces temporal reasoning for applications such as change detection and damage assessment, demonstrating strong performance on temporal sequence tasks, but struggles on more general descriptions. GeoChat \cite{kuckreja2024geochat} supports region-specific dialogue and visual grounding, enabling fine-grained interaction with high-resolution RS imagery, while SkyEyeGPT \cite{zhan2024skyeyegpt} achieves notable performance on image-level and region-level vision-language tasks with a streamlined instruction-following architecture. RS-GPT4V \cite{xu2024rs} emphasizes fine-grained object understanding and complex scene reasoning, leveraging a hierarchical instruction-following approach. While these models represent significant progress across a wide range of tasks, such models have been limited by their ability to produce only \textit{natural language outputs}. Our proposed work, \lisat addresses this limitation by natively producing segmentation masks in addition to answering natural language queries. 

\subsection{Reasoning Segmentation}

Beyond just producing segmentation masks for single classes, a goal of \lisat is to perform ``reasoning segmentation,'' the task of generating a segmentation mask from a complex or implicit query text (\autoref{fig:teaser}). Two overarching approaches have been developed for this task. LISA \cite{lai2024lisa} introduced this concept with an embedding-as-mask approach, allowing segmentation via a \texttt{[SEG]} token which is decoded into a segmentation mask using a SAM decoder \cite{kirillov2023segment}. PixelLM \cite{ren2024pixellm} expanded on this method by leveraging lightweight pixel decoder and segmentation codebook to improve multi-target differentiation in the same paradigm, while GLaMM \cite{rasheed2024glamm} also targeted the granularity problem through additional focused data. GSVA \cite{xia2024gsva} extended the \texttt{[SEG]} paradigm by introducing a \texttt{[REJ]} token to handle ambiguous or absent targets in queries. In the second paradigm, models such as Shikra \cite{chen2023shikra}, Kosmos-2 \cite{peng2024grounding} and others \cite{zhang2024groundhog, xuan2024pink, liu2024llava, liu2024visual} focus on solving reasoning segmentation tasks with natural language alignment: represent visual concepts as sequences of natural language tokens (such as the literal coordinates of a bounding box). Despite these advances, existing models often fall short when applied to remote sensing due to challenges like varying spatial resolutions, fine differences between target classes, and the lack of domain-tailored datasets (See \autoref{lisa_models_performance}). Our proposed work, \lisat, extends the embedding-as-mask approach to top-down remote-sensing data.

\section{Geospatial Reasoning Segmentation Dataset}
\label{sec:gres}


\textcolor{black}{The development of vision-language models (VLMs) for remote sensing has been hindered by the lack of high-quality remote sensing imagery paired with natural language data, a key challenge outlined in our introduction. Unlike natural image datasets, remote-sensing data require fine-grained, context-aware segmentation that accounts for extreme variations in scale, subtle object differences, and the ability to reason across complex spatial relationships. To help alleviate this need, we introduce the Geospatial Reasoning Segmentation Dataset (GRES), a collection of vision and language data designed around remote-sensing applications. \textbf{GRES} consists of two core components: \textbf{PreGRES}, a dataset consisting of over 1M remote-sensing specific visual instruction-tuning Q/A pairs for pre-training geospatial models, and \textbf{GRES}, a semisynthetic dataset specialized for reasoning segmentation of remote-sensing data.
With this structure of \textbf{GRES}, we enable LISAT to overcome both data scarcity and the domain transfer limitations faced by general-purpose models. The dataset is specifically designed to handle scale variability, object diversity, and complex reasoning queries, making it a critical resource for advancing geospatial VLMs.}

\subsection{PreGRES}
\label{sec:pregres}

PreGRES is a large-scale structured collection of existing smaller-scale geospatial datasets designed for fine-tuning vision-language models in remote sensing applications. It integrates multiple sources, each contributing to different aspects of geospatial data understanding. The datasets within GRES provide coverage across image captioning, visual question answering, and visual grounding tasks: \begin{enumerate}
    \item \textbf{Image Captioning:} \texttt{NWPU-Captions}~\cite{cheng2022nwpu}, \texttt{RSICD}~\cite{lu2017exploring}, \texttt{RSITMD}~\cite{yuan2022exploring}, \texttt{Sydney-Captions}~\cite{qu2016deep}, and \texttt{UCM-Captions}~\cite{qu2016deep}. Each contributes paired image-text data, and contains long-form descriptions of top-down imagery across different geospatial environments, increasing the diversity of language supervision.  
    \item \textbf{Visual Question Answering (VQA):}  \texttt{RSVQA\_LR}~\cite{lobry2020rsvqa}, \texttt{RSVQA\_HR}~\cite{lobry2020rsvqa}, \texttt{FloodNet}~\cite{rahnemoonfar2021floodnet}, and \texttt{RSIVQA}~\cite{zheng2021mutual}. Each of these datasets consists of structured question-answer pairs and supports reasoning over aerial and satellite images, covering tasks such as object identification, scene understanding, and disaster assessment. 
    \item \textbf{Visual Grounding / Region-Level Captioning:} \texttt{DIOR-RSVG}~\cite{zhan2023rsvg} provides paired text-image data for object localization and spatial reference resolution, and \texttt{NWPU-RESISC45}~\cite{cheng2017remote} supplies scene classification labels.  
\end{enumerate}

Overall, PreGRES consists of 119,279 images and 1,204,993 question-answer pairs and is used in the first-stage pre-training of the \lisat model enabling general-purpose geospatial question-answering in the final \lisat model. For more details on dataset composition, see \autoref{task_data_sources_overview}.

\subsection{GRES}
\label{sec:gres_gres}

\begin{figure*}
    \centering
    \includegraphics[width=\linewidth]{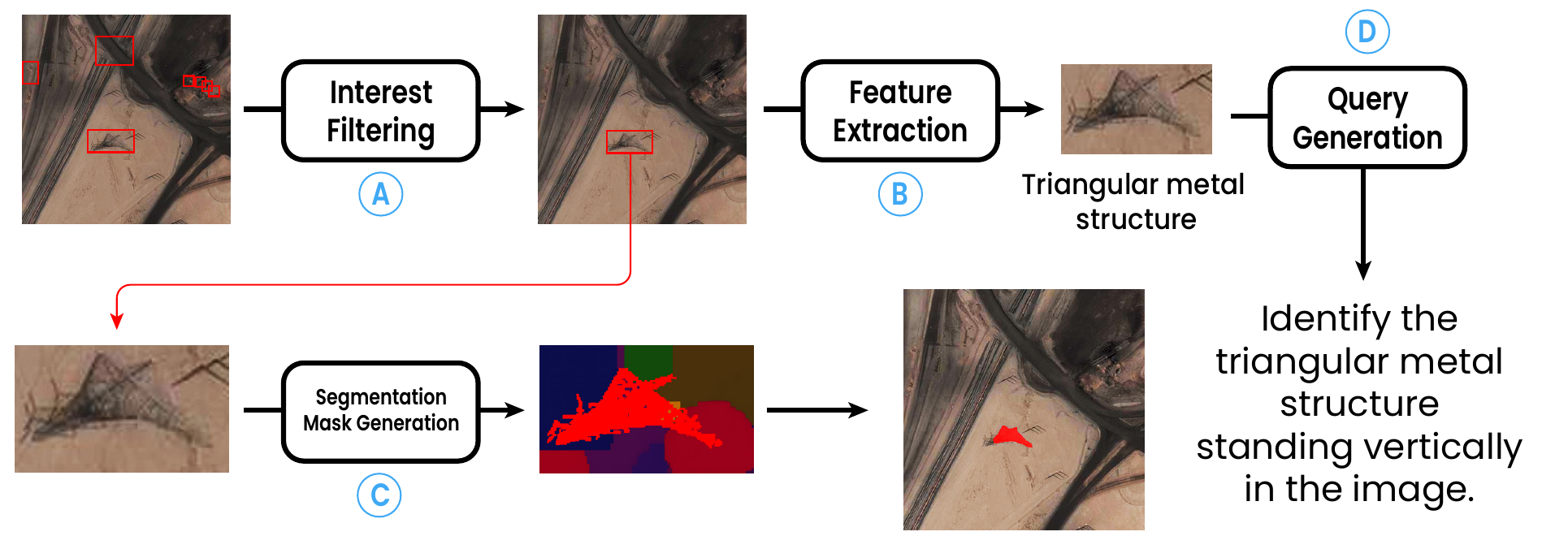}
    \caption{To generate synthetic data, we start with a seed detection dataset (\texttt{xView}). We then filter detections for those that are both visually interesting and highly distinguishable (A). For those detection, we then generate a natural language description (B), and a pixel-wise segmentation mask (C). Finally, the natural language description is used to generate a localization query (D). Together, the segmentation mask and the query form a ground-truth pair for the \lisat reasoning segmentation fine-tuning. }
    \label{fig:gres}
\end{figure*}

GRES is a semi-synthetic dataset designed explicitly for geospatial reasoning segmentation. Each sample in GRES consists of an image, a natural language query referring to a single object in that image, and a pixel-level segmentation mask (See \autoref{fig:gres} for an example of a GRES query/image pair). This task allows us to train the \lisat model to correctly localize images at a pixel level within the scene, even in the case of multiple objects requiring disambiguation. 

To build the dataset, we begin with a subset of the \texttt{xView} dataset \cite{lam2018xview} consisting of $26,541$ high-resolution satellite images spanning approximately $1,400$ square kilometers, covering more than $60$ classes. \texttt{xView} consists of paired images and object detections within the images in bounding box form. To convert \texttt{xView} images/annotations to GRES annotations/images, we follow the process overviewed in \autoref{fig:gres}. 

\textcolor{black}{Given an input image of size 512 × 512, we divide it into 4 quadrants, where the top-left quadrant is defined by \( 0 \leq x \leq 255, 0 \leq y \leq 255 \); the top-right quadrant is defined by \( 256 \leq x \leq 511, 0 \leq y \leq 255 \); the bottom-left quadrant is defined by \( 0 \leq x \leq 255, 256 \leq y \leq 511 \); the bottom-right quadrant is defined by \( 256 \leq x \leq 511, 256 \leq y \leq 511 \).}

In the first part of the pipeline, we need to generate a ``disambiguating query'' that selects for a single object within the scene from the large set of objects. To do so, we first filter the scenes for two key objectives: (1) uniqueness (i.e. can objects be easily disambiguated with a natural language query), and (2) interest (i.e. are the objects visually interesting) (\autoref{fig:gres}, A). An object is considered ``unique'' in an image if it is one of less than $2$ detections of its class in its respective quadrant, and an object is considered ``visually interesting'' if it belongs to a class appearing in less than 50\% of the overall subset of \texttt{xView} detections. Comprehensive statistics of object categories after filtering are available in \autoref{object_category_summary}. \textcolor{black}{To ensure a balanced evaluation, our dataset includes queries with and without explicit spatial references, each with a 50\% probability}.

After the filtering stage, we convert the object detection to a query using a set of structured queries to a large vision and language model trained on natural images (in our case, GPT-4v \cite{achiam2023gpt}, \autoref{fig:gres}, B). In the first prompting stage, we ask the VLM to identify unique characteristics of the class within the bounding box by asking the model to \texttt{``Find visual features (color, shape, size, etc.) that to help find or segment \{class\_name\} in the image.''}. We then ask the VLM to come up with a sentence describing the object in the bounding box within the scene using the collected unique characteristics (See the full prompt in \autoref{Prompt_Engineering_1}). Given these features, we prompt the VLM again with the full image, along with other detections in the image and the position of the bounding box to produce a query(see the full prompt in \autoref{Prompt_Engineering_2}, \autoref{fig:gres}, D). 


In the second part of the pipeline (\autoref{fig:gres}, C), we need to generate the pixel-based mask from the bounding box. To do this, we leverage a GeoSAM model \cite{sultana2023geosam} with a custom high-resolution inference configuration (128 points per side, 0.95 prediction IoU threshold, and 0.95 stability score with an 80-pixel minimum mask region area) to produce a part-wise segmentation of each bounding box. We then add any sub-parts that cover more than $80px$ of the underlying bounding box to the final pixel mask.

We then asked the VLM to rephrase each query two separate ways which added to the initially generated query gives us 3 queries per image. This pipeline overall results in a dataset consisting of 9,205 images and 27,615 natural language queries/answers within those images. From this dataset, we generate train, test, and validation splits consisting of 7,205, 1,500, and 500 images respectively.


\section{Training VLMs for Geospatial Reasoning Segmentation}

\begin{figure}[t]
    \centering
    \includegraphics[width=0.9\linewidth]{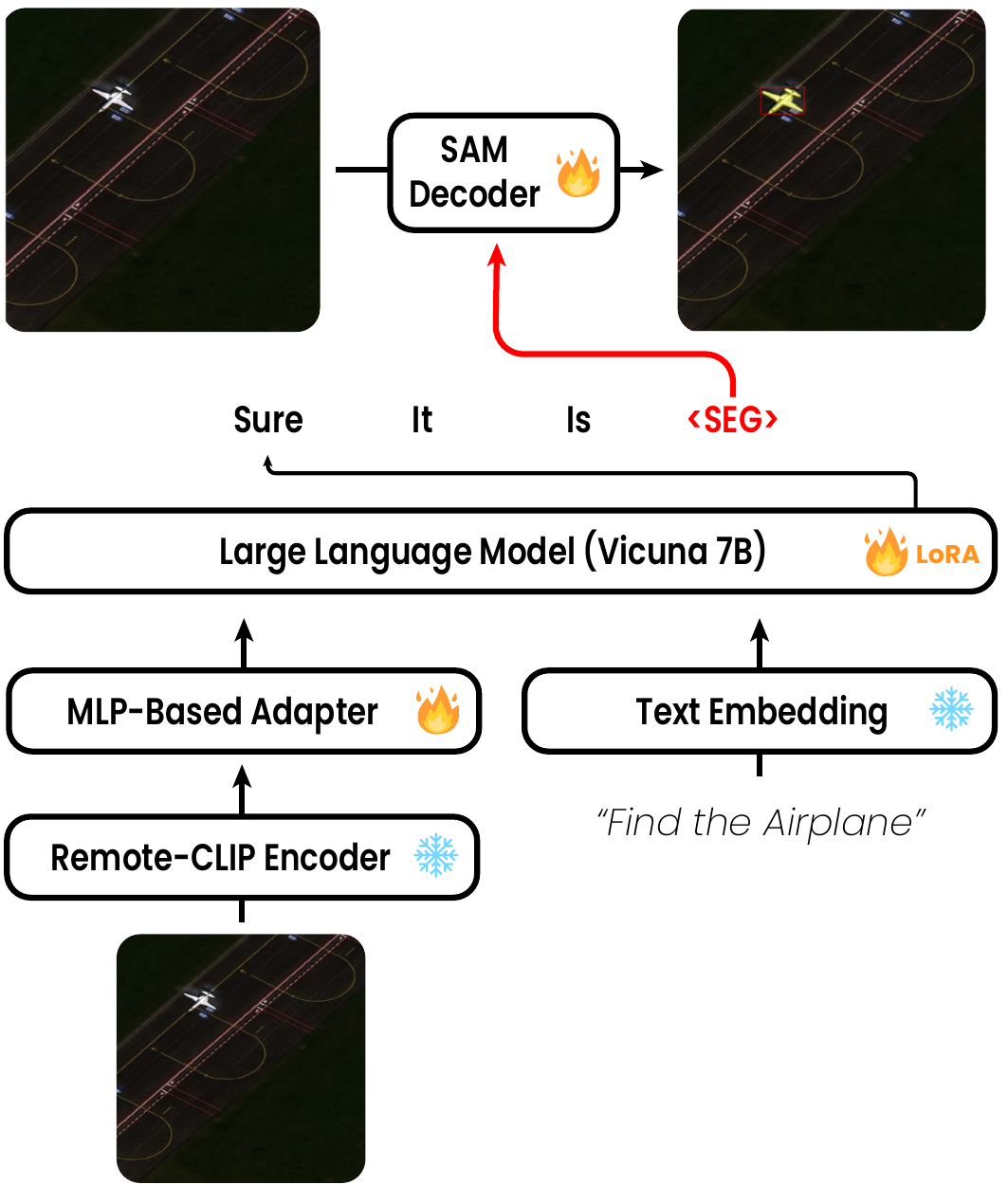}
    \caption{\lisat integrates a geospatial multimodal large language model (MLLM) with a segmentation decoder to enable reasoning-based segmentation. \lisat first pre-trains a Remote-CLIP-based MLLM on PreGRES before fine-tuning on GRES. We then expand the LMM vocabulary with a segmentation token (\texttt{<SEG>}), whose final-layer embedding is projected into the SAM segmentation query space and combined with image features to produce a segmentation mask.}
    \label{fig:method}
\end{figure}

Inspired by LISA \cite{lai2024lisa}, \lisat integrates a multimodal large language model (LLM) with a segmentation model. The multimodal LLM processes both textual and visual inputs, leveraging datasets that contain image-text pairs for instruction-following and reasoning \cite{liu2024visual} while the segmentation model uses a dataset designed for high-quality mask generation \cite{kirillov2023segment}. An overview of the architecture is given in \autoref{fig:method}.

\subsection{Geospatial Multimodal Language Models}

While LISA \cite{lai2024lisa} leverages a pre-trained LLaVA \cite{liu2024llava, liu2024visual} model as a vision and language backbone, we found that leveraging LLaVA alone was insufficient to capture the range of queries and visual variance in remote-sensing applications. To solve this problem, in the first stage of our training process we trained a remote-sensing specific multimodal large language model to serve as the base MLLM for the segmentation backbone. Our architecture generally follows LLaVA \cite{liu2024llava, liu2024visual} with several modifications for remote-sensing applications.

For the base language model, we leverage \textcolor{black}{default} Vicuna\textcolor{black}{-7B} \cite{vicuna2023} to embed a text query $\mathbf{X}_l$. For the visual backbone, \lisat adopts the Remote-CLIP ViT-L/14 encoder \cite{liu2024remoteclip} to extract visual features from an input image $\mathbf{X}_v$.  To align visual representations with the language model's word embedding space, we use a simple linear projection matrix to produce a sequence of visual tokens that match the dimensionality of the word embeddings in the language model. 
A pre-trained Vicuna base model combined with the vision encoder is further pre-trained on PreGRES (see \autoref{Experiments}) with LoRA \cite{hu2021lora} prior to being trained on GRES. We refer to this pre-trained variant as \lisatpre. 

\subsection{\textcolor{black}{Preliminaries}}

Existing multimodal LLMs for remote sensing, such as RS-GPT4V \cite{xu2024rs} and EarthGPT \cite{zhang2024earthgpt}, support images and text as input but output only text. To produce segmentation masks, \lisat leverages the ``embedding-as-a-mask'' paradigm introduced by LISA \cite{lai2024lisa}, and expands the LLM vocabulary with a new token, \texttt{<SEG>}, which represents segmentation requests. When the model produces an output containing the \texttt{<SEG>} token, we extract the final layer embedding of that token, and project it via an MLP layer to the query space of a SAM-based segmentation decoder \cite{kirillov2023segment}. The segmentation decoder combines the query-projected final embedding and a set of visual features extracted from the base image to produce a final segmentation mask $\hat{\mathbf{M}}$. 

\subsection{Training Objectives}

\lisat is trained end-to-end with a loss function that combines text generation and segmentation objectives. The total loss $\mathcal{L}$ is the weighted sum of two components:

\begin{equation}
    \mathcal{L} = \lambda_{txt} \mathcal{L}_{txt} + \lambda_{mask} \mathcal{L}_{mask}.
\end{equation}

where the text generation loss $\mathcal{L}_{txt}$ is an autoregressive cross-entropy loss:

\begin{equation}
    \mathcal{L}_{txt} = \mathbf{CE}(\hat{\mathbf{y}}_{txt}, \mathbf{y}_{txt}).
\end{equation}

and the segmentation loss $\mathcal{L}_{mask}$ consists of a per-pixel binary cross-entropy (BCE) loss and a DICE loss, weighted by $\lambda_{bce}$ and $\lambda_{dice}$:

\begin{equation}
    \mathcal{L}_{mask} = \lambda_{bce} \mathbf{BCE}(\hat{\mathbf{M}}, \mathbf{M}) + \lambda_{dice} \mathbf{DICE}(\hat{\mathbf{M}}, \mathbf{M}).
\end{equation}







\section{Experimental Results}
\label{Experiments}
\label{Experimental_Setting}
\label{Evaluation_Metrics}

\begin{table*}[t]
\centering
\caption{Qualitative examples of the segmentations generated by \lisat on the GRES dataset.  \vspace{0.3em}}
\resizebox{\textwidth}{!}{
\begin{tabular}{>{\centering\arraybackslash}p{0.18\textwidth}m{0.22\textwidth}m{0.22\textwidth}m{0.22\textwidth}m{0.22\textwidth}}
\hline
\multicolumn{1}{c}{\textbf{Queries}} & 
\multicolumn{1}{c}{\textbf{RGB}} & 
\multicolumn{1}{c}{\textbf{LISA}} & 
\multicolumn{1}{c}{\textbf{\lisat (Ours)}} & 
\multicolumn{1}{c}{\textbf{Ground Truth}} \\
\toprule

\centering\arraybackslash\parbox{0.18\textwidth}{\centering\vspace{-9pt}Locate the building with a large rectangular structure, dark roof, and symmetrical window patterns.
} & 
\includegraphics[width=1.05\linewidth]{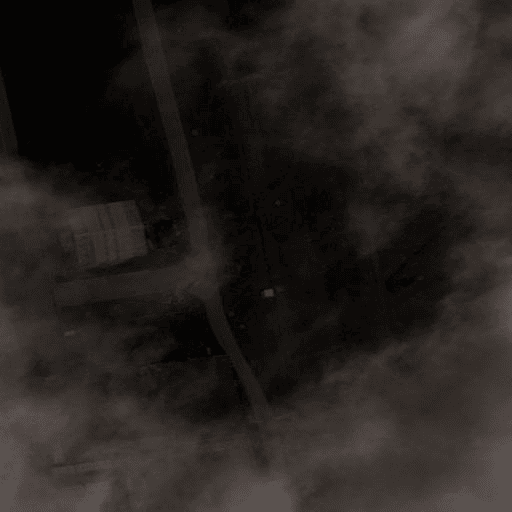} & 
\includegraphics[width=1.05\linewidth]{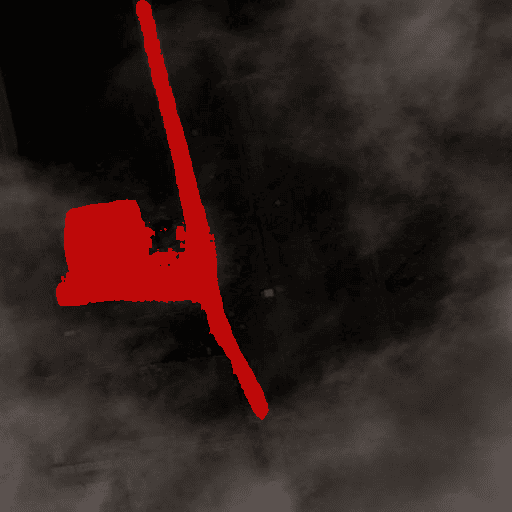} & 
\includegraphics[width=1.05\linewidth]{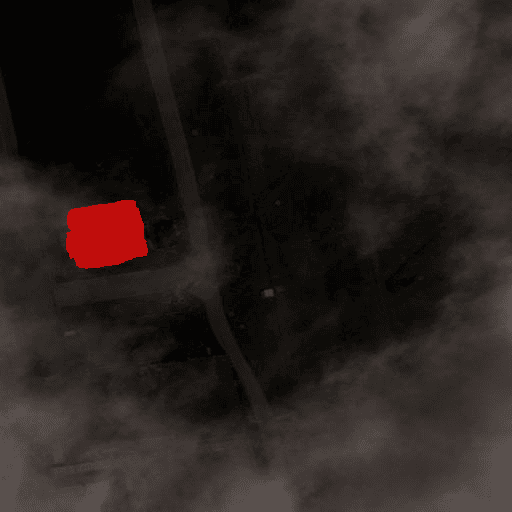} & 
\includegraphics[width=1.05\linewidth]{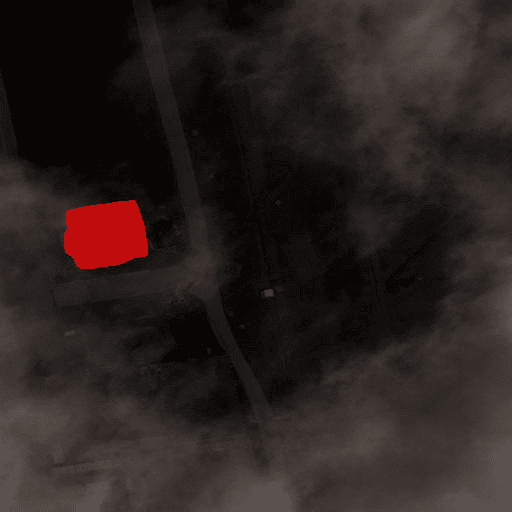} \\

\centering\arraybackslash\parbox{0.18\textwidth}{\centering\vspace{-9pt}Identify the facility in the center-left of the image.
} & 
\includegraphics[width=1.05\linewidth]{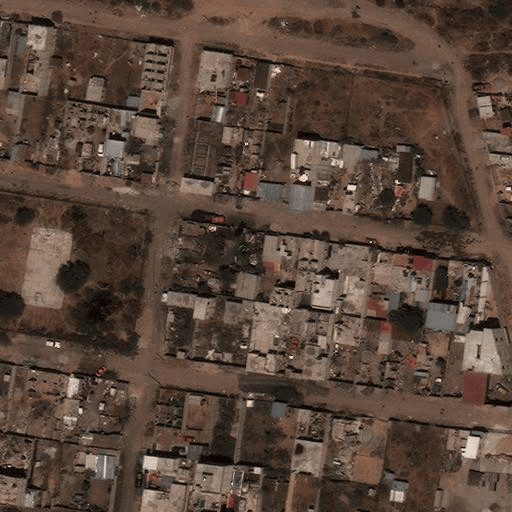} & 
\includegraphics[width=1.05\linewidth]{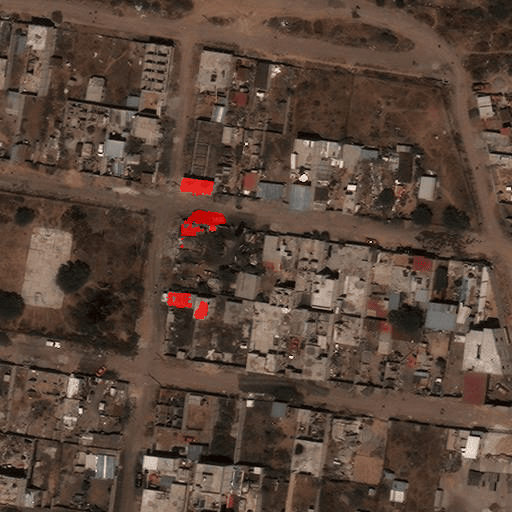} & 
\includegraphics[width=1.05\linewidth]{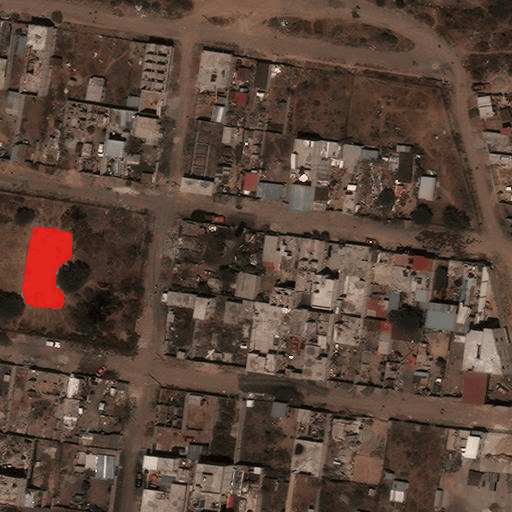} & 
\includegraphics[width=1.05\linewidth]{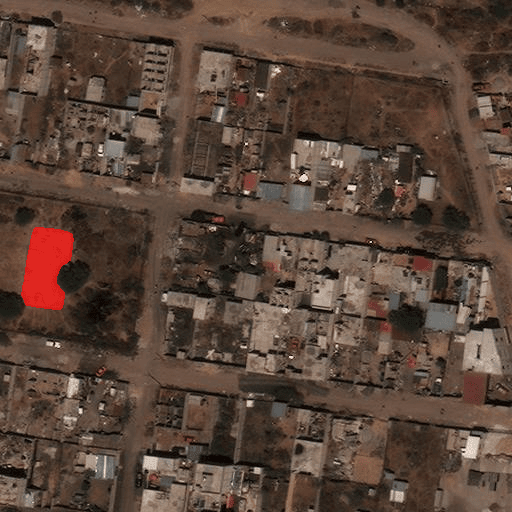} \\
\

\centering\arraybackslash\parbox{0.18\textwidth}{\centering\vspace{-9pt}Identify the damaged building in the center of the image.
} & 
\includegraphics[width=1.05\linewidth]{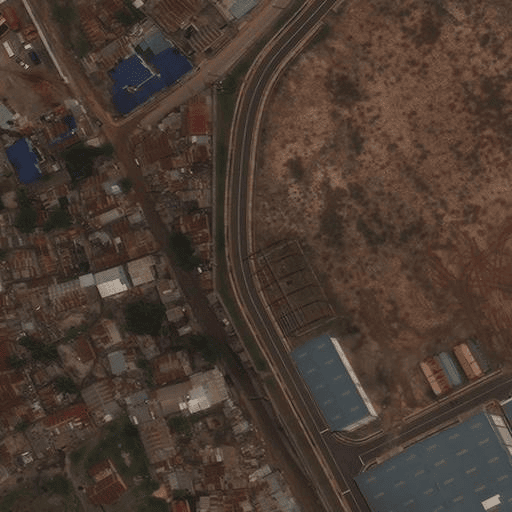} & 
\includegraphics[width=1.05\linewidth]{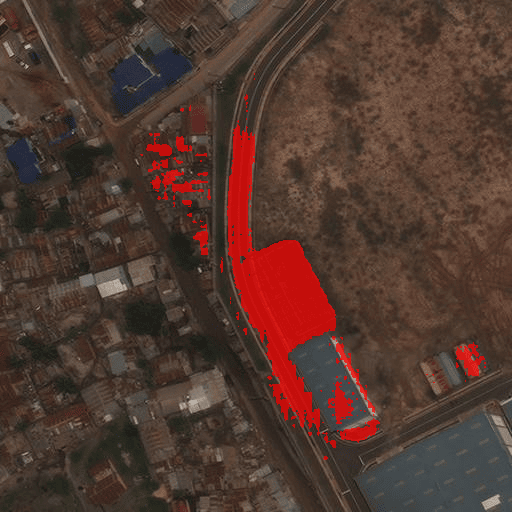} & 
\includegraphics[width=1.05\linewidth]{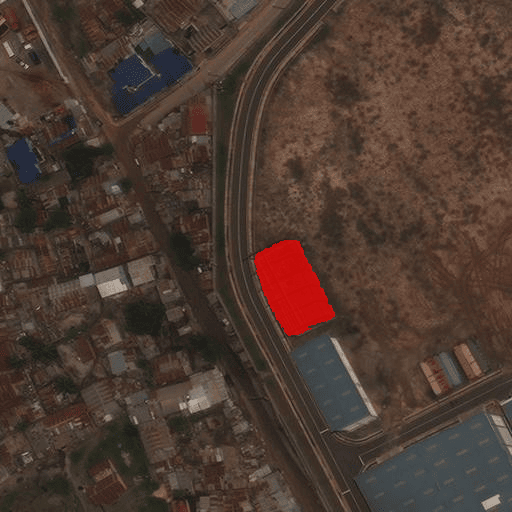} & 
\includegraphics[width=1.05\linewidth]{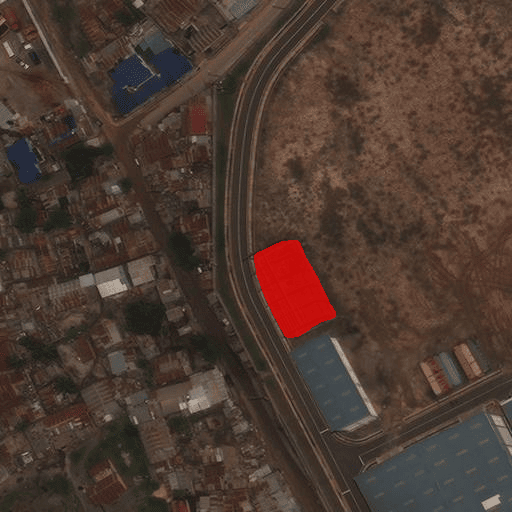} \\


\midrule
\centering\arraybackslash\parbox{0.18\textwidth}
{\centering\vspace{-9pt}\textcolor{red}{\textbf{Failure Case:} } Locate the dark, elongated rectangular shape with a red outline against the dark background to identify the barge.
} &
\includegraphics[width=1.05\linewidth]{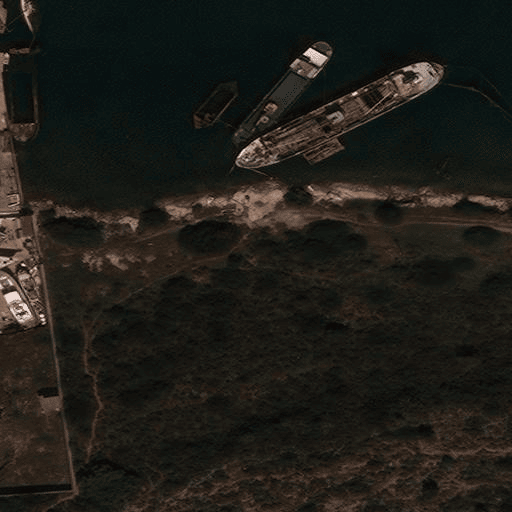} & 
\includegraphics[width=1.05\linewidth]{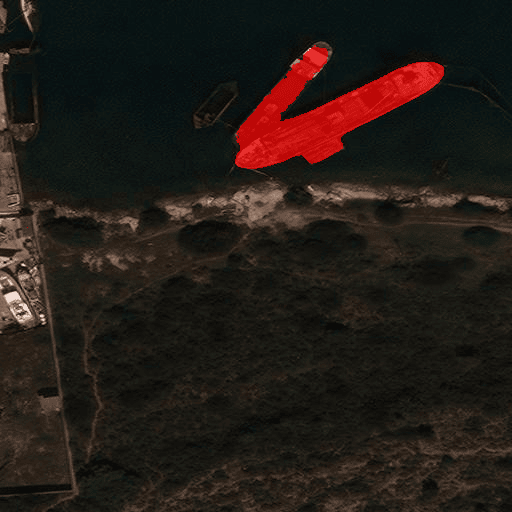} & 
\includegraphics[width=1.05\linewidth]{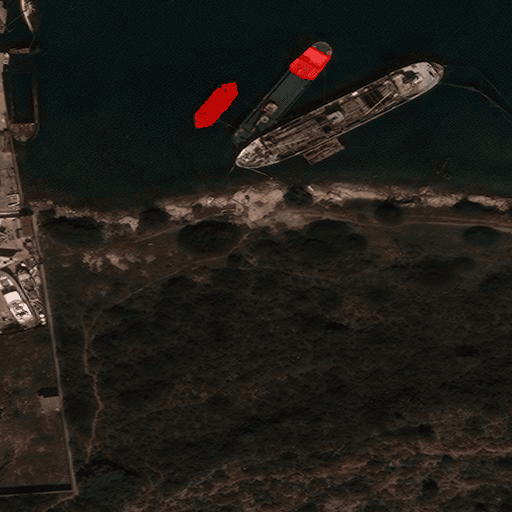} & 
\includegraphics[width=1.05\linewidth]{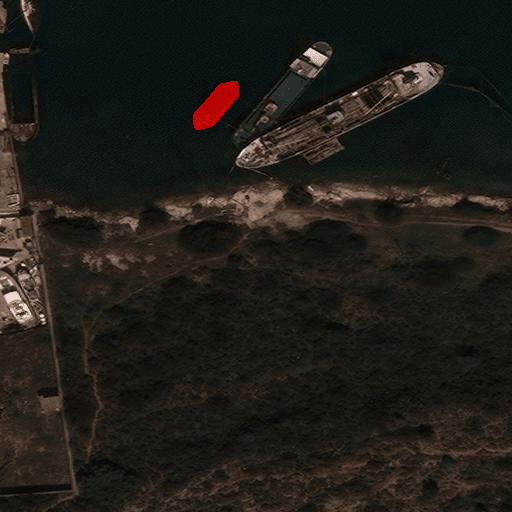} \\

\bottomrule
\end{tabular}
}
\label{tab:lisat_qual}
\end{table*}

\paragraph{Implementation Details:} \lisat and \lisatpre are trained on eight DGX A100 80GB GPUs. In the first stage, we pretrain \lisatpre (context length = 2048) using LoRA \cite{hu2021lora} for 1 epoch on PreGRES (described in \autoref{sec:pregres}) with next-token prediction cross-entropy loss. We employ the AdamW optimizer \cite{loshchilov2017decoupled} with a learning rate of $3e^{-4}$ and a cosine-decay learning rate scheduler, setting the batch size to 2 and gradient accumulation steps to 6.

In the second stage, we train \lisat using GRES, as well as two traditional natural image referring segmentation datasets, FP-Ref-COCO \cite{wu2024see} and ReasonSeg \cite{lai2024lisa}. LoRA is applied to \lisatpre, while the SAM decoder undergoes full fine-tuning. The learning rate is set to $3e^{-4}$, with all other configurations remaining the same. For the loss function, \textcolor{black}{we empirically found that setting} the weight for text generation loss ($\lambda_{txt}$) and mask loss ($\lambda_{mask}$) to 1.0, while the binary cross-entropy loss (BCE) ($\lambda_{bce}$) and Dice loss ($\lambda_{dice}$) are assigned weights of 2.0 and 0.5, respectively performs better.  The total training time was approximately 12 hours on eight DGX A100 80GB GPUs. 

\paragraph{Evaluation Protocol:} We use the GRES test set to evaluate segmentation performance. We focus on two subsets of the GRES test set, Small and Large, to evaluate performance on small and large objects, respectively. We define a threshold of $500$ pixels$^{2}$ and categorize any object in the test set that covers an area less than the threshold to be \texttt{Small} and bigger to be \texttt{Large}. We evaluate segmentation performance using generalized Intersection-over-Union (gIoU) and cumulative Intersection-over-Union (cIoU) \cite{lai2024lisa}. To evaluate the performance of our approach on traditional vision and language tasks, we use several existing datasets, including NWPU-Captions \cite{cheng2022nwpu}, UCM-Captions \cite{qu2016deep}, Sydney-Captions \cite{qu2016deep}, and RSICD \cite{lu2017exploring}. Following prior work, we report standard evaluation metrics: BLEU \cite{papineni2002bleu}, CIDEr \cite{vedantam2015cider}, and SPICE \cite{anderson2016spice}.

\subsection{Segmentation}

\begin{table}[t]
    \centering
    \small
    \caption{Comparative performance of \lisat against LISA-7B-v1 and LISA-13B-Llama2-v1 on GRES across different object sizes. \lisat-7B consistently outperforms the baseline models, particularly in the Small object category.\vspace{0.3em}}
    
    \begin{tabularx}{\linewidth}{Xccc}
        \toprule
        \textbf{Model} & \textbf{Obj. Size} & \textbf{cIoU} & \textbf{gIoU} \\
        \midrule
         LISA-7B    & All & $0.122_{\pm 0.014}$ & $0.113_{\pm 0.007}$ \\
         & Small  & $0.104_{\pm 0.022}$ & $0.062_{\pm 0.008}$ \\
             & Large & $0.157_{\pm 0.017}$ & $0.222_{\pm 0.013}$ \\
        \midrule
        LISA-13B & All  & $0.122_{\pm 0.014}$ & $0.139_{\pm 0.006}$ \\
        (llama2)                                    & Small  & $0.106_{\pm 0.016}$ & $0.089_{\pm 0.007}$ \\
                                           & Large  & $0.148_{\pm 0.018}$ & $0.244_{\pm 0.019}$ \\
                                            
        \midrule
        \lisat (Ours) & All  & $\textbf{0.245}_{\pm 0.023}$ & $\textbf{0.275}_{\pm 0.009}$ \\
         & Small  & $\textbf{0.232}_{\pm 0.024}$ & $\textbf{0.240}_{\pm 0.009}$ \\
        & Large  & $\textbf{0.250}_{\pm 0.029}$ & $\textbf{0.348}_{\pm 0.015}$ \\
        
        \bottomrule
    \end{tabularx}
    \label{lisa_models_performance}
\end{table}

\begin{figure}[t]
    \centering
    \includegraphics[width=\columnwidth]{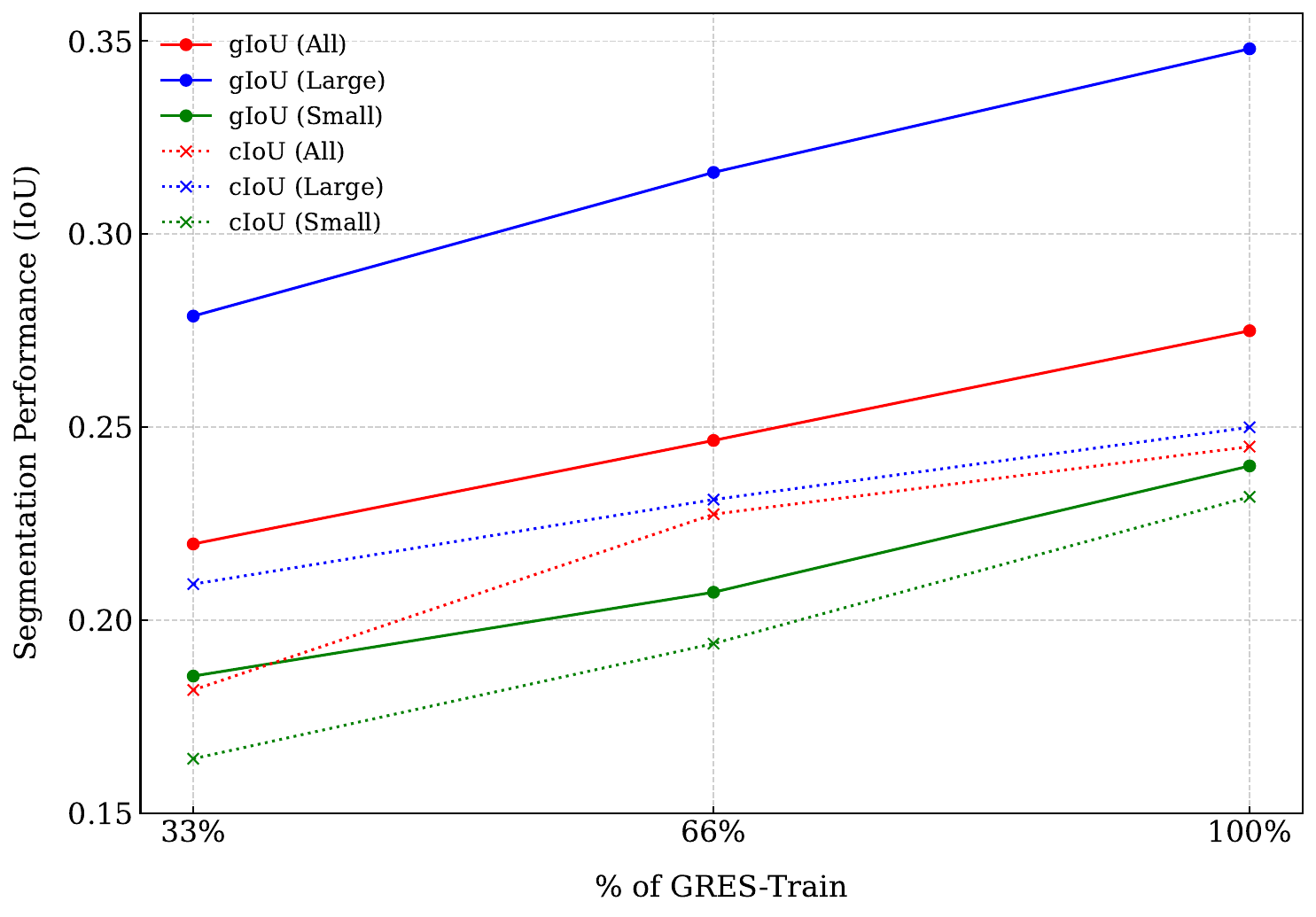} 
    \caption{Scaling behavior of \lisat on the GRES dataset. While adding additional data is helpful, even with $7K$ \textit{training} images (the full GRES dataset), we observe the beginning of a plateau in performance, particularly on cIOU scores. This suggests that more data alone may not be helpful, and instead, we may need additional data variance outside the \texttt{xView} classes.}
    \label{scaling_lisat} 
\end{figure}

\begin{table}[t]
    \centering
    \small
    \caption{Comparison of \lisat's performance using GeoSAM vs. SAM for segmentation on the \texttt{All} dataset configuration.\vspace{0.3em}}
    \begin{tabularx}{\linewidth}{Xccc}
        \toprule
        \textbf{Segmentation Model} & \textbf{cIoU} & \textbf{gIoU} \\
        \midrule
        GeoSAM \cite{sultana2023geosam} & $0.220_{\pm 0.019}$ & $0.238_{\pm 0.007}$ \\
        SAM \cite{kirillov2023segment}  & $\textbf{0.245}_{\pm 0.023}$ & $\textbf{0.275}_{\pm 0.009}$ \\
        \bottomrule
    \end{tabularx}
    \label{tab:geo_sam_vs}
\end{table}

\autoref{lisa_models_performance} compares \lisat with LISA-7B-v1 and LISA-13B-Llama2-v1 \cite{lai2024lisa} across different dataset configurations (All, Small, Large). \lisat consistently and significantly outperforms both natural-image trained referring segmentation models. Notably, for smaller objects, \lisat has larger relative gains compared to large models, suggesting that \lisat is more effective for capturing fine-grained spatial details, which is important for applications involving dense scenes or small-scale features in remote sensing imagery.

Some qualitative examples are given in \autoref{tab:lisat_qual}. The first three rows represent success cases, where \lisat correctly identifies and localizes objects based on the queries. In the first, \lisat correctly segments the building against a noisy background, and when many of the ground features match the visual features of the target object. In the second and third, \lisat correctly identifies the key object of interest, ignoring other potential distractor objects. In the failure case, \lisat fails to correctly identify the barge alone from the two ships, likely due to the color patterns on the first ship, but still manages to outperform LISA, which only focuses on the larger ship objects. 

\autoref{scaling_lisat} demonstrates the influence of training dataset size on \lisat’s performance. With an increasing number of training images, \lisat demonstrates notable improvements in both cIoU and gIoU scores. These results indicate that \lisat benefits from larger training datasets thereby exhibiting some good scaling properties, as its segmentation performance improves with more data, particularly for small objects.

\autoref{tab:geo_sam_vs} compares \lisat's performance using GeoSAM and SAM as base segmentation models on the \texttt{All} dataset. While both models yield competitive results, SAM achieves slightly higher cIoU (0.245) and gIoU (0.275) than GeoSAM. This suggests that despite being designed for geospatial tasks, GeoSAM alone without specific language-aligned fine-tuning may be limited by training-specific biases, whereas SAM's broader training on diverse natural images enables more adaptable feature extraction, leading to improved segmentation performance.

\subsection{Captioning and Question-Answering}
\label{Geospatial-focused_multimodal_LLM_Results}

\begin{table}[t]
    \centering
    \small
    \caption{Comparison of captioning performance on the UCM-Captions dataset. Results are reported for BLEU-4 and CIDEr metrics.\vspace{0.3em}}
    \begin{tabularx}{\linewidth}{Xcc}
        \toprule 
        \textbf{Method} & \textbf{BLEU-4} & \textbf{CIDEr} \\
        \midrule
        SAA \cite{lobry2020rsvqa} & 64.77 & 294.51 \\
        SD-RSIC \cite{sumbul2020sd} & 53.80 & 213.20 \\
        RTRMN (semantic) \cite{wang2020retrieval} & 35.87 & 180.25 \\
        RTRMN (statistical) \cite{wang2020retrieval} & 63.93 & 312.70 \\
        SVM-D BOW \cite{hoxha2021novel} & 51.95 & 271.42 \\
        SVM-D CONC \cite{hoxha2021novel} & 59.42 & 292.28 \\
        Post-processing \cite{huang2023language} & 62.62 & 309.64 \\
        LLaVA-v1.5-7b \cite{liu2024visual} & \ \ 5.54  & \ \ 32.67 \\
        LLaVA-v1.6-7b \cite{liu2024llava} & \ \ 5.44  & \ \ 23.86 \\
        RS-GPT4V \cite{xu2024rs} & 65.74 & 333.23 \\
        
        LISA-7B (fine-tuned on GRES) & 8.73 & 59.96 \\
        
        \lisatpre (Ours) & \textbf{72.34} & \textbf{355.32} \\
        \bottomrule
    \end{tabularx}
    \label{performance_comparison_ucm_captions}
\end{table}

\begin{table}[t]
    \centering
    \small
    \caption{Comparison of captioning performance on the NWPU-Captions dataset. Results are reported for BLEU-4 and SPICE metrics.\vspace{0.3em}}
    \begin{tabularx}{\linewidth}{Xcc}
        \toprule
        \textbf{Method} & \textbf{BLEU-4} & \textbf{SPICE} \\
        \midrule
        CSMLF \cite{touvron2023llama} & 47.1 & 26.5 \\
        Multimodal \cite{qu2016deep} & 45.5 & 27.6 \\
        Attention (hard) \cite{lu2017exploring} & 46.4 & 28.4 \\
        FC-Att \cite{zhang2019description} & 46.9 & 28.3 \\
        MLCA-Net \cite{cheng2022nwpu} & 47.8 & 28.5 \\
        LLaVA-v1.5-7b \cite{liu2024visual}  & \ \ 4.8  & 11.1\\
        LLaVA-v1.6-7b \cite{liu2024llava}  & \ \ 2.9  & \ \ 8.7\\
        EarthGPT\cite{zhang2024earthgpt} & 65.5 & \textbf{32.2} \\ 

        
        LISA-7B (fine-tuned on GRES) & 39.9 & 19.52 \\
        
        \lisatpre (Ours) & \textbf{65.8} & \textbf{32.2} \\
        \bottomrule
    \end{tabularx}
    \label{tab:nwpu}
\end{table}

On the UCM-Captions dataset (\autoref{performance_comparison_ucm_captions}), \lisatpre achieves the highest BLEU-4 (72.34) and CIDEr (355.32) scores, surpassing previous geospatial models such as RS-GPT4V \cite{xu2024rs} and post-processing methods \cite{huang2023language}, as well as general-purpose vision-language models such as LLaVA-v1.5 and LLaVA-v1.6 \cite{liu2024visual, liu2024llava}. For NWPU-Captions (\autoref{tab:nwpu}), \lisatpre achieves the highest BLEU-4 score and matches the best SPICE performance, outperforming prior geospatial captioning models such as MLCA-Net \cite{cheng2022nwpu} and multimodal attention-based methods \cite{lu2017exploring}. General-purpose vision-language models (LLaVA-v1.5 and LLaVA-v1.6) \cite{liu2024visual, liu2024llava} perform significantly worse, highlighting the benefits of domain-specific training. Similar trends are observed on RSICD (\autoref{performance_comparison_RSICD}) and Sydney-Captions (\autoref{performance_comparison_sydney_captions}).

\begin{table}[t]
\centering
\tiny
\caption{Performance on RSVQA-LR (\% accuracy).\vspace{0.3em}}
\begin{tabularx}{\linewidth}{Xccc}
\toprule
\textbf{Model}   & \textbf{Count} & \textbf{Presence} & \textbf{Comparison} \\ \midrule
RSVQA \cite{lobry2020rsvqa}                    & 67.01          & 87.46             & 81.50               \\
EasyToHard \cite{yuan2022easy}               & 69.22          & 90.66             & 87.49               \\
Bi-Modal \cite{bazi2022bi}                 & \textbf{72.22} & 91.06             & 91.16               \\
SHRNet \cite{zhang2023spatial}                   & 73.87          & 91.03             & 90.48               \\
LLaVA-1.5 \cite{liu2024visual}                & 26.81          & 54.72             & 66.22               \\
InternLM-XC2 \cite{internlmxcomposer2}             & 26.91          & 55.74             & 64.89               \\
RS-GPT4V \cite{xu2024rs}                    & -              & 91.17             & 91.70               \\
GeoChat \cite{zhang2023multi}                  & -              & 91.09             & 90.33               \\ 
Full-FT \cite{xu2024rs}                   & 70.48          & 91.10             & 92.23               \\
RS-GPT4V-LoRA-FT \cite{xu2024rs}                     & 70.34          & 92.24             & 92.10               \\
RS-GPT4V-MoE-LoRA-FT \cite{xu2024rs}                 & 71.06          & 91.10             & \textbf{92.55}      \\

LLaVA-v1.5-7b \cite{liu2024visual}  & 18.66 & 53.98 & 66.22  \\
LLaVA-v1.6-7b \cite{liu2024llava}  & 19.65 & 57.53 & 62.32   \\

        
LISA-7B (fine-tuned on GRES) & 25.86 & 79.80 & 84.41\\
        
\lisatpre (Ours) & 70.24       & \textbf{92.36}    &  92.20             \\ \bottomrule
\end{tabularx}
\label{tab:rsvqa}
\end{table}

\autoref{tab:rsvqa} presents the performance of \lisatpre on the RSVQA-LR dataset across Count, Presence, and Comparison categories. The model achieves the highest Presence accuracy (92.36) and Comparison accuracy (92.20), indicating strong performance in these tasks. In contrast, models such as LLaVA-1.5 and InternLM-XC2 report lower scores in Count and Presence. These results suggest that \lisatpre effectively handles multimodal reasoning and task-specific fine-tuning, particularly in Presence-based evaluations.

\begin{table}[t]
\centering
\small
\caption{Ablations of the base language model and visual feature extractor for \lisatpre on the UCM-Captions dataset.\vspace{0.3em}}
\begin{tabularx}{\linewidth}{lXccc}
        \toprule
        \textbf{Vision Encoder} & \textbf{LLM} & \textbf{BLEU-4} & \textbf{CIDEr} & \textbf{SPICE} \\
        \midrule
        CLIP    & Llama 2 & 69.03 & 328.82 & 52.21 \\

        CLIP336    & Llama 2 & 66.97 & 324.61 & 50.46 \\

        SAT-CLIP    & Llama 2  & 8.82  & 30.41  & 8.15  \\
        
        Geo-CLIP    & Llama 2  & 12.77 & 44.64  & 11.67 \\
        
        RemoteCLIP  & Llama 2  & 68.31 & 330.94 & 52.17 \\
        
        CLIP        & Vicuna & 66.68 & 329.32 & 52.00 \\
        CLIP336     & Vicuna & 68.28 & 324.89 & 51.55 \\
        SAT-CLIP    & Vicuna & 16.87 & 63.92  & 15.08 \\
        Geo-CLIP    & Vicuna & 24.56 & 109.20 & 21.15 \\
        
        RemoteCLIP  & Vicuna & \textbf{72.34} & \textbf{355.32} & \textbf{54.15} \\
        \bottomrule
\end{tabularx}
\label{tab:vision_language_ablation}
\end{table}

The ablation study in \autoref{tab:vision_language_ablation} evaluates different vision encoders and language models for \lisatpre on the UCM-Captions dataset. Among the vision encoders, RemoteCLIP (which we use in \lisat) significantly outperforms both Geo-CLIP and Sat-CLIP on all domains, while slightly outperforming the base CLIP models. Models using LLama 2 as a base LLM are notably worse than Vicuna. These findings highlight that both the vision encoder and the language model play crucial roles, with RemoteCLIP and Vicuna forming the most effective pairing for remote sensing imagery.

\subsection{Limitations and Failure Cases}
\label{limitations}

While \lisat outperforms all existing reasoning segmentation models, it is not perfect. \autoref{Failure_Cases_of_LISAt} highlights examples of failure cases in our pipeline. In some instances, LISAt struggles to produce accurate predictions when images are cloudy or when key features are obscured. Other challenges arise when the query is too vague like \texttt{``Identify the plane in
the bottom-right of the image.``} while there are several planes in the bottom right corner of the image. It is also the case when similar objects to the target item appear in the image, leading to ambiguity. We hypothesize that training on a larger dataset and refining the query design could help mitigate these issues. Another issue arises from the ground truth masks generated by GeoSAM in the GRES dataset. In some cases, the underlying ground truth mask is incorrect, and LISAt is occasionally penalized even when making correct predictions, as demonstrated in \autoref{Ground_Truth_Error_Cases}.

\section{Conclusion}
\label{conclusion}

In this paper we introduce \lisat, an open-source, open-data foundation model for geospatial reasoning segmentation, and GRES, an open dataset to help support applications in remote-sensing referring segmentation. \lisat is only the first step towards models that can produce text and task-specific outputs such as masks and boxes when reasoning about a geospatial world. Future work can focus on scaling \lisat for use with large rasters, integrate \lisat with additional segmentation models other than SAM and GeoSAM, or incorporate multimodal/hyperspectral data sources.  Overall, we hope that \lisat lays the groundwork for future advancements in geospatial artificial intelligence, paving the way for more sophisticated models that seamlessly integrate vision and language to better understand and interact with our dynamic geospatial world.

\section*{Impact Statement}
This paper presents advancements in reasoning segmentation for remote sensing tasks. \lisat is a method that is able to reason over arbitrary remote sensing images and output both explanations and segmentation masks for objects of interest. These kinds of workflows are extremely common across multiple fields. For example, disaster management personnel may want to know which roads leading to an airport are undamaged, and why. \lisat is the first such model that can simultaneously answer both components of such questions.

Broadly, \lisat has impacts in numerous domains such as environmental monitoring, urban planning, and search and rescue. However, one of the biggest uses of satellite imaging is surveillance. Being cognizant of this, our work is primarily based on datasets that have been widely adopted by the remote sensing community over interesting, cluttered scenes that do not capture any individual entity.

We encourage responsible deployment and continued discourse on the implications of geospatial AI in real-world applications.

\bibliography{example_paper}

\begin{thebibliography}{59}
\providecommand{\natexlab}[1]{#1}
\providecommand{\url}[1]{\texttt{#1}}
\expandafter\ifx\csname urlstyle\endcsname\relax
  \providecommand{\doi}[1]{doi: #1}\else
  \providecommand{\doi}{doi: \begingroup \urlstyle{rm}\Url}\fi

\bibitem[Achiam et~al.(2023)Achiam, Adler, Agarwal, Ahmad, Akkaya, Aleman, Almeida, Altenschmidt, Altman, Anadkat, et~al.]{achiam2023gpt}
Achiam, J., Adler, S., Agarwal, S., Ahmad, L., Akkaya, I., Aleman, F.~L., Almeida, D., Altenschmidt, J., Altman, S., Anadkat, S., et~al.
\newblock Gpt-4 technical report.
\newblock \emph{arXiv preprint arXiv:2303.08774}, 2023.

\bibitem[Anderson et~al.(2016)Anderson, Fernando, Johnson, and Gould]{anderson2016spice}
Anderson, P., Fernando, B., Johnson, M., and Gould, S.
\newblock Spice: Semantic propositional image caption evaluation.
\newblock In \emph{Computer Vision--ECCV 2016: 14th European Conference, Amsterdam, The Netherlands, October 11-14, 2016, Proceedings, Part V 14}, pp.\  382--398. Springer, 2016.

\bibitem[Bazi et~al.(2022)Bazi, Al~Rahhal, Mekhalfi, Al~Zuair, and Melgani]{bazi2022bi}
Bazi, Y., Al~Rahhal, M.~M., Mekhalfi, M.~L., Al~Zuair, M.~A., and Melgani, F.
\newblock Bi-modal transformer-based approach for visual question answering in remote sensing imagery.
\newblock \emph{IEEE Transactions on Geoscience and Remote Sensing}, 60:\penalty0 1--11, 2022.

\bibitem[Chen et~al.(2023)Chen, Zhang, Zeng, Zhang, Zhu, and Zhao]{chen2023shikra}
Chen, K., Zhang, Z., Zeng, W., Zhang, R., Zhu, F., and Zhao, R.
\newblock Shikra: Unleashing multimodal llm's referential dialogue magic.
\newblock \emph{arXiv preprint arXiv:2306.15195}, 2023.

\bibitem[Chen et~al.(2024)Chen, Qu, Zhang, Liu, Wang, and Zhang]{chen2024edge}
Chen, L., Qu, Z., Zhang, Y., Liu, J., Wang, R., and Zhang, D.
\newblock Edge enhanced gciffnet: A multiclass semantic segmentation network based on edge enhancement and multiscale attention mechanism.
\newblock \emph{IEEE Journal of Selected Topics in Applied Earth Observations and Remote Sensing}, 2024.

\bibitem[Cheng et~al.(2017)Cheng, Han, and Lu]{cheng2017remote}
Cheng, G., Han, J., and Lu, X.
\newblock Remote sensing image scene classification: Benchmark and state of the art.
\newblock \emph{Proceedings of the IEEE}, 105\penalty0 (10):\penalty0 1865--1883, 2017.

\bibitem[Cheng et~al.(2022)Cheng, Huang, Xu, Zhou, Li, and Wang]{cheng2022nwpu}
Cheng, Q., Huang, H., Xu, Y., Zhou, Y., Li, H., and Wang, Z.
\newblock Nwpu-captions dataset and mlca-net for remote sensing image captioning.
\newblock \emph{IEEE Transactions on Geoscience and Remote Sensing}, 60:\penalty0 1--19, 2022.

\bibitem[Chiang et~al.(2023)Chiang, Li, Lin, Sheng, Wu, Zhang, Zheng, Zhuang, Zhuang, Gonzalez, Stoica, and Xing]{vicuna2023}
Chiang, W.-L., Li, Z., Lin, Z., Sheng, Y., Wu, Z., Zhang, H., Zheng, L., Zhuang, S., Zhuang, Y., Gonzalez, J.~E., Stoica, I., and Xing, E.~P.
\newblock Vicuna: An open-source chatbot impressing gpt-4 with 90\%* chatgpt quality, March 2023.
\newblock URL \url{https://lmsys.org/blog/2023-03-30-vicuna/}.

\bibitem[Dong et~al.(2024)Dong, Zhang, Zang, Cao, Wang, Ouyang, Wei, Zhang, Duan, Cao, Zhang, Li, Yan, Gao, Zhang, Li, Li, Chen, He, Zhang, Qiao, Lin, and Wang]{internlmxcomposer2}
Dong, X., Zhang, P., Zang, Y., Cao, Y., Wang, B., Ouyang, L., Wei, X., Zhang, S., Duan, H., Cao, M., Zhang, W., Li, Y., Yan, H., Gao, Y., Zhang, X., Li, W., Li, J., Chen, K., He, C., Zhang, X., Qiao, Y., Lin, D., and Wang, J.
\newblock Internlm-xcomposer2: Mastering free-form text-image composition and comprehension in vision-language large model.
\newblock \emph{arXiv preprint arXiv:2401.16420}, 2024.

\bibitem[Flotzinger et~al.(2024)Flotzinger, R{\"o}sch, Benz, Ahmad, Cankaya, Mayer, Rodehorst, Oswald, and Braml]{flotzinger2024dacl}
Flotzinger, J., R{\"o}sch, P.~J., Benz, C., Ahmad, M., Cankaya, M., Mayer, H., Rodehorst, V., Oswald, N., and Braml, T.
\newblock dacl-challenge: Semantic segmentation during visual bridge inspections.
\newblock In \emph{Proceedings of the IEEE/CVF Winter Conference on Applications of Computer Vision}, pp.\  716--725, 2024.

\bibitem[Gao et~al.(2024)Gao, Ao, Wang, Zhao, Ma, Xie, Fu, Ren, and Gao]{gao2024enrich}
Gao, T., Ao, W., Wang, X.-A., Zhao, Y., Ma, P., Xie, M., Fu, H., Ren, J., and Gao, Z.
\newblock Enrich distill and fuse: Generalized few-shot semantic segmentation in remote sensing leveraging foundation model's assistance.
\newblock In \emph{Proceedings of the IEEE/CVF Conference on Computer Vision and Pattern Recognition}, pp.\  2771--2780, 2024.

\bibitem[Hansen et~al.(2024)Hansen, Jensen, Philipsen, M{\o}gelmose, Bodum, and Moeslund]{hansen2024opentrench3d}
Hansen, L.~H., Jensen, S.~B., Philipsen, M.~P., M{\o}gelmose, A., Bodum, L., and Moeslund, T.~B.
\newblock Opentrench3d: A photogrammetric 3d point cloud dataset for semantic segmentation of underground utilities.
\newblock In \emph{Proceedings of the IEEE/CVF Conference on Computer Vision and Pattern Recognition}, pp.\  7646--7655, 2024.

\bibitem[Hoxha et~al.(2021)]{hoxha2021novel}
Hoxha, G. et~al.
\newblock A novel svm-based decoder for remote sensing image captioning.
\newblock \emph{IEEE Transactions on Geoscience and Remote Sensing}, 60:\penalty0 1--14, 2021.

\bibitem[Hu et~al.(2021)Hu, Shen, Wallis, Allen-Zhu, Li, Wang, Wang, and Chen]{hu2021lora}
Hu, E.~J., Shen, Y., Wallis, P., Allen-Zhu, Z., Li, Y., Wang, S., Wang, L., and Chen, W.
\newblock Lora: Low-rank adaptation of large language models.
\newblock \emph{arXiv preprint arXiv:2106.09685}, 2021.

\bibitem[Hu et~al.(2016)Hu, Rohrbach, and Darrell]{hu2016segmentation}
Hu, R., Rohrbach, M., and Darrell, T.
\newblock Segmentation from natural language expressions.
\newblock In \emph{Computer Vision--ECCV 2016: 14th European Conference, Amsterdam, The Netherlands, October 11--14, 2016, Proceedings, Part I 14}, pp.\  108--124. Springer, 2016.

\bibitem[Huang et~al.(2023)Huang, Dong, Wang, Hao, Singhal, Ma, Lv, Cui, Mohammed, Patra, et~al.]{huang2023language}
Huang, S., Dong, L., Wang, W., Hao, Y., Singhal, S., Ma, S., Lv, T., Cui, L., Mohammed, O.~K., Patra, B., et~al.
\newblock Language is not all you need: Aligning perception with language models.
\newblock \emph{Advances in Neural Information Processing Systems}, 36:\penalty0 72096--72109, 2023.

\bibitem[Irvin et~al.(2024)Irvin, Liu, Chen, Dormoy, Kim, Khanna, Zheng, and Ermon]{irvin2024teochat}
Irvin, J.~A., Liu, E.~R., Chen, J.~C., Dormoy, I., Kim, J., Khanna, S., Zheng, Z., and Ermon, S.
\newblock Teochat: A large vision-language assistant for temporal earth observation data.
\newblock \emph{arXiv preprint arXiv:2410.06234}, 2024.

\bibitem[Kirillov et~al.(2023)Kirillov, Mintun, Ravi, Mao, Rolland, Gustafson, Xiao, Whitehead, Berg, Lo, et~al.]{kirillov2023segment}
Kirillov, A., Mintun, E., Ravi, N., Mao, H., Rolland, C., Gustafson, L., Xiao, T., Whitehead, S., Berg, A.~C., Lo, W.-Y., et~al.
\newblock Segment anything.
\newblock In \emph{Proceedings of the IEEE/CVF International Conference on Computer Vision}, pp.\  4015--4026, 2023.

\bibitem[Kuckreja et~al.(2024)Kuckreja, Danish, Naseer, Das, Khan, and Khan]{kuckreja2024geochat}
Kuckreja, K., Danish, M.~S., Naseer, M., Das, A., Khan, S., and Khan, F.~S.
\newblock Geochat: Grounded large vision-language model for remote sensing.
\newblock In \emph{Proceedings of the IEEE/CVF Conference on Computer Vision and Pattern Recognition}, pp.\  27831--27840, 2024.

\bibitem[Lai et~al.(2024)Lai, Tian, Chen, Li, Yuan, Liu, and Jia]{lai2024lisa}
Lai, X., Tian, Z., Chen, Y., Li, Y., Yuan, Y., Liu, S., and Jia, J.
\newblock Lisa: Reasoning segmentation via large language model.
\newblock In \emph{Proceedings of the IEEE/CVF Conference on Computer Vision and Pattern Recognition}, pp.\  9579--9589, 2024.

\bibitem[Lam et~al.(2018)Lam, Kuzma, McGee, Dooley, Laielli, Klaric, Bulatov, and McCord]{lam2018xview}
Lam, D., Kuzma, R., McGee, K., Dooley, S., Laielli, M., Klaric, M., Bulatov, Y., and McCord, B.
\newblock xview: Objects in context in overhead imagery.
\newblock \emph{arXiv preprint arXiv:1802.07856}, 2018.

\bibitem[Li(2024)]{li2024cpseg}
Li, L.
\newblock Cpseg: Finer-grained image semantic segmentation via chain-of-thought language prompting.
\newblock In \emph{Proceedings of the IEEE/CVF Winter Conference on Applications of Computer Vision}, pp.\  513--522, 2024.

\bibitem[Li et~al.(2018)Li, Yuan, and Lu]{li2018multi}
Li, X., Yuan, A., and Lu, X.
\newblock Multi-modal gated recurrent units for image description.
\newblock \emph{Multimedia Tools and Applications}, 77\penalty0 (22):\penalty0 29847--29869, 2018.

\bibitem[Liu et~al.(2022)Liu, Zhao, and Shi]{liu2022remote}
Liu, C., Zhao, R., and Shi, Z.
\newblock Remote-sensing image captioning based on multilayer aggregated transformer.
\newblock \emph{IEEE Geoscience and Remote Sensing Letters}, 19:\penalty0 1--5, 2022.

\bibitem[Liu et~al.(2024{\natexlab{a}})Liu, Chen, Guan, Zhou, Zhu, Ye, Fu, and Zhou]{liu2024remoteclip}
Liu, F., Chen, D., Guan, Z., Zhou, X., Zhu, J., Ye, Q., Fu, L., and Zhou, J.
\newblock Remoteclip: A vision language foundation model for remote sensing.
\newblock \emph{IEEE Transactions on Geoscience and Remote Sensing}, 2024{\natexlab{a}}.

\bibitem[Liu et~al.(2024{\natexlab{b}})Liu, Li, Li, Li, Zhang, Shen, and Lee]{liu2024llava}
Liu, H., Li, C., Li, Y., Li, B., Zhang, Y., Shen, S., and Lee, Y.~J.
\newblock Llava-next: Improved reasoning, ocr, and world knowledge, 2024{\natexlab{b}}.

\bibitem[Liu et~al.(2024{\natexlab{c}})Liu, Li, Wu, and Lee]{liu2024visual}
Liu, H., Li, C., Wu, Q., and Lee, Y.~J.
\newblock Visual instruction tuning.
\newblock \emph{Advances in neural information processing systems}, 36, 2024{\natexlab{c}}.

\bibitem[Lobry et~al.(2020)Lobry, Marcos, Murray, and Tuia]{lobry2020rsvqa}
Lobry, S., Marcos, D., Murray, J., and Tuia, D.
\newblock Rsvqa: Visual question answering for remote sensing data.
\newblock \emph{IEEE Transactions on Geoscience and Remote Sensing}, 58\penalty0 (12):\penalty0 8555--8566, 2020.

\bibitem[Loshchilov(2017)]{loshchilov2017decoupled}
Loshchilov, I.
\newblock Decoupled weight decay regularization.
\newblock \emph{arXiv preprint arXiv:1711.05101}, 2017.

\bibitem[Lu et~al.(2017)Lu, Wang, Zheng, and Li]{lu2017exploring}
Lu, X., Wang, B., Zheng, X., and Li, X.
\newblock Exploring models and data for remote sensing image caption generation.
\newblock \emph{IEEE Transactions on Geoscience and Remote Sensing}, 56\penalty0 (4):\penalty0 2183--2195, 2017.

\bibitem[Papineni et~al.(2002)Papineni, Roukos, Ward, and Zhu]{papineni2002bleu}
Papineni, K., Roukos, S., Ward, T., and Zhu, W.-J.
\newblock Bleu: a method for automatic evaluation of machine translation.
\newblock In \emph{Proceedings of the 40th annual meeting of the Association for Computational Linguistics}, pp.\  311--318, 2002.

\bibitem[Peng et~al.(2024)Peng, Wang, Dong, Hao, Huang, Ma, Ye, and Wei]{peng2024grounding}
Peng, Z., Wang, W., Dong, L., Hao, Y., Huang, S., Ma, S., Ye, Q., and Wei, F.
\newblock Grounding multimodal large language models to the world.
\newblock In \emph{The Twelfth International Conference on Learning Representations}, 2024.

\bibitem[Qu et~al.(2016)Qu, Li, Tao, and Lu]{qu2016deep}
Qu, B., Li, X., Tao, D., and Lu, X.
\newblock Deep semantic understanding of high resolution remote sensing image.
\newblock In \emph{2016 International conference on computer, information and telecommunication systems (Cits)}, pp.\  1--5. IEEE, 2016.

\bibitem[Radford et~al.(2021)Radford, Kim, Hallacy, Ramesh, Goh, Agarwal, Sastry, Askell, Mishkin, Clark, et~al.]{radford2021learning}
Radford, A., Kim, J.~W., Hallacy, C., Ramesh, A., Goh, G., Agarwal, S., Sastry, G., Askell, A., Mishkin, P., Clark, J., et~al.
\newblock Learning transferable visual models from natural language supervision.
\newblock In \emph{International conference on machine learning}, pp.\  8748--8763. PMLR, 2021.

\bibitem[Rahnemoonfar et~al.(2021)Rahnemoonfar, Chowdhury, Sarkar, Varshney, Yari, and Murphy]{rahnemoonfar2021floodnet}
Rahnemoonfar, M., Chowdhury, T., Sarkar, A., Varshney, D., Yari, M., and Murphy, R.~R.
\newblock Floodnet: A high resolution aerial imagery dataset for post flood scene understanding.
\newblock \emph{IEEE Access}, 9:\penalty0 89644--89654, 2021.

\bibitem[Rasheed et~al.(2024)Rasheed, Maaz, Shaji, Shaker, Khan, Cholakkal, Anwer, Xing, Yang, and Khan]{rasheed2024glamm}
Rasheed, H., Maaz, M., Shaji, S., Shaker, A., Khan, S., Cholakkal, H., Anwer, R.~M., Xing, E., Yang, M.-H., and Khan, F.~S.
\newblock Glamm: Pixel grounding large multimodal model.
\newblock In \emph{Proceedings of the IEEE/CVF Conference on Computer Vision and Pattern Recognition}, pp.\  13009--13018, 2024.

\bibitem[Ren et~al.(2024)Ren, Huang, Wei, Zhao, Fu, Feng, and Jin]{ren2024pixellm}
Ren, Z., Huang, Z., Wei, Y., Zhao, Y., Fu, D., Feng, J., and Jin, X.
\newblock Pixellm: Pixel reasoning with large multimodal model.
\newblock In \emph{Proceedings of the IEEE/CVF Conference on Computer Vision and Pattern Recognition}, pp.\  26374--26383, 2024.

\bibitem[Subudhi et~al.(2021)Subudhi, Patro, Biswal, and Dell’Acqua]{subudhi2021survey}
Subudhi, S., Patro, R.~N., Biswal, P.~K., and Dell’Acqua, F.
\newblock A survey on superpixel segmentation as a preprocessing step in hyperspectral image analysis.
\newblock \emph{IEEE Journal of Selected Topics in Applied Earth Observations and Remote Sensing}, 14:\penalty0 5015--5035, 2021.

\bibitem[Sultana et~al.(2023)Sultana, Lia, Zhua, Khanduria, Brocanellib, and Zhua]{sultana2023geosam}
Sultana, R.~I., Lia, C., Zhua, H., Khanduria, P., Brocanellib, M., and Zhua, D.
\newblock Geosam: Fine-tuning sam with multi-modal prompts for mobility infrastructure segmentation.
\newblock \emph{arXiv preprint arXiv:2311.11319}, 2023.

\bibitem[Sumbul et~al.(2020)Sumbul, Nayak, and Demir]{sumbul2020sd}
Sumbul, G., Nayak, S., and Demir, B.
\newblock Sd-rsic: Summarization-driven deep remote sensing image captioning.
\newblock \emph{IEEE Transactions on Geoscience and Remote Sensing}, 59\penalty0 (8):\penalty0 6922--6934, 2020.

\bibitem[Touvron et~al.(2023)Touvron, Lavril, Izacard, Martinet, Lachaux, Lacroix, Rozi{\`e}re, Goyal, Hambro, Azhar, et~al.]{touvron2023llama}
Touvron, H., Lavril, T., Izacard, G., Martinet, X., Lachaux, M.-A., Lacroix, T., Rozi{\`e}re, B., Goyal, N., Hambro, E., Azhar, F., et~al.
\newblock Llama: open and efficient foundation language models. arxiv.
\newblock \emph{arXiv preprint arXiv:2302.13971}, 2023.

\bibitem[Vedantam et~al.(2015)Vedantam, Lawrence~Zitnick, and Parikh]{vedantam2015cider}
Vedantam, R., Lawrence~Zitnick, C., and Parikh, D.
\newblock Cider: Consensus-based image description evaluation.
\newblock In \emph{Proceedings of the IEEE conference on computer vision and pattern recognition}, pp.\  4566--4575, 2015.

\bibitem[Wang et~al.(2020)Wang, Zheng, Qu, and Lu]{wang2020retrieval}
Wang, B., Zheng, X., Qu, B., and Lu, X.
\newblock Retrieval topic recurrent memory network for remote sensing image captioning.
\newblock \emph{IEEE Journal of Selected Topics in Applied Earth Observations and Remote Sensing}, 13:\penalty0 256--270, 2020.

\bibitem[Weiss et~al.(2020)Weiss, Jacob, and Duveiller]{weiss2020remote}
Weiss, M., Jacob, F., and Duveiller, G.
\newblock Remote sensing for agricultural applications: A meta-review.
\newblock \emph{Remote sensing of environment}, 236:\penalty0 111402, 2020.

\bibitem[Wu et~al.(2024)Wu, Biamby, Chan, Dunlap, Gupta, Wang, Gonzalez, and Darrell]{wu2024see}
Wu, T.-H., Biamby, G., Chan, D., Dunlap, L., Gupta, R., Wang, X., Gonzalez, J.~E., and Darrell, T.
\newblock See say and segment: Teaching lmms to overcome false premises.
\newblock In \emph{Proceedings of the IEEE/CVF Conference on Computer Vision and Pattern Recognition}, pp.\  13459--13469, 2024.

\bibitem[Xia et~al.(2024)Xia, Han, Han, Pan, Song, and Huang]{xia2024gsva}
Xia, Z., Han, D., Han, Y., Pan, X., Song, S., and Huang, G.
\newblock Gsva: Generalized segmentation via multimodal large language models.
\newblock In \emph{Proceedings of the IEEE/CVF Conference on Computer Vision and Pattern Recognition}, pp.\  3858--3869, 2024.

\bibitem[Xu(2015)]{xu2015show}
Xu, K.
\newblock Show, attend and tell: Neural image caption generation with visual attention.
\newblock \emph{arXiv preprint arXiv:1502.03044}, 2015.

\bibitem[Xu et~al.(2024)Xu, Zhao, Guo, Li, Long, Zou, Wang, and Li]{xu2024rs}
Xu, L., Zhao, L., Guo, W., Li, Q., Long, K., Zou, K., Wang, Y., and Li, H.
\newblock Rs-gpt4v: A unified multimodal instruction-following dataset for remote sensing image understanding.
\newblock \emph{arXiv preprint arXiv:2406.12479}, 2024.

\bibitem[Xuan et~al.(2024)Xuan, Guo, Yang, and Zhang]{xuan2024pink}
Xuan, S., Guo, Q., Yang, M., and Zhang, S.
\newblock Pink: Unveiling the power of referential comprehension for multi-modal llms.
\newblock In \emph{Proceedings of the IEEE/CVF Conference on Computer Vision and Pattern Recognition}, pp.\  13838--13848, 2024.

\bibitem[Yuan et~al.(2022{\natexlab{a}})Yuan, Mou, Wang, and Zhu]{yuan2022easy}
Yuan, Z., Mou, L., Wang, Q., and Zhu, X.~X.
\newblock From easy to hard: Learning language-guided curriculum for visual question answering on remote sensing data.
\newblock \emph{IEEE transactions on geoscience and remote sensing}, 60:\penalty0 1--11, 2022{\natexlab{a}}.

\bibitem[Yuan et~al.(2022{\natexlab{b}})Yuan, Zhang, Fu, Li, Deng, Wang, and Sun]{yuan2022exploring}
Yuan, Z., Zhang, W., Fu, K., Li, X., Deng, C., Wang, H., and Sun, X.
\newblock Exploring a fine-grained multiscale method for cross-modal remote sensing image retrieval.
\newblock \emph{arXiv preprint arXiv:2204.09868}, 2022{\natexlab{b}}.

\bibitem[Zhan et~al.(2023)Zhan, Xiong, and Yuan]{zhan2023rsvg}
Zhan, Y., Xiong, Z., and Yuan, Y.
\newblock Rsvg: Exploring data and models for visual grounding on remote sensing data.
\newblock \emph{IEEE Transactions on Geoscience and Remote Sensing}, 61:\penalty0 1--13, 2023.

\bibitem[Zhan et~al.(2024)Zhan, Xiong, and Yuan]{zhan2024skyeyegpt}
Zhan, Y., Xiong, Z., and Yuan, Y.
\newblock Skyeyegpt: Unifying remote sensing vision-language tasks via instruction tuning with large language model.
\newblock \emph{arXiv preprint arXiv:2401.09712}, 2024.

\bibitem[Zhang et~al.(2023{\natexlab{a}})Zhang, Chen, and Li]{zhang2023multi}
Zhang, M., Chen, F., and Li, B.
\newblock Multi-step question-driven visual question answering for remote sensing.
\newblock \emph{IEEE Transactions on Geoscience and Remote Sensing}, 2023{\natexlab{a}}.

\bibitem[Zhang et~al.(2024{\natexlab{a}})Zhang, Cai, Zhang, Zhuang, and Mao]{zhang2024earthgpt}
Zhang, W., Cai, M., Zhang, T., Zhuang, Y., and Mao, X.
\newblock Earthgpt: A universal multi-modal large language model for multi-sensor image comprehension in remote sensing domain.
\newblock \emph{IEEE Transactions on Geoscience and Remote Sensing}, 2024{\natexlab{a}}.

\bibitem[Zhang et~al.(2019)Zhang, Wang, Tang, Zhou, and Li]{zhang2019description}
Zhang, X., Wang, X., Tang, X., Zhou, H., and Li, C.
\newblock Description generation for remote sensing images using attribute attention mechanism.
\newblock \emph{Remote Sensing}, 11\penalty0 (6):\penalty0 612, 2019.

\bibitem[Zhang et~al.(2024{\natexlab{b}})Zhang, Ma, Gao, Shakiah, Gao, and Chai]{zhang2024groundhog}
Zhang, Y., Ma, Z., Gao, X., Shakiah, S., Gao, Q., and Chai, J.
\newblock Groundhog: Grounding large language models to holistic segmentation.
\newblock In \emph{Proceedings of the IEEE/CVF conference on computer vision and pattern recognition}, pp.\  14227--14238, 2024{\natexlab{b}}.

\bibitem[Zhang et~al.(2023{\natexlab{b}})Zhang, Jiao, Li, Liu, Chen, Liu, Li, and Guo]{zhang2023spatial}
Zhang, Z., Jiao, L., Li, L., Liu, X., Chen, P., Liu, F., Li, Y., and Guo, Z.
\newblock A spatial hierarchical reasoning network for remote sensing visual question answering.
\newblock \emph{IEEE Transactions on Geoscience and Remote Sensing}, 61:\penalty0 1--15, 2023{\natexlab{b}}.

\bibitem[Zheng et~al.(2021)Zheng, Wang, Du, and Lu]{zheng2021mutual}
Zheng, X., Wang, B., Du, X., and Lu, X.
\newblock Mutual attention inception network for remote sensing visual question answering.
\newblock \emph{IEEE Transactions on Geoscience and Remote Sensing}, 60:\penalty0 1--14, 2021.

\end{thebibliography}
\bibliographystyle{icml2025}

\newpage
\appendix

\renewcommand{\theequation}{\thesection.\arabic{equation}}
\renewcommand{\thefigure}{\thesection.\arabic{figure}}
\renewcommand{\thetable}{\thesection.\arabic{table}}

\onecolumn
\section*{Appendix}

In this appendix, we include several additional discussions:

\begin{itemize}
\item \autoref{code_release} details the code release, including links to the codebases and datasets used in this project.
\item \autoref{More_on_the_LISAt_Dataset} outlines the prompt structure used for engineering the GRES dataset for \lisat and provides further details on its class distribution and quality verification.
\item \autoref{More_on_the_RS_LLaVA_Dataset} presents additional details on the PreGRES dataset used to fine-tune \lisatpre, discussing its composition and further evaluations.
\item \autoref{Qualitative_Analysis} showcases qualitative results, highlighting both successful and failure cases, as well as instances where \lisat was penalized due to incomplete Ground Truth annotations generated by GeoSAM (GT).
\end{itemize}

\section{Code Release}
\label{code_release}

The project page for this paper is available \href{https://lisat-bair.github.io/LISAt/}{here}. Our code for \lisat, derived from the Apache 2.0-licensed LISA codebase \citep{lai2024lisa}, as well as the curated datasets are publicly released under the MIT license (or their respective licenses) and could also be found on the same page.

\section{More on GRES}
\label{More_on_the_LISAt_Dataset}

\subsection{Prompt Engineering}
\label{Prompt_Engineering}

As outlined in Section \ref{sec:gres}, we used GPT-4o to generate the final prompt in two stages, detailed below.

\subsubsection{Promt Engineering Stage 1}
\label{Prompt_Engineering_1}

In the first stage, we input a $512 \times 512$ chip into the model and prompt it, following the template below, to generate a sentence that accurately describes the item within the bounding box provided, as specified by the Ground Truth from \texttt{xView}

\begin{quote}
\begin{lstlisting}[basicstyle=\ttfamily\scriptsize, frame=single, breaklines=true]
The size of the original image is (512,512).
This original image, where the image's origin is at the top left corner, contains the following objects: {classes_list_str}.
Only focus on {class_name} in the image.
If {class_name} contains the word 'Other', remove the word 'Other' and use only the second word in {class_name} describing the class. In that case, make sure that second word in {class_name} starts with a lowercase letter.
The following are the bounding boxes [x, y, width, height] of objects of class {class_name}, where (x,y) represents the top left corner of the bounding box, and 'width' represents the bounding box's width, and 'height' represents the bounding box's height.
The bounding box of the {class_name} is at coordinates {bbox}.
Find visual features (color, shape, size, etc.) that can help find or segment {class_name} in the image.
Generate a sentence (not a question) that can uniquely segment or identify or find or locate {class_name} in this image, be concise and clear.
\end{lstlisting}
\label{object_identification_prompt_stage_1}
\end{quote}

Where \texttt{\{classes\_list\_str\}}, \texttt{\{class\_name\}}, and \texttt{\{bbox\}} are the ground truth list of classes, the object class name or category, and the bounding box of the object from the \texttt{xView} dataset bounding box and class annotations.

The model outputs a descriptive sentence in the variable \texttt{\{unique\_characteristics.query\}}, which is then used to query the model again in the second stage, as shown below.

\subsubsection{Prompt Engineering stage 2}
\label{Prompt_Engineering_2}

Once the uniquely descriptive sentence is generated, we asked the model using the template below to come with a question to which the given sentence in \texttt{\{unique\_characteristics.query\}} will be the answer.

\begin{minipage}{\linewidth}
\begin{quote}
\begin{lstlisting}[basicstyle=\ttfamily\scriptsize, frame=single, breaklines=true]
The size of the original image is (512,512).
Only focus on {class_name} in the image.
In the original image, where the image's origin is at the top left corner, the object is a {class_name} located at bounding box coordinates {bbox}.
The following are the bounding boxes [x, y, width, height] of objects of class {class_name}, where (x,y) represents the top left corner of the bounding box, and 'width' represents the bounding box's width, and 'height' represents the bounding box's height:
This original image, where the image's origin is at the top left corner, contains the following objects: {classes_list_str}.
If {class_name} contains the word 'Other', remove the word 'Other' and use only the second word in {class_name} describing the class. In that case, make sure that second word in {class_name} starts with a lowercase letter.
{ ' located at bounding box coordinates {bbox}.' if include_bbox else '.'}
Please generate a query that would help locate this {class_name} in the original image.
Your query will be the question to the answer provided by {unique_characteristics.query}.
For example, if the value contained in {unique_characteristics.query} is 'Look for a long rectangular shape with distinct wheels, typically metallic or painted in color, connected to a truck cab at the front', your query should be:
'Segment the blue car in the bottom right of the image with a long rectangular shape with distinct wheels, typically metallic or painted in color, connected to a truck cab at the front'
'Identify the blue car in the bottom right of the image with a long rectangular shape with distinct wheels, typically metallic or painted in color, connected to a truck cab at the front'
'Find the blue car in the bottom right of the image with a long rectangular shape with distinct wheels, typically metallic or painted in color, connected to a truck cab at the front'
'Locate the blue car in the bottom right of the image with a long rectangular shape with distinct wheels, typically metallic or painted in color, connected to a truck cab at the front'
'Show the blue car in the bottom right of the image with a long rectangular shape with distinct wheels, typically metallic or painted in color, connected to a truck cab at the front'.
Generate the query considering the sentence: {unique_characteristics.query}
{ 'and the location described by the bounding box.' if include_bbox else '.'}
Make sure to vary the start of your queries with key words such as 'Segment, Find, Locate, Show, Identify' and similar synonyms. Do not overuse one over the others.
Rephrase the generated query to make it sound better.
{ 'Do not mention or use any location-related info such as: top, near the center in your query.' if not include_bbox else ''}
Do not output the exact bounding box coordinates, instead, output the locations such as: bottom-left, top-right, top-left, bottom-right, center, etc.
The response to the generated queries should be a JSON object in the following format and contain nothing else:
The response to the generated query should be a sentence, not a question.
Be concise and clear, start the sentence with: Locate, Segment, or Identify.
{"query": "<your_query_here>"}
\end{lstlisting}
\label{object_identification_prompt_stage_2}
\end{quote}
\end{minipage}

Where \texttt{\{class\_name\}}, \texttt{\{bbox\}}, \texttt{\{unique\_characteristics.query\}}, and \texttt{\{class\_name\}} are the ground truth class name or category of the object class name or category, its bounding box and the unique characteristics obtained from GPT-4 \cite{achiam2023gpt} in the first stage.

The final query is then treated as the principal query. To enhance query diversity, we ask GPT to rephrase the principal query into two additional variants, resulting in three distinct queries per image. We then use GeoSAM to generate corresponding masks, forming image-queries-mask tuples.

\subsection{Dataset Quality Assurance}
\label{dataset_quality_assurance}

We use RGB images from the \texttt{xView} dataset \cite{lam2018xview}, as referenced in our manuscript. We employ \textit{Human Verification}, where multiple team members manually inspect randomly selected subsets of the dataset to verify the accuracy of query-image-annotation triplets.


\subsection{GRES Dataset Summary}
\label{LISAt_Dataset_Summary}

\begin{figure*}[h]
    \centering
    \includegraphics[width=\linewidth]{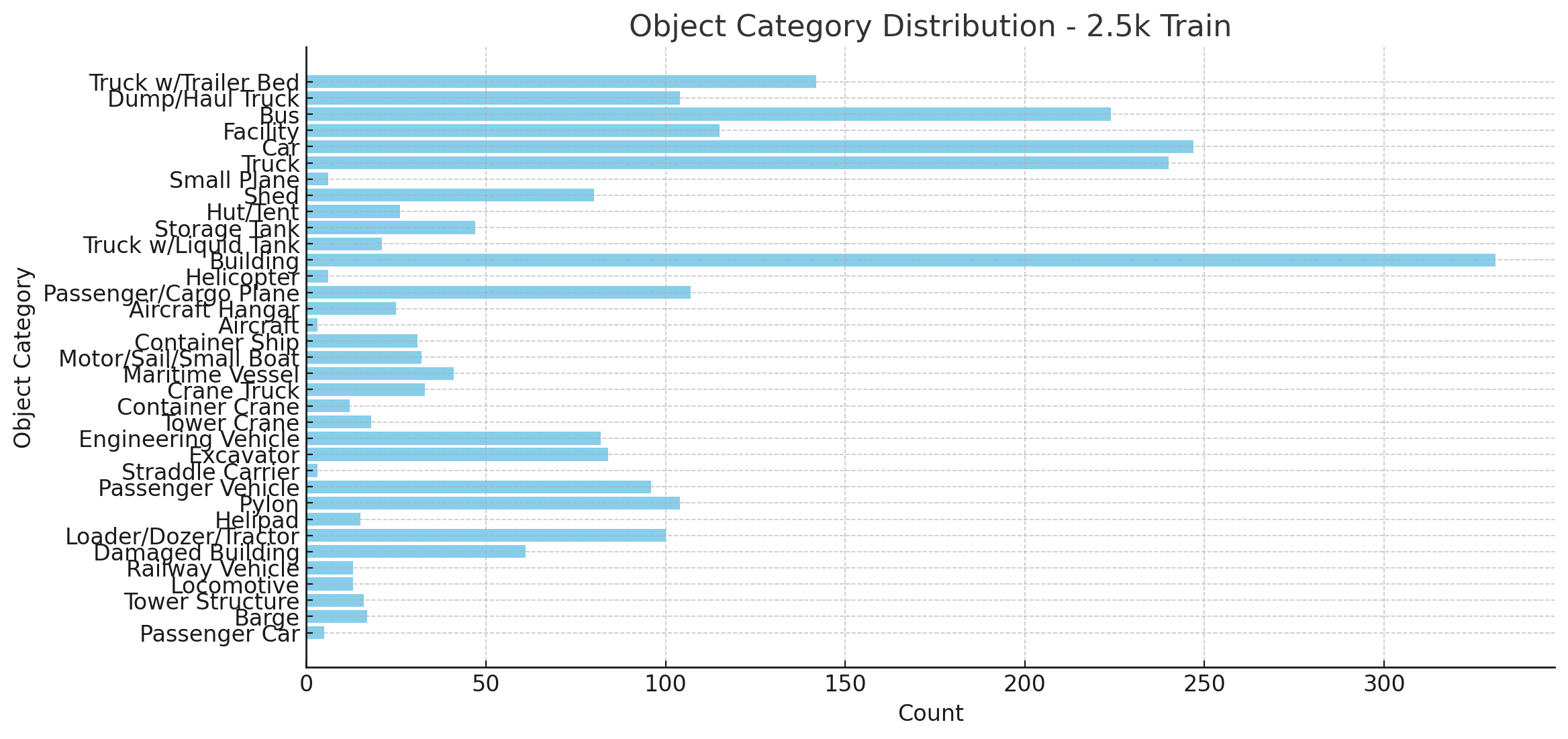}
    \caption{Class Distribution of 33\% Training Set}
    \label{fig:2_5k_train}
\end{figure*}

\begin{figure*}[h]
    \centering
    \includegraphics[width=\linewidth]{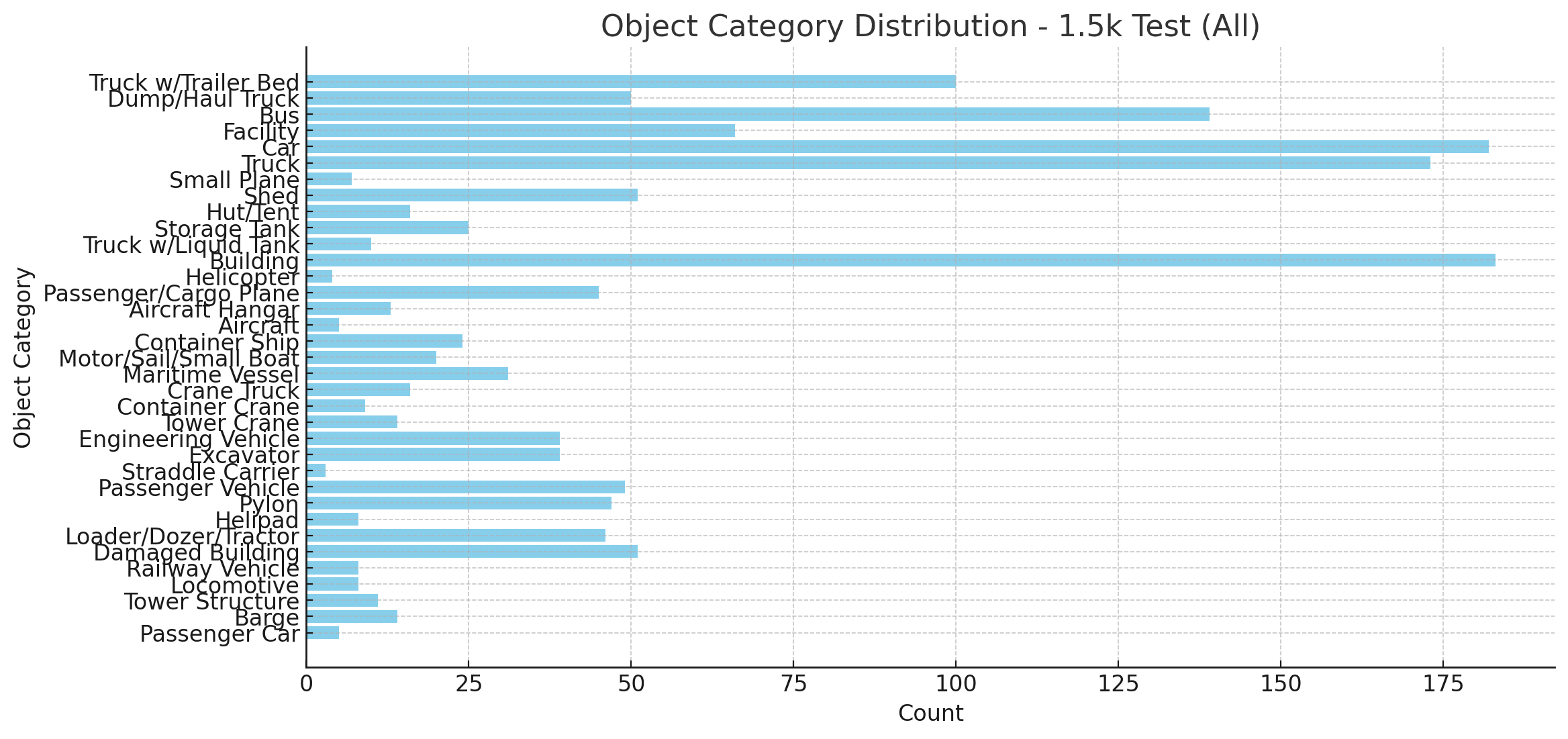}
    \caption{Class Distribution of Testing Set}
    \label{fig:1_5k_test}
\end{figure*}

Table \ref{object_category_summary} below shows the \lisat dataset distribution per class. \textcolor{black}{We have also provided bar charts for the dataset distributions in Figures \ref{fig:2_5k_train}} through \ref{fig:1_5k_test}.


\begin{table}[h]
\centering
\small
\caption{Summary of Object Categories Across Train, Validation, and Test Sets}

\begin{tabularx}{\linewidth}{lccccccc}
\toprule
\textbf{Object Category} & \textbf{2.5k Train} & \textbf{4.5k Train} & \textbf{7.2k Train} & \textbf{.5k Val} & \textbf{1.5k Test (All)} & \textbf{.5k Test (Large)} & \textbf{1k Test (Small)} \\ 
\midrule
Truck w/Trailer Bed       & 142 & 298 & 469 & 25 & 100 & 34 & 66 \\ 
Dump/Haul Truck           & 104 & 148 & 208 & 16 & 50 & 18 & 32 \\ 
Bus                       & 224 & 417 & 671 & 61 & 139 & 8 & 131 \\ 
Facility                  & 115 & 197 & 370 & 28 & 66 & 44 & 22 \\ 
Car                       & 247 & 546 & 914 & 65 & 182 & 2 & 180 \\ 
Truck                     & 240 & 518 & 932 & 75 & 173 & 12 & 161 \\ 
Small Plane               & 6 & 18 & 39 & 2 & 7 & 3 & 4 \\ 
Shed                      & 80 & 152 & 249 & 7 & 51 & 16 & 35 \\ 
Hut/Tent                  & 26 & 46 & 82 & 9 & 16 & 8 & 8 \\ 
Storage Tank              & 47 & 74 & 120 & 8 & 25 & 14 & 11 \\ 
Truck w/Liquid Tank       & 21 & 29 & 45 & 3 & 10 & 4 & 6 \\ 
Building                  & 331 & 548 & 937 & 69 & 183 & 102 & 81 \\ 
Helicopter                & 6 & 12 & 19 & 2 & 4 & 1 & 3 \\ 
Passenger/Cargo Plane     & 107 & 135 & 198 & 11 & 45 & 25 & 20 \\ 
Aircraft Hangar           & 25 & 39 & 73 & 6 & 13 & 9 & 4 \\ 
Aircraft                  & 3 & 15 & 29 & 0 & 5 & 3 & 2 \\ 
Container Ship            & 31 & 72 & 102 & 5 & 24 & 11 & 13 \\ 
Motor/Sail/Small Boat     & 32 & 58 & 87 & 7 & 20 & 2 & 18 \\ 
Maritime Vessel           & 41 & 92 & 134 & 12 & 31 & 21 & 10 \\ 
Crane Truck               & 33 & 48 & 70 & 2 & 16 & 5 & 11 \\ 
Container Crane           & 12 & 25 & 38 & 4 & 9 & 0 & 9 \\ 
Tower Crane               & 18 & 42 & 57 & 6 & 14 & 7 & 7 \\ 
Engineering Vehicle       & 82 & 115 & 166 & 15 & 39 & 11 & 28 \\ 
Excavator                 & 84 & 115 & 161 & 12 & 39 & 10 & 29 \\ 
Straddle Carrier          & 3 & 7 & 14 & 2 & 3 & 2 & 1 \\ 
Passenger Vehicle         & 96 & 145 & 215 & 15 & 49 & 0 & 49 \\ 
Pylon                     & 104 & 140 & 177 & 6 & 47 & 34 & 13 \\ 
Helipad                   & 15 & 21 & 32 & 2 & 8 & 6 & 2 \\ 
Loader/Dozer/Tractor      & 100 & 137 & 186 & 7 & 46 & 7 & 39 \\ 
Damaged Building          & 61 & 151 & 226 & 8 & 51 & 37 & 14 \\ 
Railway Vehicle           & 13 & 22 & 26 & 1 & 8 & 8 & 0 \\ 
Locomotive                & 13 & 21 & 32 & 3 & 8 & 4 & 4 \\ 
Tower Structure           & 16 & 30 & 41 & 1 & 11 & 6 & 5 \\ 
Barge                     & 17 & 42 & 59 & 5 & 14 & 13 & 1 \\ 
Passenger Car             & 5 & 14 & 27 & 1 & 5 & 1 & 4 \\ 
\midrule
\textbf{Total}            & 2500 & 4489 & 7205 & 500 & 1500 & 488 & 1023 \\ 
\bottomrule
\end{tabularx}
\label{object_category_summary}
\end{table}

\begin{figure*}[h]
    \centering
    \includegraphics[width=\linewidth]{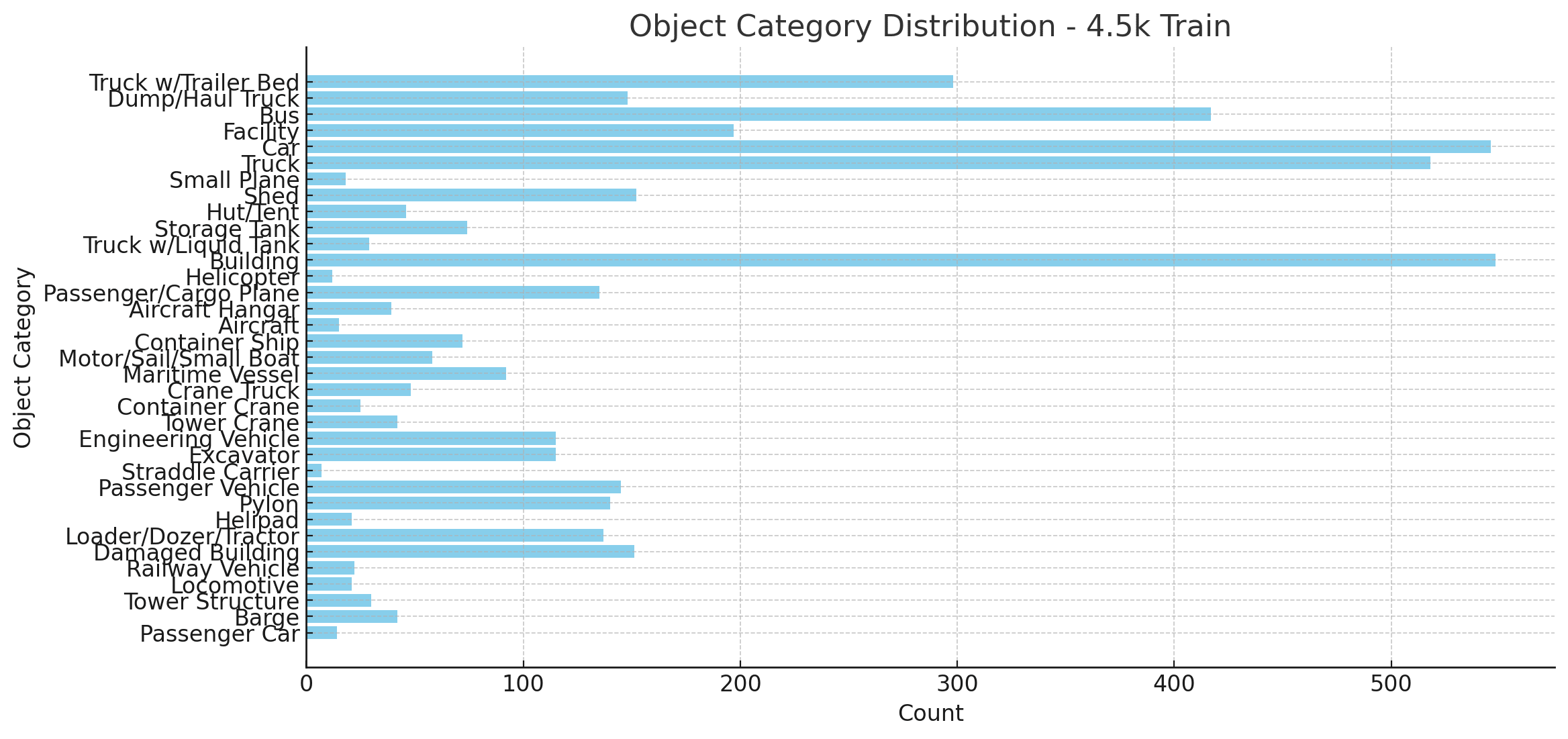}
    \caption{Class Distribution of 66\% Training Set}
    \label{fig:4_5k_train}
\end{figure*}

\begin{figure*}[h]
    \centering
    \includegraphics[width=\linewidth]{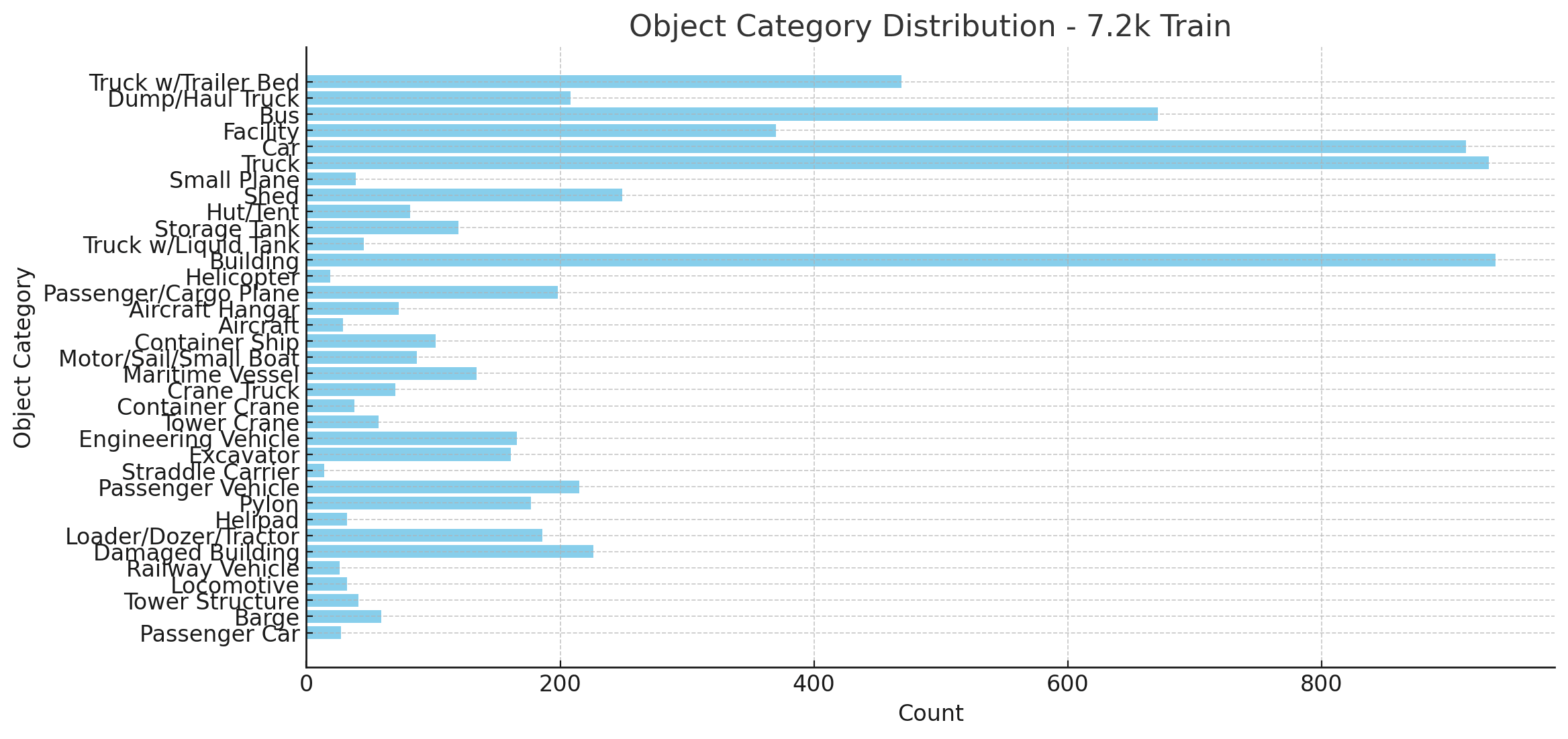}
    \caption{Class Distribution of 100\% Training Set}
    \label{fig:7_2k_train}
\end{figure*}

\begin{figure*}[h]
    \centering
    \includegraphics[width=\linewidth]{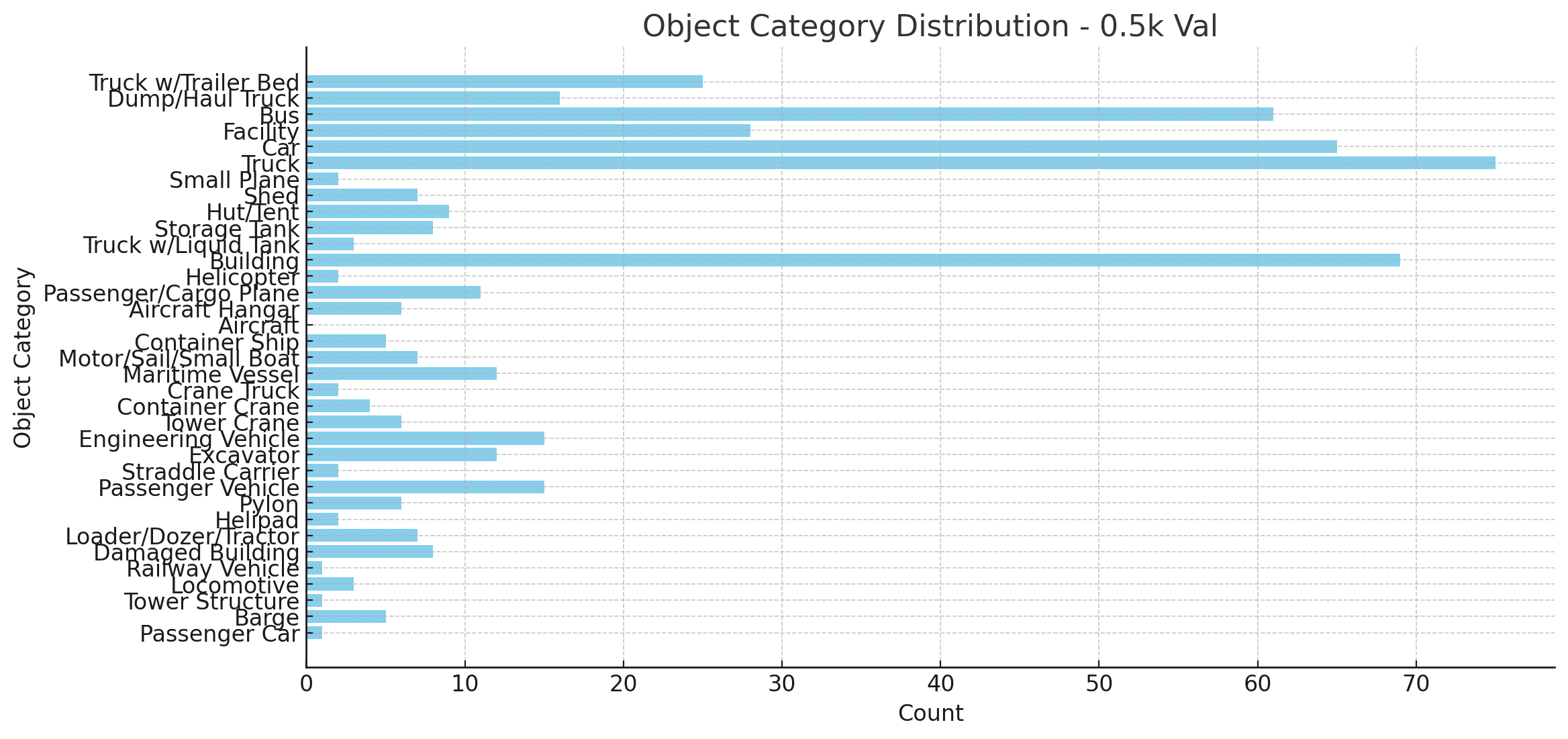}
    \caption{Class Distribution of Validation Set}
    \label{fig:0_5k_val}
\end{figure*}

\section{More on PreGRES}
\label{More_on_the_RS_LLaVA_Dataset}

We conducted additional evaluations of \lisatpre. We show evaluation results on the NWPU Caption in Table \ref{tab:nwpu},  RSICD in Table \ref{performance_comparison_RSICD}, Sidney-Caption in Table \ref{performance_comparison_sydney_captions}. We also ran Count, Presence, Comparisaon and Area evaluation as was done in \cite{xu2024rs} in Table \ref{performance_metrics_models}. 

\begin{table}[h]
    \centering
    \caption{Overview of Task Data Sources and Statistics}
    \begin{tabular}{lcccccc}
        \hline
        \textbf{Task} & \textbf{Data Source} & \textbf{Train Images} & \textbf{Train QA Pairs} & \textbf{Test Images} & \textbf{Test QA Pairs} \\
        \hline
        \multirow{5}{*}{Image Captioning} 
        & NWPU-Captions & 25200 & 125894 & 3150 & 1093 \\
        & RSICD & 8734 & 17813 & 1093 & 1093 \\
        & RSITMD & 4291 & 20096 & - & - \\
        & Sydney-Captions & 497 & 2294 & 58 & 58 \\
        & UCM-Captions & 1680 & 7999 & 210 & 210 \\
        \hline
        \multirow{4}{*}{Visual Question Answering} 
        & RSVQA-LR & 572 & 57223 & 100 & 10004 \\
        & RSVQA-HR & 6251 & 625340 & 2226 & 222684 \\
        & FloodNet & 1448 & 4511 & - & - \\
        & RSIVQA & 5401 & 19218 & - & - \\
        \hline
        Visual Grounding 
        & DIOR-RSVG & 9466 & 19643 & 7936 & 18677 \\
        \hline
        Region-level Captioning 
        & DIOR-RSVG & 9466 & 19643 & - & - \\
        \hline
        Scene Classification 
        & NWPU-RESISC45 & 31500 & 31500 & - & - \\
        \hline
        \textbf{Total} 
        & - & \textbf{104506} & \textbf{951174} & \textbf{14773} & \textbf{253819} \\
        \hline
    \end{tabular}
    \label{task_data_sources_overview}
\end{table}

\begin{table}[h]
    \centering
    \caption{Comparison of Various Models for \lisatpre on RSICD}
    \begin{tabular}{lccccccc}
        \hline
        \textbf{Method} & \textbf{BLEU-1} & \textbf{BLEU-2} & \textbf{BLEU-3} & \textbf{BLEU-4} & \textbf{METEOR} & \textbf{ROUGE L} & \textbf{CIDEr} \\
        \hline
        VLAD + RNN \cite{lu2017exploring} & 49.38 & 30.91 & 22.09 & 16.77 & 19.96 & 42.42 & 103.92 \\
        VLAD + LSTM \cite{lu2017exploring} & 50.04 & 31.95 & 23.19 & 17.78 & 20.46 & 43.34 & 118.01 \\
        mRNN \cite{qu2016deep} & 45.58 & 28.25 & 18.09 & 12.13 & 15.69 & 31.26 & 19.15 \\
        mLSTM \cite{qu2016deep} & 50.57 & 32.42 & 23.29 & 17.46 & 17.84 & 35.02 & 31.61 \\
        mGRU \cite{li2018multi} & 42.56 & 29.99 & 22.91 & 17.98 & 19.41 & 37.97 & 124.82 \\
        mGRU embedword \cite{li2018multi} & 60.94 & 46.24 & 36.80 & 29.81 & 26.14 & 48.20 & \textbf{159.54} \\
        CSMLF \cite{touvron2023llama} & 57.59 & 38.59 & 28.32 & 22.17 & 21.28 & 44.55 & 52.97 \\
        SAA \cite{lobry2020rsvqa} & 59.35 & 45.11 & 35.29 & 28.08 & 26.11 & 49.57 & 132.35 \\
        Soft-attention \cite{xu2015show} & 65.13 & 49.04 & 39.00 & 32.30 & 26.39 & 49.69 & 90.58 \\
        SD-RSIC \cite{sumbul2020sd} & 64.50 & 47.10 & 36.40 & 29.40 & 24.90 & 51.90 & 77.50 \\
        RTRMN (semantic) \cite{wang2020retrieval} & 62.01 & 46.23 & 36.44 & 29.71 & 28.29 & \textbf{55.39} & 151.46 \\
        RTRMN (statistical) \cite{wang2020retrieval} & 61.02 & 45.14 & 35.35 & 28.59 & 27.51 & 54.52 & 148.20 \\
        SVM-D BOW \cite{hoxha2021novel} & 61.12 & 42.77 & 31.53 & 24.11 & 23.03 & 45.88 & 68.25 \\
        SVM-D CONC \cite{hoxha2021novel} & 59.99 & 43.47 & 33.55 & 26.89 & 22.99 & 45.57 & 68.54 \\
        MLAT \cite{liu2022remote} & 66.90 & 51.13 & 41.14 & 34.21 & 27.31 & 50.57 & 94.27 \\
        Post-processing \cite{huang2023language} & 62.90 & 45.99 & 35.68 & 28.68 & 25.30 & 47.34 & 75.56 \\
        RS-GPT4V \cite{xu2024rs} & 70.32 & 54.23 & \textbf{44.02} & \textbf{36.83} & 30.10 & 53.34 & 102.94 \\

        LLaVA-v1.5-7b \cite{liu2024visual}  & 38.36 & 18.27 & 8.46  & 3.57  & 14.64 & 27.36 & 16.96 \\
        LLaVA-v1.6-7b \cite{liu2024llava}  & 29.31 & 13.40 & 6.00  & 2.44  & 13.11 & 24.40 & 10.69 \\
        
        \lisatpre (Ours) & \textbf{72.51} & \textbf{54.98} & 43.77 & 36.10 & \textbf{30.28} & 53.80 & 118.39 \\
        \hline
    \end{tabular}
    \label{performance_comparison_RSICD}
\end{table}


\begin{table}[h]
    \centering
    \caption{Comparison of Various Models for \lisatpre on Sydney-Captions}
    \begin{tabular}{lccccccc}
        \hline
        \textbf{Method} & \textbf{BLEU-1} & \textbf{BLEU-2} & \textbf{BLEU-3} & \textbf{BLEU-4} & \textbf{METEOR} & \textbf{ROUGE L} & \textbf{CIDEr} \\
        \hline
        VLAD + RNN \cite{lu2017exploring} & 56.58 & 45.14 & 38.07 & 32.79 & 26.72 & 52.71 & 93.72 \\
        VLAD + LSTM \cite{lu2017exploring} & 49.13 & 34.12 & 27.60 & 23.14 & 19.30 & 42.01 & 91.64 \\
        mRNN \cite{qu2016deep} & 51.30 & 37.50 & 20.40 & 19.30 & 18.50 & - & 161.00 \\
        mLSTM\cite{qu2016deep} & 54.60 & 39.50 & 22.30 & 21.20 & 20.50 & - & 186.00 \\
        mGRU \cite{li2018multi} & 69.64 & 60.92 & 52.39 & 44.21 & 31.12 & 59.17 & 171.55 \\
        mGRU embedword \cite{li2018multi} & 68.85 & 60.03 & 51.81 & 44.29 & 30.36 & 57.47 & 168.94 \\
        CSMLF \cite{touvron2023llama} & 59.98 & 45.83 & 38.69 & 34.33 & 24.75 & 50.18 & 75.55 \\
        SAA \cite{lobry2020rsvqa} & 68.82 & 60.73 & 52.94 & 45.39 & 30.49 & 58.20 & 170.52 \\
        Soft-attention \cite{xu2015show} & 73.22 & 66.74 & 62.23 & 58.20 & 39.42 & 71.27 & 249.93 \\
        Hard-attention \cite{xu2015show} & 75.91 & 66.10 & 58.89 & 52.58 & 38.98 & 71.89 & 218.19 \\
        SD-RSIC \cite{sumbul2020sd} & 72.40 & 62.10 & 53.20 & 45.10 & 34.20 & 63.60 & 139.50 \\
        SVM-D BOW \cite{hoxha2021novel} & 77.87 & 68.35 & 60.23 & 53.05 & 37.97 & 69.92 & 227.22 \\
        SVM-D CONC \cite{hoxha2021novel} & 75.47 & 67.11 & 59.70 & 53.08 & 36.43 & 67.46 & 222.22 \\
        Post-processing \cite{huang2023language} & 78.37 & 69.85 & 63.22 & 57.17 & 39.49 & 71.06 & 255.53 \\

        LLaVA-v1.5-7b \cite{liu2024visual}  & 41.04 & 19.62 & 10.80  & 4.69  & 13.71 & 31.38 & 10.89 \\
        LLaVA-v1.6-7b \cite{liu2024llava}  & 32.25 & 17.15 & 9.98  & 5.92  & 14.11 & 29.17 & 12.20 \\
        
        RS-GPT4V \cite{xu2024rs} & \textbf{82.26} & \textbf{75.28} & \textbf{68.57} & \textbf{62.23} & \textbf{41.37} & \textbf{74.77} & \textbf{273.08} \\
        
        \lisatpre (Ours) & 77.92 & 68.30 & 60.75 & 54.24 & 38.50 & 69.92 & 216.36 \\
        
        \hline
    \end{tabular}
    \label{performance_comparison_sydney_captions}
\end{table}

\begin{sidewaystable}[h]
    \centering
    \caption{Comparison of Vision and Language Encoders for \lisatpre on UCM-Captions, NWPU-Captions, RSICD, and Sydney-Captions}
    \small
    \begin{tabular}{c c cccccccc}
        \hline
        \textbf{Vision Encoder} & \textbf{Language Encoder} & \textbf{BLEU-1} & \textbf{BLEU-2} & \textbf{BLEU-3} & \textbf{BLEU-4} & \textbf{METEOR} & \textbf{ROUGE L} & \textbf{CIDEr} & \textbf{SPICE} \\
        \hline
        \multicolumn{10}{c}{\textbf{\centering UCM-Captions}} \\

        CLIP    & Llama 2  & 85.57 & 79.02 & 73.81 & 69.03 & 45.49 & 80.10 & 328.82  & 52.21 \\

        CLIP336    & Llama 2  & 84.86 & 77.81 & 72.06 & 66.97 & 44.70 & 78.97 & 324.61  & 50.46 \\
        
        SAT-CLIP    & Llama 2  & 41.24 & 32.38 & 12.74 & 8.82  & 13.90 & 28.30 & 30.41  & 8.15  \\
        
        Geo-CLIP    & Llama 2  & 44.57 & 26.22 & 17.37 & 12.77 & 15.61 & 32.22 & 44.64  & 11.67 \\
        
        RemoteCLIP  & Llama 2  & 85.95 & 79.00 & 73.38 & 68.31 & 45.80 & 79.99 & 330.94 & 52.17 \\
        
        CLIP        & Vicuna & 84.93 & 77.80 & 72.05 & 66.68 & 45.62 & 80.04 & 329.32 & 52.00 \\
        
        CLIP336     & Vicuna & 85.40 & 78.81 & 73.34 & 68.28 & 45.84 & 79.66 & 324.89 & 51.55 \\
        
        SAT-CLIP    & Vicuna & 47.21 & 29.55 & 21.21 & 16.87 & 17.35 & 34.20 & 63.92  & 15.08 \\
        
        Geo-CLIP    & Vicuna & 53.77 & 37.86 & 29.50 & 24.56 & 21.77 & 42.86 & 109.20 & 21.15 \\
        
        RemoteCLIP  & Vicuna & \textbf{88.23} & \textbf{82.07} & \textbf{77.08} & \textbf{72.34} & \textbf{47.78} & \textbf{83.13} & \textbf{355.32} & \textbf{54.15} \\
        \hline
        \multicolumn{10}{c}{\textbf{\centering NWPU-Captions}} \\

        CLIP    & Llama 2  & 87.25 & 77.53 & 69.89 & 63.53 & 43.33 & 76.59 & 180.81  & 31.38 \\

        CLIP336    & Llama 2  & 86.70 & 76.38 & 68.43 & 62.04 & 42.55 & 75.56 & 176.54  & 30.75 \\

        SAT-CLIP    & Llama 2  & 69.51 & 50.90 & 39.28 & 31.48 & 25.83 & 52.34 & 68.94  & 16.15 \\
        
        Geo-CLIP    & Llama 2 & 74.36 & 58.26 & 47.59 & 39.96 & 30.12 & 58.48 & 97.67  & 19.78 \\
        
        RemoteCLIP  & Llama 2  & 87.25 & 77.67 & 70.06 & 63.76 & 43.44 & 76.64 & 181.01 & 31.48 \\
        
        CLIP        & Vicuna & 86.62 & 76.76 & 69.03 & 62.59 & 43.09 & 76.18 & 179.96 & 31.09 \\
        
        CLIP336     & Vicuna & 87.47 & 77.79 & 70.13 & 63.78 & 43.47 & 76.42 & 181.94 & 31.15 \\
        
        SAT-CLIP    & Vicuna & 75.87 & 60.51 & 50.35 & 43.00 & 31.78 & 60.51 & 105.53 & 21.25 \\
        
        Geo-CLIP    & Vicuna & 77.98 & 63.94 & 54.24 & 46.96 & 34.10 & 63.87 & 121.54 & 23.44 \\
        
        RemoteCLIP  & Vicuna & \textbf{88.5} & \textbf{79.3} & \textbf{72.0} & \textbf{65.8} & \textbf{44.4} & \textbf{77.5} & \textbf{185.7} & \textbf{32.2} \\
        \hline
        \multicolumn{10}{c}{\textbf{\centering RSICD}} \\
        CLIP    & Llama 2  & 60.51 & 43.02 & 32.34 & 25.60 & 25.41 & 46.20 & 76.30 & 25.46 \\

        CLIP336    & Llama 2  & 70.05 & 51.72 & 40.06 & 32.36 & 28.19 & 50.45 & 107.18  & 28.35 \\
        
        Geo-CLIP   & Llama 2  & 45.86 & 21.97 & 11.90 & 7.16  & 15.73 & 28.46 & 22.60  & 12.88 \\
        
        RemoteCLIP & Llama 2  & 68.19 & 49.65 & 38.25 & 30.74 & 27.64 & 49.91 & 101.05 & 27.94 \\
        
        SAT-CLIP   & Llama 2  & 43.68 & 19.17 & 9.82  & 5.60  & 14.61 & 26.65 & 16.71  & 10.95 \\
        
        CLIP       & Vicuna & 62.46 & 44.31 & 33.36 & 26.47 & 25.65 & 47.08 & 81.22  & 25.56 \\
        CLIP336    & Vicuna & 70.40 & 52.09 & 40.44 & 32.79 & 28.50 & 50.74 & 108.37 & 28.82 \\
        SAT-CLIP   & Vicuna & 45.95 & 22.68 & 13.00 & 8.23  & 16.02 & 29.02 & 24.77  & 12.76 \\
        Geo-CLIP   & Vicuna & 49.19 & 26.41 & 16.17 & 10.80 & 17.67 & 31.74 & 31.44  & 15.14 \\
        RemoteCLIP & Vicuna & \textbf{72.51} & \textbf{54.98} & \textbf{43.77} & \textbf{36.10} & \textbf{30.28} & \textbf{53.80} & \textbf{118.39} & \textbf{30.54} \\
        \hline
        \multicolumn{10}{c}{\textbf{\centering Sydney-Captions}} \\

        CLIP    & Llama 2  & 78.59 & 69.78 & 62.50 & 56.35 & 39.09 & 70.00 & 220.50 & 45.20 \\

        CLIP336    & Llama 2  & 78.48 & 69.35 & 62.25 & 56.14 & 38.54 & 68.98 & 211.95  & 43.25 \\

        SAT-CLIP   & Llama 2  & 58.40 & 45.47 & 37.93 & 32.27 & 25.75 & 46.95 & 85.67  & 22.96 \\
        
        Geo-CLIP   & Llama 2  & 68.91 & 56.44 & 48.91 & 43.09 & 30.75 & 55.86 & 153.84 & 31.33 \\
        
        RemoteCLIP & Llama 2  & 76.19 & 66.31 & 58.57 & 52.27 & 37.30 & 68.12 & 201.64 & 43.99 \\

        CLIP       & Vicuna & 76.42 & 67.79 & 60.49 & 53.95 & 38.02 & 68.78 & 204.76 & \textbf{44.07} \\
        CLIP336    & Vicuna & 77.68 & \textbf{68.70} & 61.37 & \textbf{55.35} & 38.41 & \textbf{70.37} & 213.92 & 45.22 \\
        SAT-CLIP   & Vicuna & 69.58 & 58.86 & 51.80 & 46.45 & 32.24 & 58.43 & 182.93 & 34.65 \\
        Geo-CLIP   & Vicuna & 71.68 & 60.83 & 53.67 & 47.76 & 34.33 & 61.49 & 177.19 & 36.60 \\
        RemoteCLIP & Vicuna & \textbf{77.92} & 68.30 & \textbf{60.75} & 54.24 & \textbf{38.50} & 69.92 & \textbf{216.36} & 44.04 \\
        \hline
    \end{tabular}
    \label{combined_vision_language_encoders}
\end{sidewaystable}


\begin{table}[h]
\centering
\caption{Performance Metrics for \lisatpre on the RSVQA\_LR}
\begin{tabular}{@{}lcccc@{}}
\toprule
\textbf{Model}   & \textbf{Count} & \textbf{Presence} & \textbf{Comparison} & \textbf{Area} \\ \midrule
RSVQA \cite{lobry2020rsvqa}                    & 67.01          & 87.46             & 81.50               & 85.24         \\
EasyToHard \cite{yuan2022easy}               & 69.22          & 90.66             & 87.49               & 85.92         \\
Bi-Modal \cite{bazi2022bi}                 & \textbf{72.22} & 91.06             & 91.16               & 86.27         \\
SHRNet \cite{zhang2023spatial}                   & 73.87          & 91.03             & 90.48               & \textbf{86.35}         \\
LLaVA-1.5 \cite{liu2024visual}                & 26.81          & 54.72             & 66.22               & 1.45          \\
InternLM-XC2 \cite{internlmxcomposer2}            & 26.91          & 55.74             & 64.89               & 5.94          \\
RS-GPT4V \cite{xu2024rs}                    & -              & 91.17             & 91.70               & -             \\
GeoChat \cite{zhang2023multi}                  & -              & 91.09             & 90.33               & -             \\ 
Full-FT \cite{xu2024rs}                   & 70.48          & 91.10             & 92.23               & 86.00         \\
LoRA \cite{xu2024rs}                     & 70.34          & 92.24             & 92.10               & 85.84         \\
MoE LoRA \cite{xu2024rs}                 & 71.06          & 91.10             & \textbf{92.55}      & 85.82         \\

LLaVA-v1.5-7b \cite{liu2024visual}  & 18.66 & 53.98 & 66.22  & 58.00  \\
LLaVA-v1.6-7b \cite{liu2024llava}  & 19.65 & 57.53 & 62.32  & 62.00   \\

\lisatpre (Ours) & 70.24       & \textbf{92.36}    &  92.20             & 61.43         \\ \bottomrule
\end{tabular}
\label{performance_metrics_models}
\end{table}


\section{Qualitative Analysis} 

\label{Qualitative_Analysis}

In this section, we present a qualitative analysis of the model's performance, showcasing a range of success cases \autoref{Successfull_Results_of_LISAt}, failure cases \autoref{Failure_Cases_of_LISAt}, and instances where the ground truth (GT) was erroneous \autoref{Ground_Truth_Error_Cases}. Success cases shown in \autoref{prediction_comparison_1}, \autoref{prediction_comparison_2}, \autoref{prediction_comparison_3}, \autoref{prediction_comparison_4}, and \autoref{prediction_comparison_5} highlight scenarios where the model successfully aligns with the expected outcomes, demonstrating its ability to handle complex tasks accurately. Failure cases shown in \autoref{failure_prediction_comparison}, however, indicate situations where the model struggles due to challenges such as occlusion, poor lighting, or ambiguous object representations, leading to incorrect predictions or missed detections. These cases reveal areas where model improvements are needed, particularly in dynamic environments or with less structured input data. Finally, GT mistake cases as shown in \autoref{gt_mistake_prediction_comparison} refer to instances where the GT was erroneous but the model aligns with the expected ground truth annotations. The model is penalized here due to inherent inconsistencies in the dataset from the mask labeling with GeoSAM. These cases underscore the challenges posed by noisy or ambiguous ground truth data, highlighting the importance of dataset refinement and improved model calibration to reduce such errors. Together, these cases provide valuable insights into the model's performance, guiding future research and optimizations.


\subsection{Success Cases of \lisat}
\label{Successfull_Results_of_LISAt}

In this subsection, we present a selection of successful cases where \lisat accurately predicted object categories and configurations. These examples highlight the model's ability to generalize and perform well under varied conditions, demonstrating its effectiveness in real-world applications.

\begin{table}[h]
\centering
\caption{Comparison of Predictions and Ground Truth Across Models}
\resizebox{\textwidth}{!}{
\begin{tabular}{>{\centering\arraybackslash}p{0.18\textwidth}m{0.22\textwidth}m{0.22\textwidth}m{0.22\textwidth}m{0.22\textwidth}}
\hline
\multicolumn{1}{c}{\textbf{Queries}} & 
\multicolumn{1}{c}{\textbf{RGB}} & 
\multicolumn{1}{c}{\textbf{LISA}} & 
\multicolumn{1}{c}{\textbf{\lisat (Ours)}} & 
\multicolumn{1}{c}{\textbf{Ground Truth}} \\
\hline
\\
\centering\arraybackslash\parbox{0.18\textwidth}{\centering\vspace{-9pt}Identify the excavator by locating the bright yellow arm and bucket against the darker background.
} & 
\includegraphics[width=1.05\linewidth]{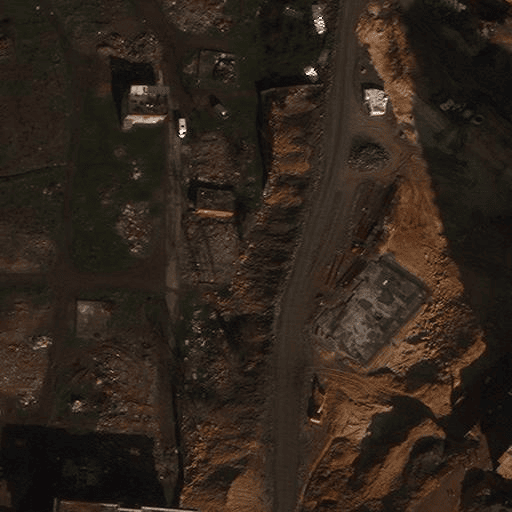} & 
\includegraphics[width=1.05\linewidth]{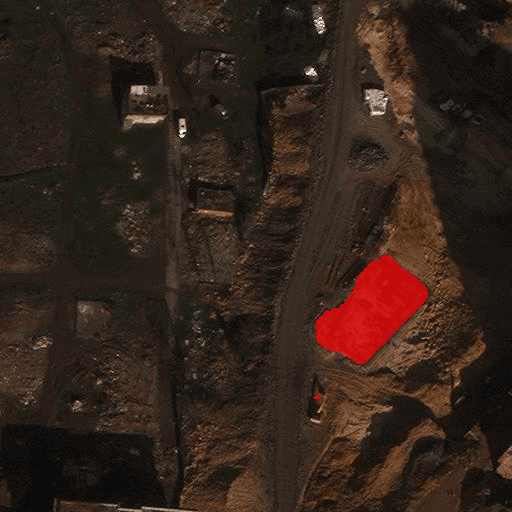} & 
\includegraphics[width=1.05\linewidth]{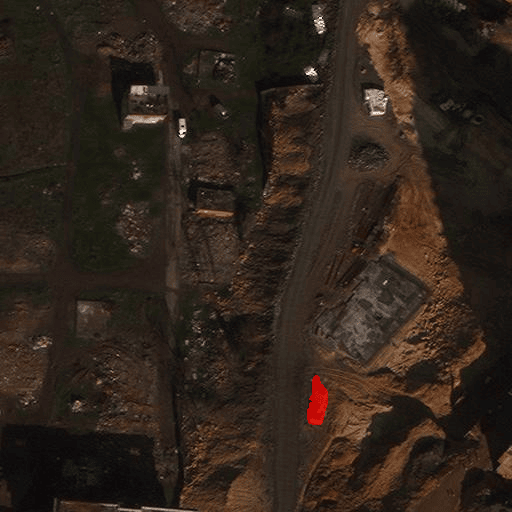} & 
\includegraphics[width=1.05\linewidth]{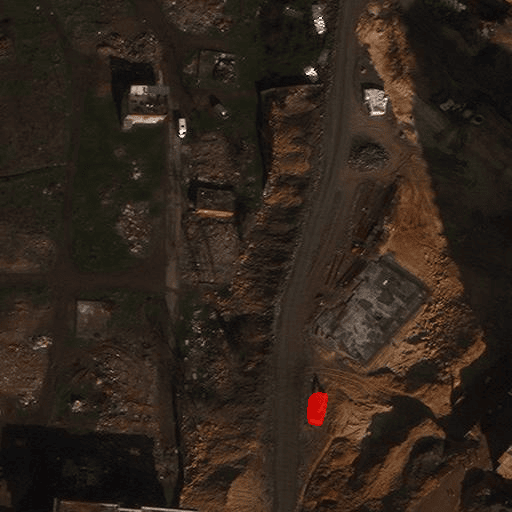} \\

\centering\arraybackslash\parbox{0.18\textwidth}{\centering\vspace{-9pt}Locate the building with a beige facade and a dark brown roof in the image.
} & 
\includegraphics[width=1.05\linewidth]{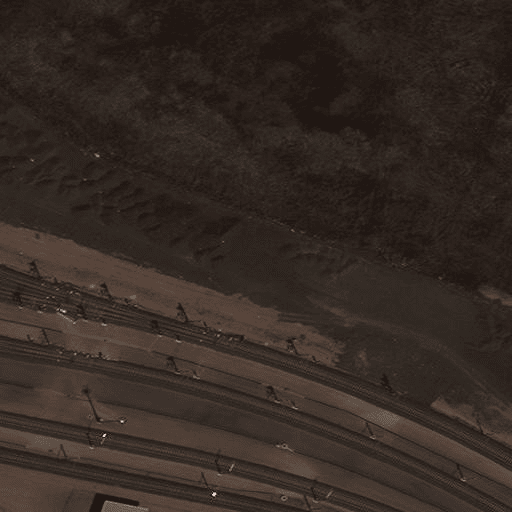} & 
\includegraphics[width=1.05\linewidth]{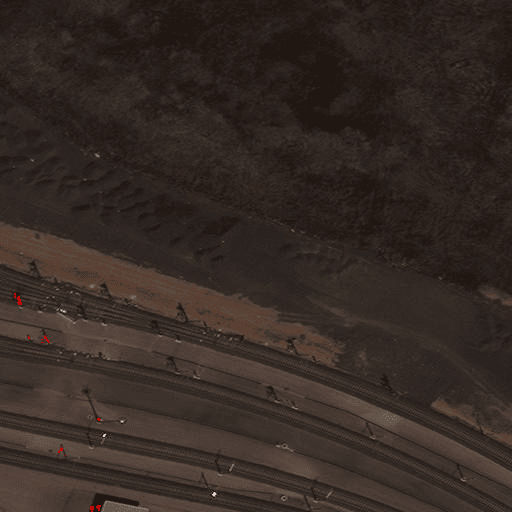} & 
\includegraphics[width=1.05\linewidth]{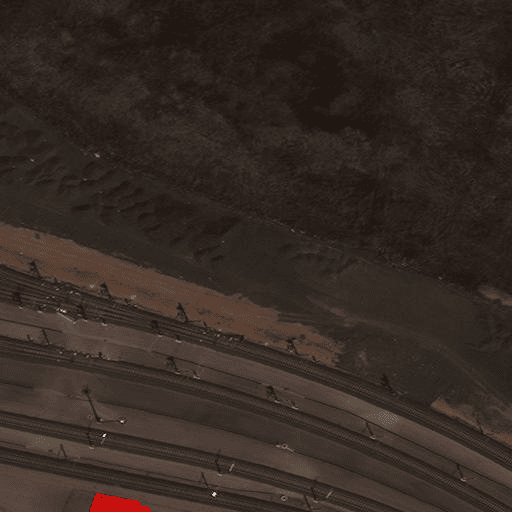} & 
\includegraphics[width=1.05\linewidth]{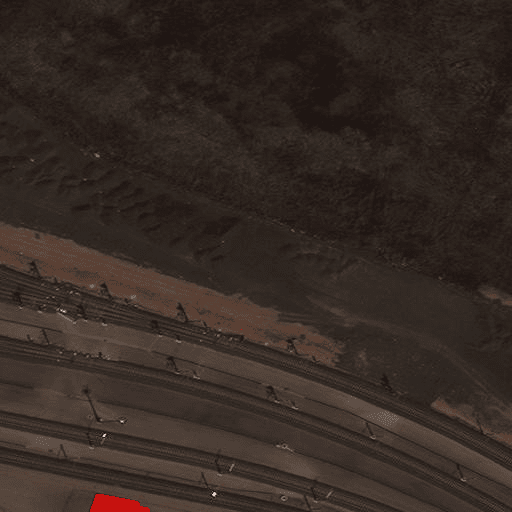} \\

\centering\arraybackslash\parbox{0.18\textwidth}{\centering\vspace{-9pt}Locate the large, elongated structure with stacked rectangular containers and a reddish-brown deck, characteristic of a container ship, against the dark water background.
} & 
\includegraphics[width=1.05\linewidth]{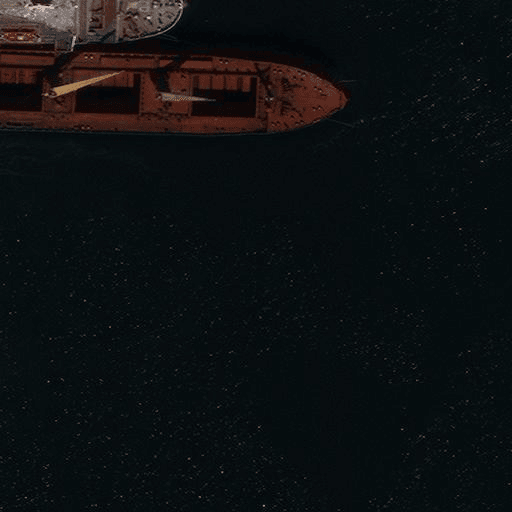} & 
\includegraphics[width=1.05\linewidth]{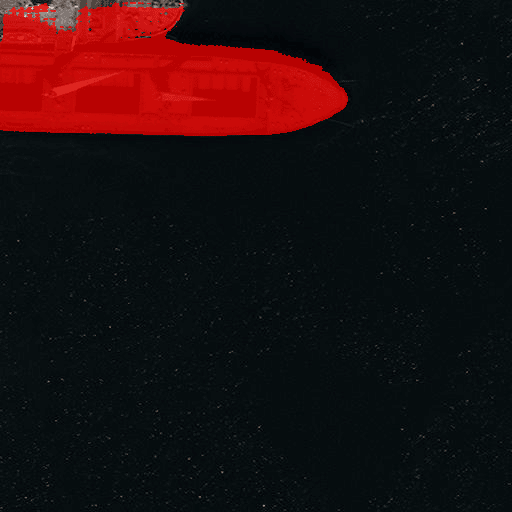} & 
\includegraphics[width=1.05\linewidth]{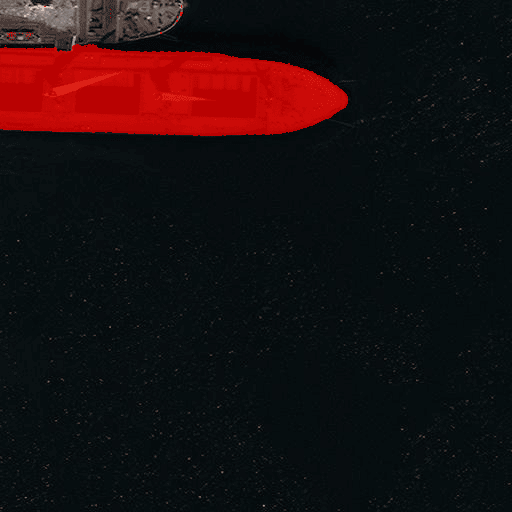} & 
\includegraphics[width=1.05\linewidth]{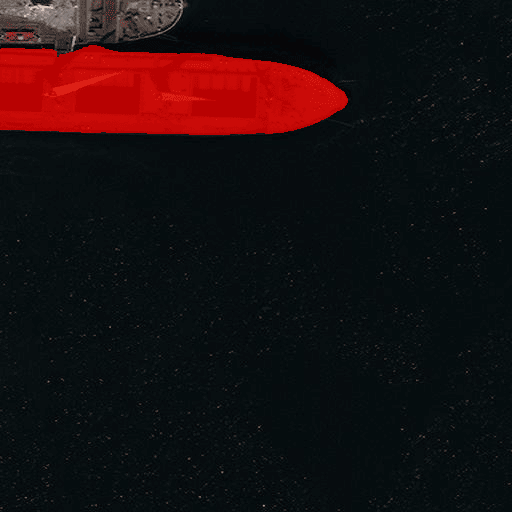} \\

\centering\arraybackslash\parbox{0.18\textwidth}{\centering\vspace{-9pt}Locate the building in the center-left of the image.
} & 
\includegraphics[width=1.05\linewidth]{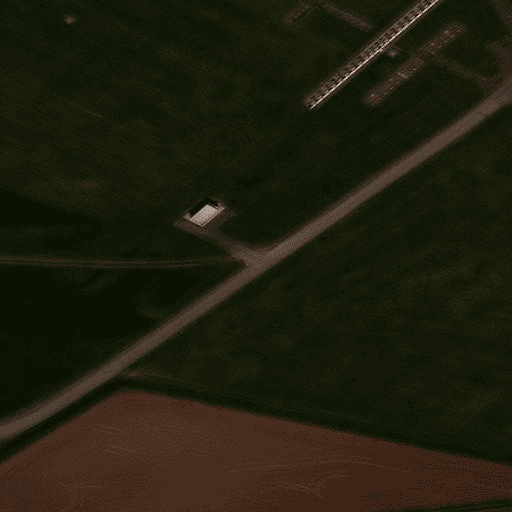} & 
\includegraphics[width=1.05\linewidth]{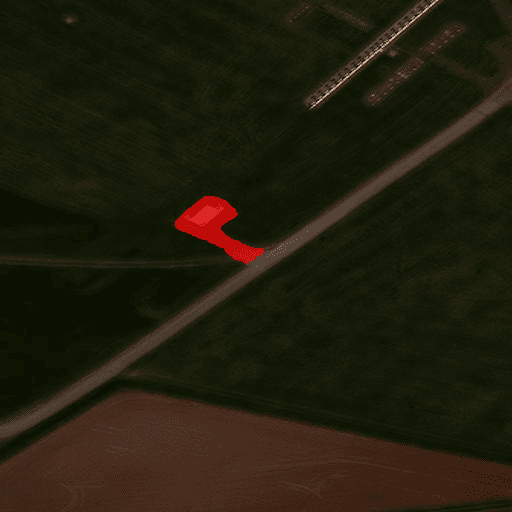} & 
\includegraphics[width=1.05\linewidth]{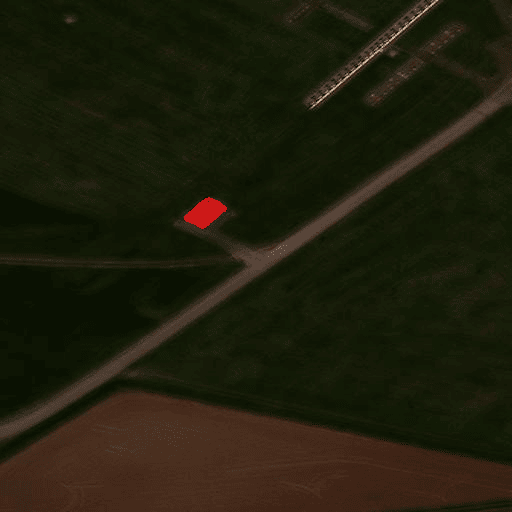} & 
\includegraphics[width=1.05\linewidth]{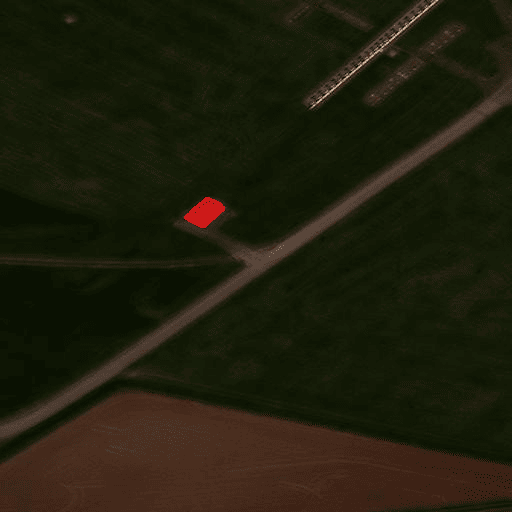} \\

\centering\arraybackslash\parbox{0.18\textwidth}{\centering\vspace{-9pt}Locate the long, green vehicle with rectangular windows and wheels, positioned horizontally across the image.
} & 
\includegraphics[width=1.05\linewidth]{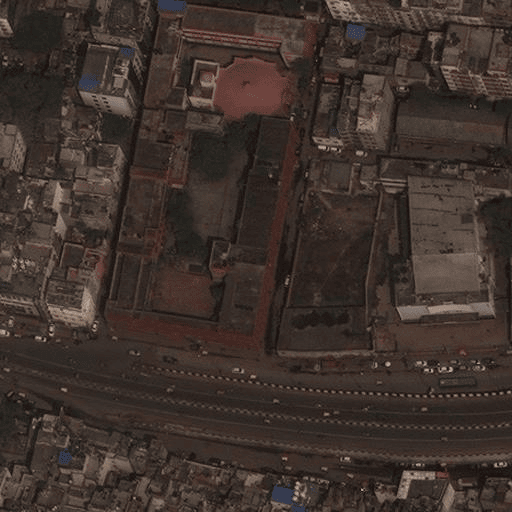} & 
\includegraphics[width=1.05\linewidth]{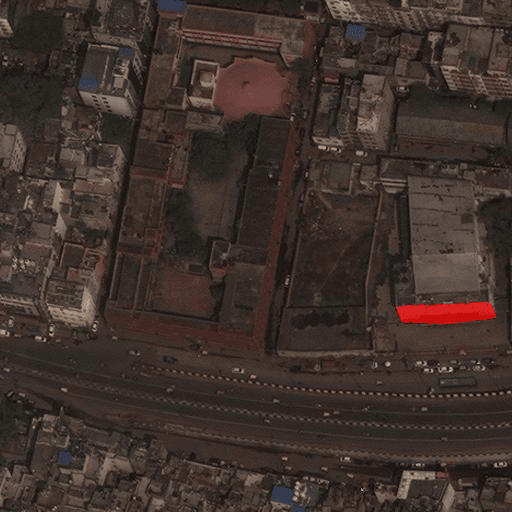} & 
\includegraphics[width=1.05\linewidth]{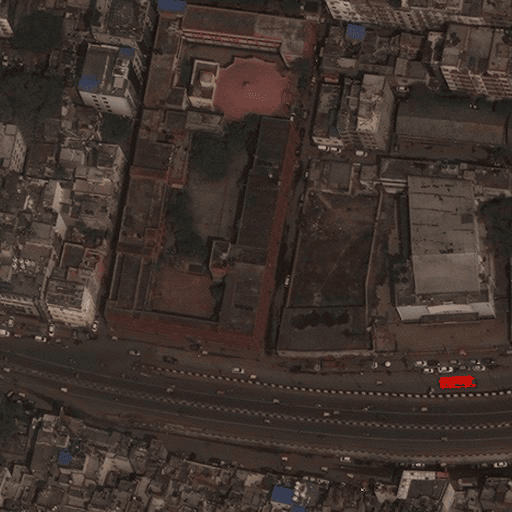} & 
\includegraphics[width=1.05\linewidth]{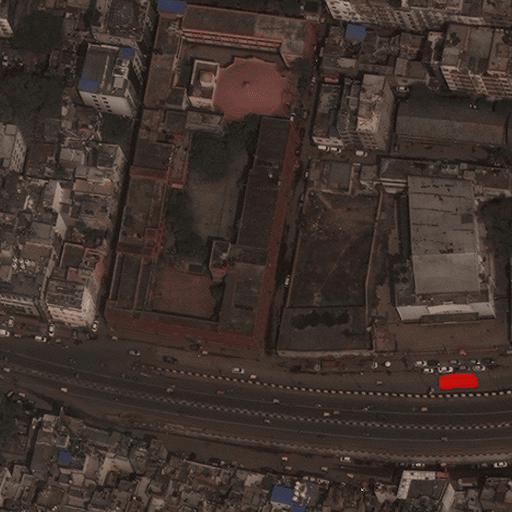} \\

\centering\arraybackslash\parbox{0.18\textwidth}{\centering\vspace{-9pt}Locate the building in the top-left of the image.
} & 
\includegraphics[width=1.05\linewidth]{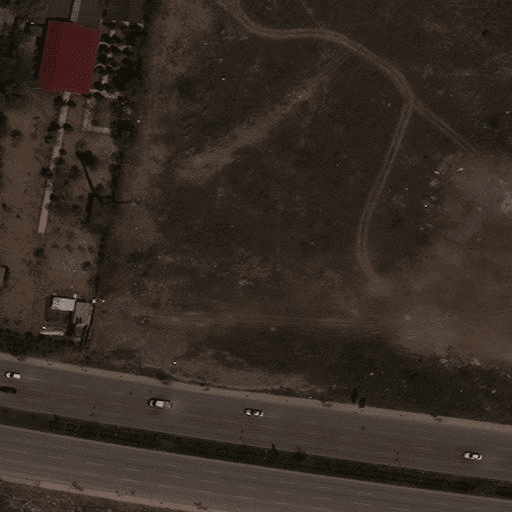} & 
\includegraphics[width=1.05\linewidth]{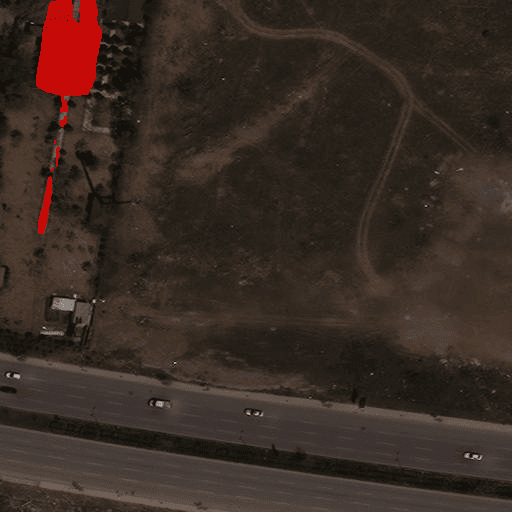} & 
\includegraphics[width=1.05\linewidth]{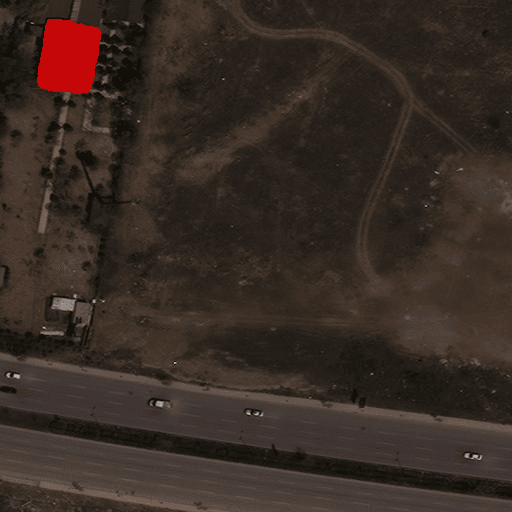} & 
\includegraphics[width=1.05\linewidth]{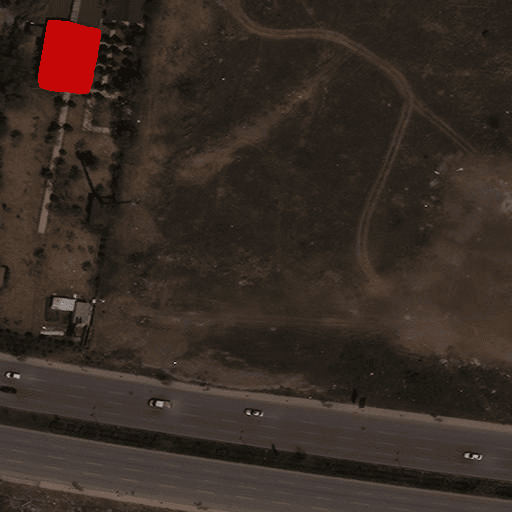} \\

\hline
\end{tabular}
}
\label{prediction_comparison_1}
\end{table}


\begin{table}[h]
\centering
\caption{Comparison of Predictions and Ground Truth Across Models (Cont.)}
\resizebox{\textwidth}{!}{
\begin{tabular}{>{\centering\arraybackslash}p{0.18\textwidth}m{0.22\textwidth}m{0.22\textwidth}m{0.22\textwidth}m{0.22\textwidth}}
\hline
\multicolumn{1}{c}{\textbf{Queries}} & 
\multicolumn{1}{c}{\textbf{RGB}} & 
\multicolumn{1}{c}{\textbf{LISA}} & 
\multicolumn{1}{c}{\textbf{\lisat (Ours)}} & 
\multicolumn{1}{c}{\textbf{Ground Truth}} \\
\hline
\\

\centering\arraybackslash\parbox{0.18\textwidth}{\centering\vspace{-9pt}Identify the triangular metal structure with intersecting lines, standing vertically in the image.
} & 
\includegraphics[width=1.05\linewidth]{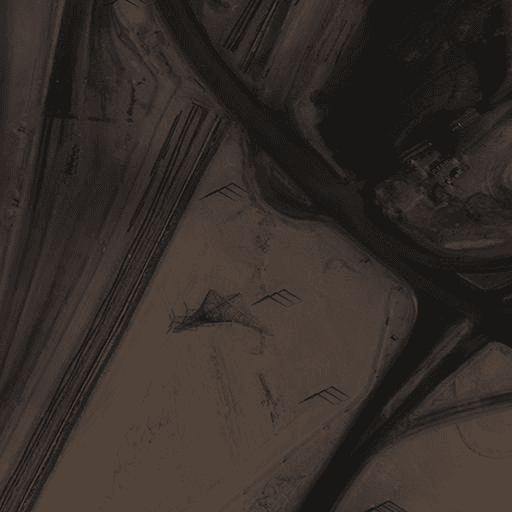} & 
\includegraphics[width=1.05\linewidth]{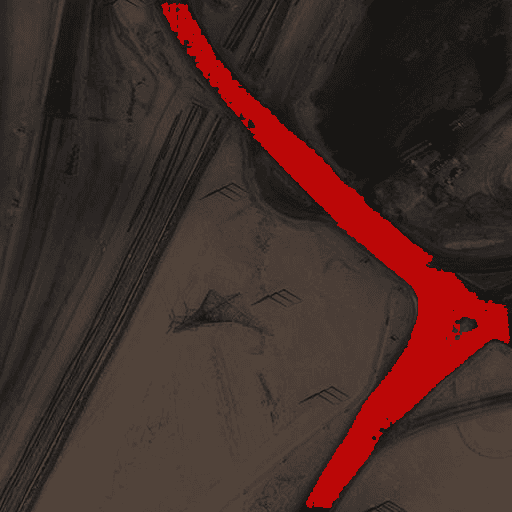} & 
\includegraphics[width=1.05\linewidth]{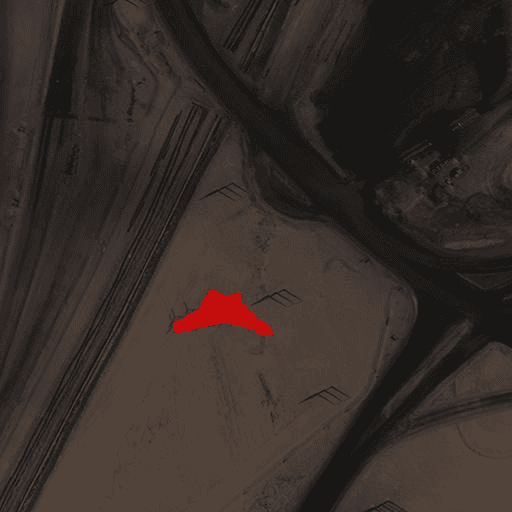} & 
\includegraphics[width=1.05\linewidth]{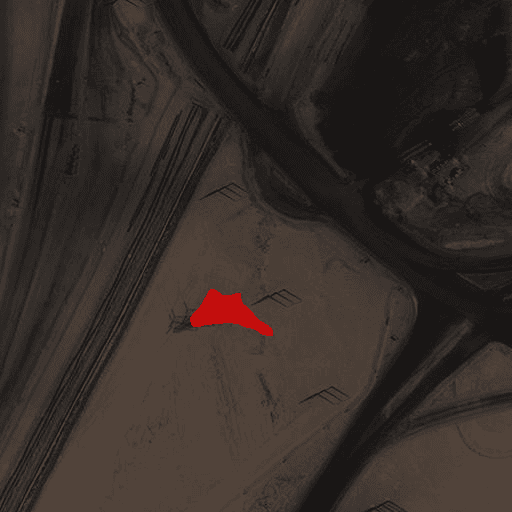} \\

\centering\arraybackslash\parbox{0.18\textwidth}{\centering\vspace{-9pt}Identify the circular structure with a metallic appearance and distinct shadow, contrasting against the surrounding terrain.
} & 
\includegraphics[width=1.05\linewidth]{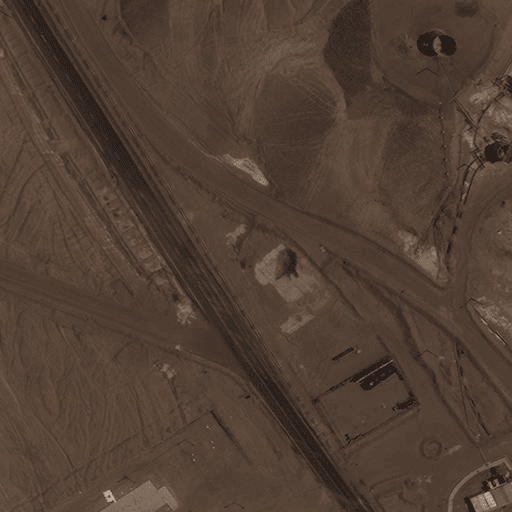} & 
\includegraphics[width=1.05\linewidth]{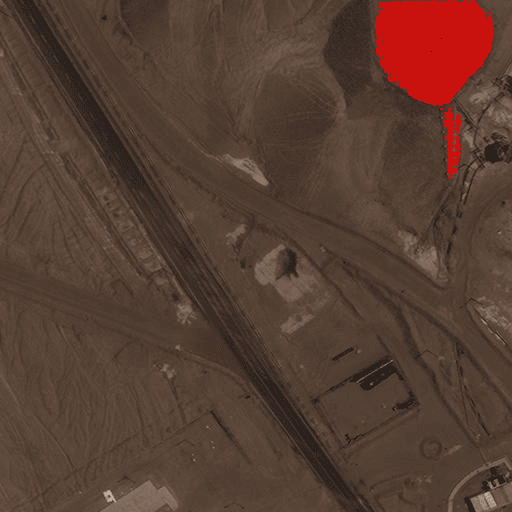} & 
\includegraphics[width=1.05\linewidth]{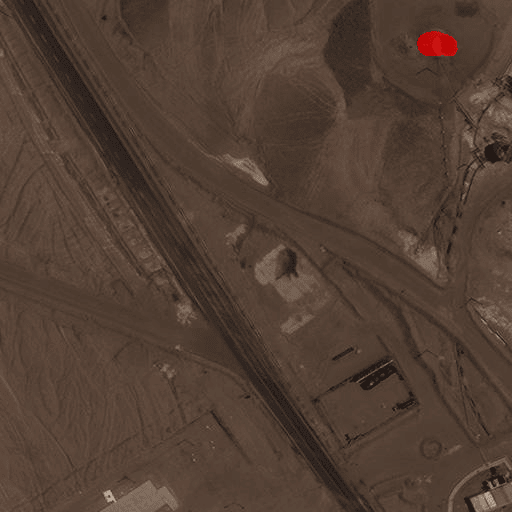} & 
\includegraphics[width=1.05\linewidth]{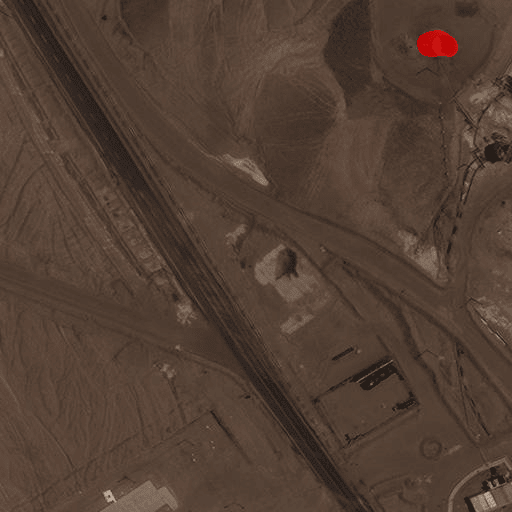} \\

\centering\arraybackslash\parbox{0.18\textwidth}{\centering\vspace{-9pt}Identify the pylon in the top-left area of the image.
} & 
\includegraphics[width=1.05\linewidth]{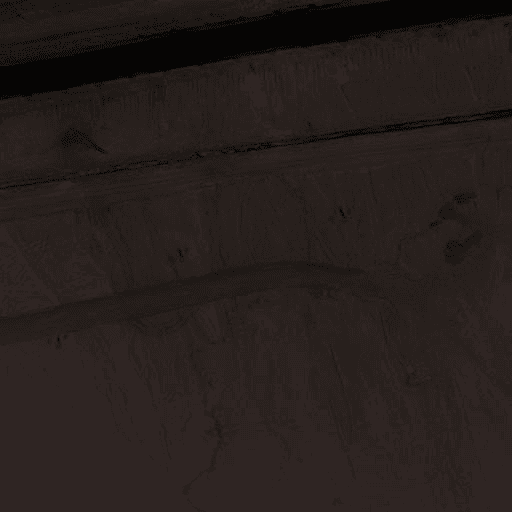} & 
\includegraphics[width=1.05\linewidth]{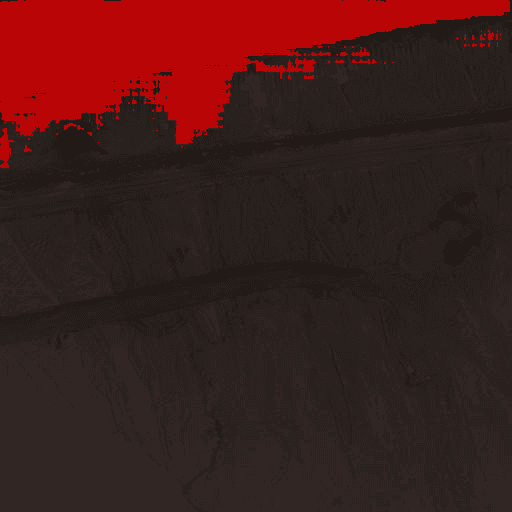} & 
\includegraphics[width=1.05\linewidth]{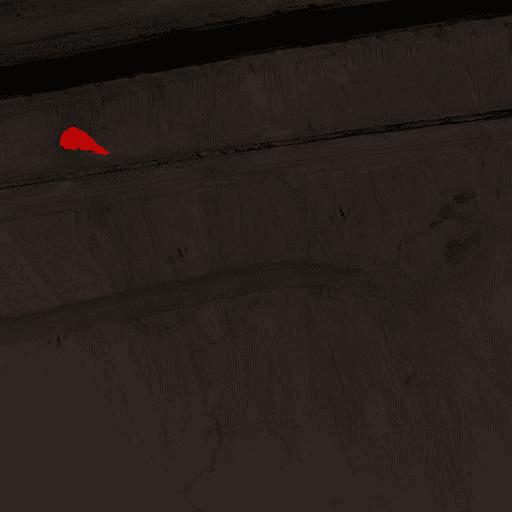} & 
\includegraphics[width=1.05\linewidth]{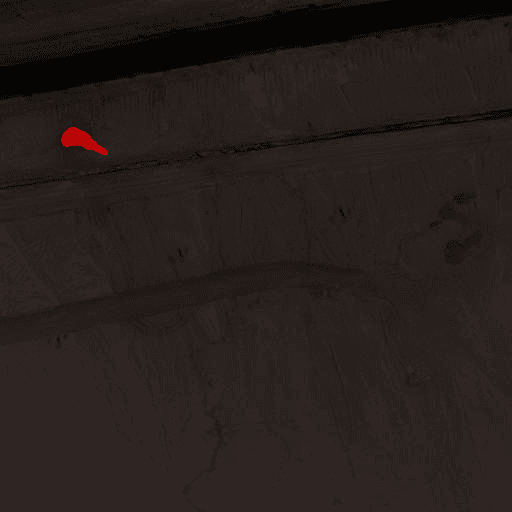} \\

\centering\arraybackslash\parbox{0.18\textwidth}{\centering\vspace{-9pt}Identify the pylon located in the bottom-right of the image.
} & 
\includegraphics[width=1.05\linewidth]{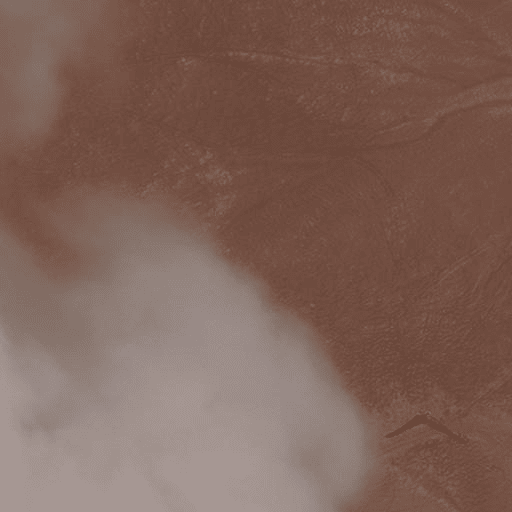} & 
\includegraphics[width=1.05\linewidth]{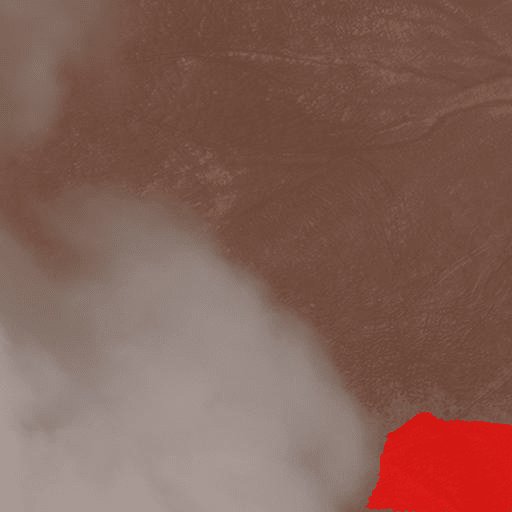} & 
\includegraphics[width=1.05\linewidth]{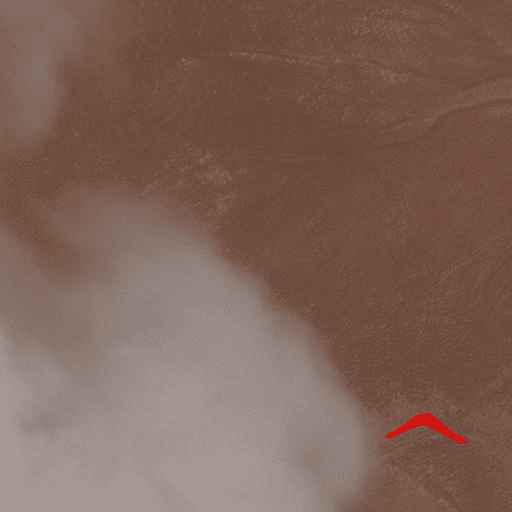} & 
\includegraphics[width=1.05\linewidth]{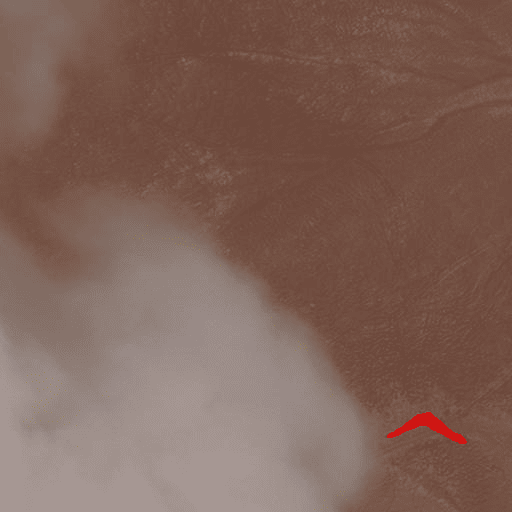} \\

\centering\arraybackslash\parbox{0.18\textwidth}{\centering\vspace{-9pt}Identify the engineering vehicle with a metallic appearance and distinct geometric shapes against the brown background.
} & 
\includegraphics[width=1.05\linewidth]{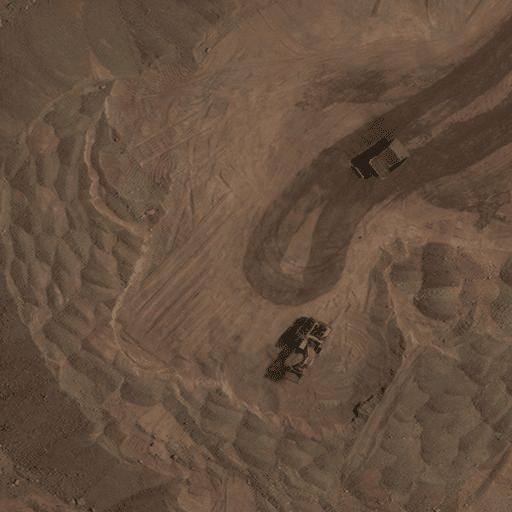} & 
\includegraphics[width=1.05\linewidth]{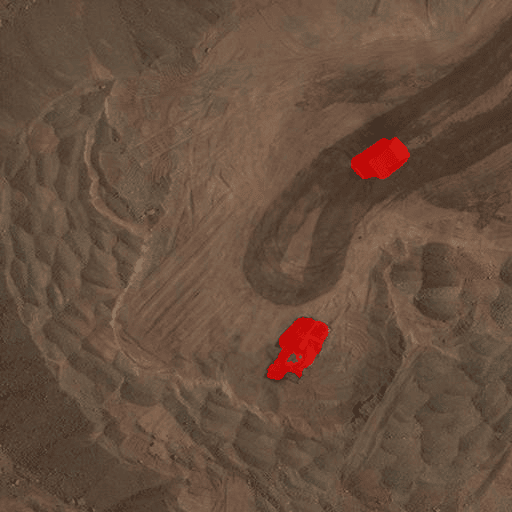} & 
\includegraphics[width=1.05\linewidth]{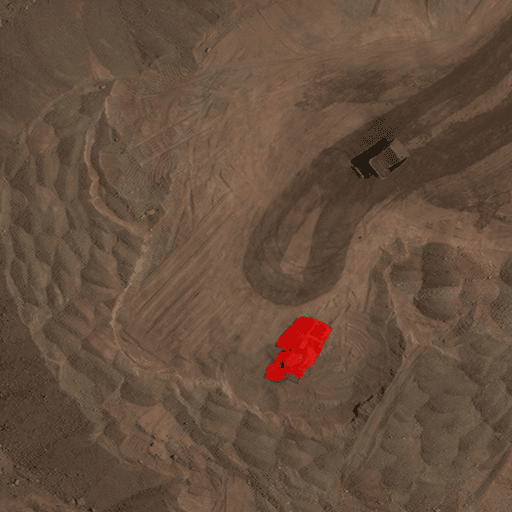} & 
\includegraphics[width=1.05\linewidth]{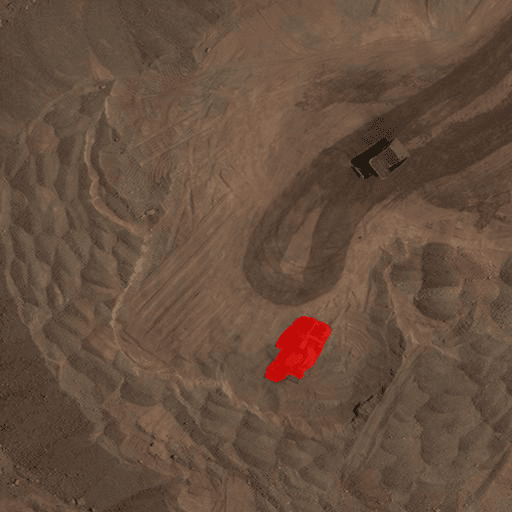} \\

\centering\arraybackslash\parbox{0.18\textwidth}{\centering\vspace{-9pt}Identify the damaged building with an irregular, fragmented roof structure and scattered debris contrasting with surrounding vegetation.
} & 
\includegraphics[width=1.05\linewidth]{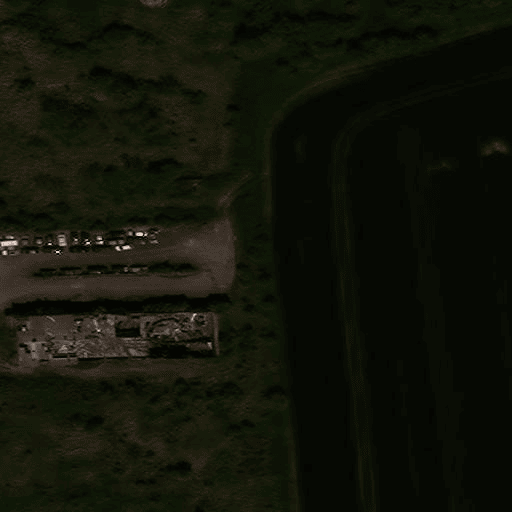} & 
\includegraphics[width=1.05\linewidth]{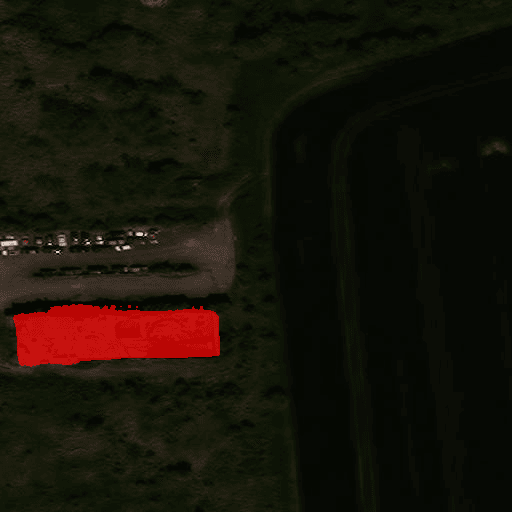} & 
\includegraphics[width=1.05\linewidth]{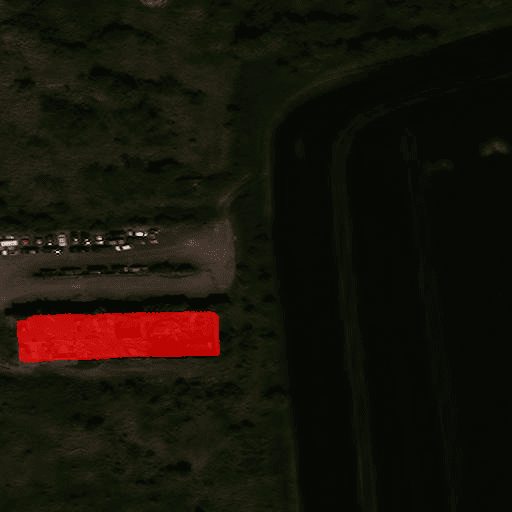} & 
\includegraphics[width=1.05\linewidth]{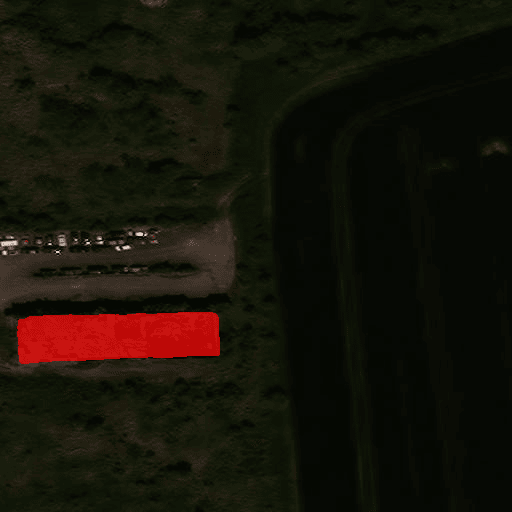} \\

\hline
\end{tabular}
}
\label{prediction_comparison_2}
\end{table}


\begin{table}[h]
\centering
\caption{Comparison of Predictions and Ground Truth Across Models (Cont.)}
\resizebox{\textwidth}{!}{
\begin{tabular}{>{\centering\arraybackslash}p{0.18\textwidth}m{0.22\textwidth}m{0.22\textwidth}m{0.22\textwidth}m{0.22\textwidth}}
\hline
\multicolumn{1}{c}{\textbf{Queries}} & 
\multicolumn{1}{c}{\textbf{RGB}} & 
\multicolumn{1}{c}{\textbf{LISA}} & 
\multicolumn{1}{c}{\textbf{\lisat (Ours)}} & 
\multicolumn{1}{c}{\textbf{Ground Truth}} \\
\hline
\\
\centering\arraybackslash\parbox{0.18\textwidth}{\centering\vspace{-9pt}Segment the damaged building located in the top-right of the image.
} & 
\includegraphics[width=1.05\linewidth]{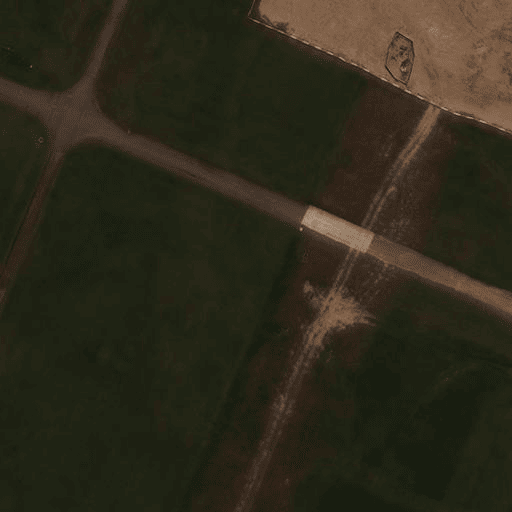} & 
\includegraphics[width=1.05\linewidth]{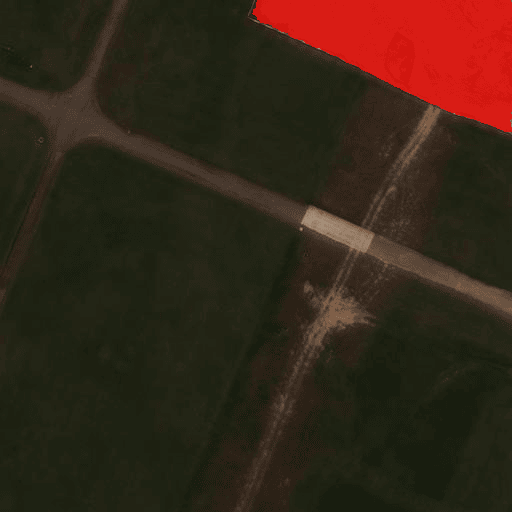} & 
\includegraphics[width=1.05\linewidth]{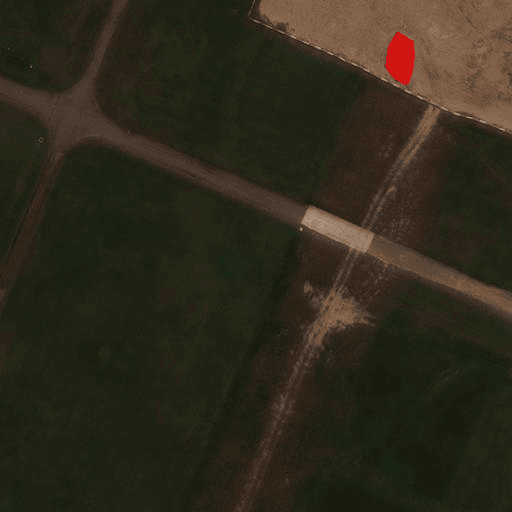} & 
\includegraphics[width=1.05\linewidth]{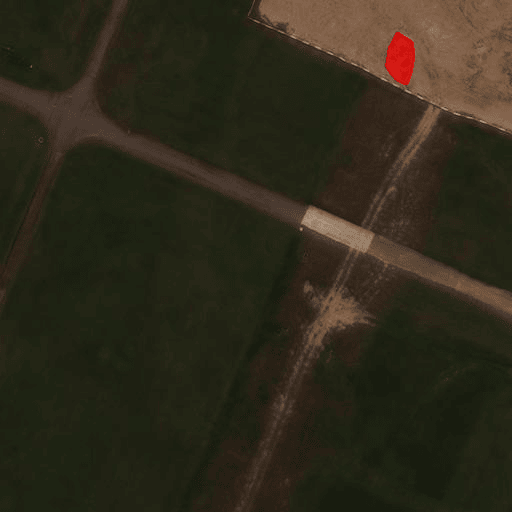} \\

\centering\arraybackslash\parbox{0.18\textwidth}{\centering\vspace{-9pt}Identify the building in the center-left of the image.
} & 
\includegraphics[width=1.05\linewidth]{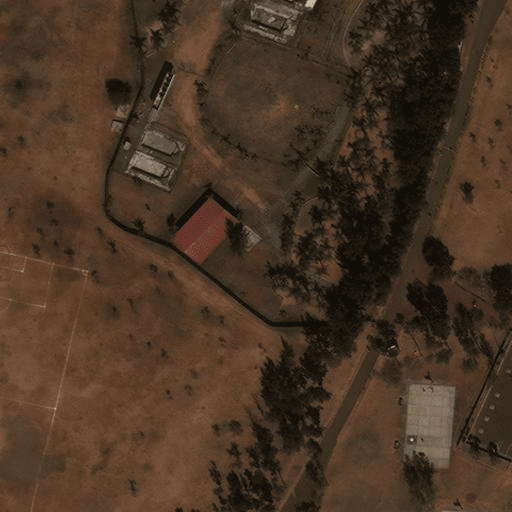} & 
\includegraphics[width=1.05\linewidth]{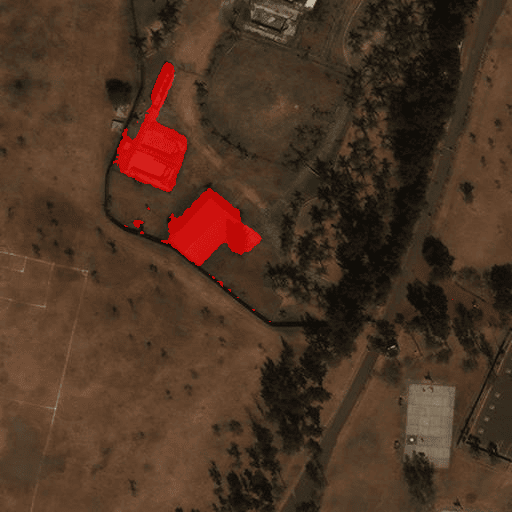} & 
\includegraphics[width=1.05\linewidth]{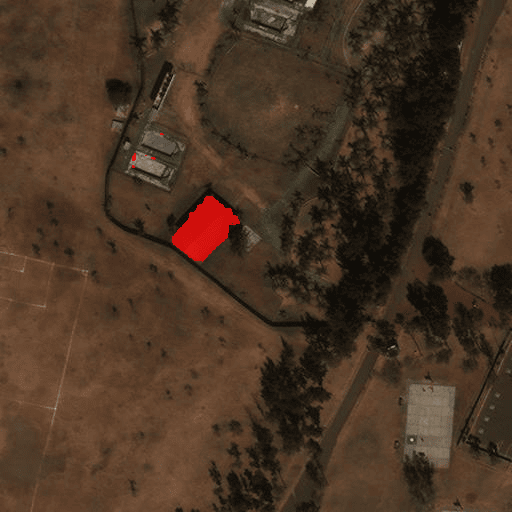} & 
\includegraphics[width=1.05\linewidth]{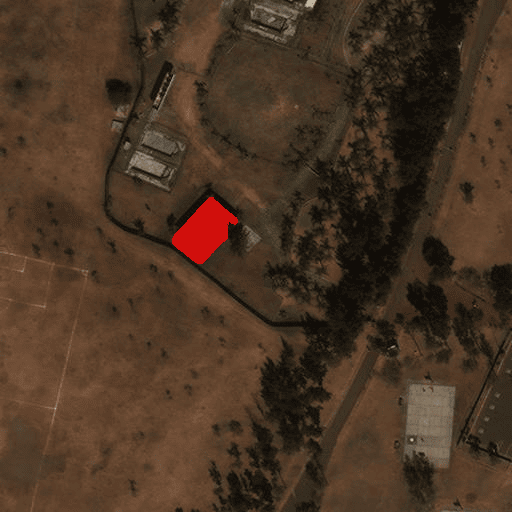} \\

\centering\arraybackslash\parbox{0.18\textwidth}{\centering\vspace{-9pt}Identify the building with a unique vertical dark brown structure with a slight curvature on the edge.
} & 
\includegraphics[width=1.05\linewidth]{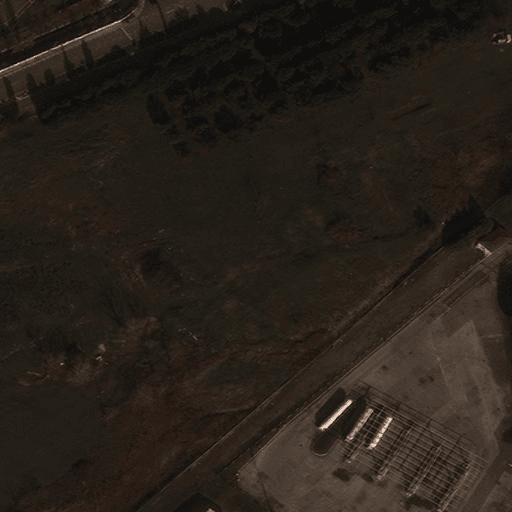} & 
\includegraphics[width=1.05\linewidth]{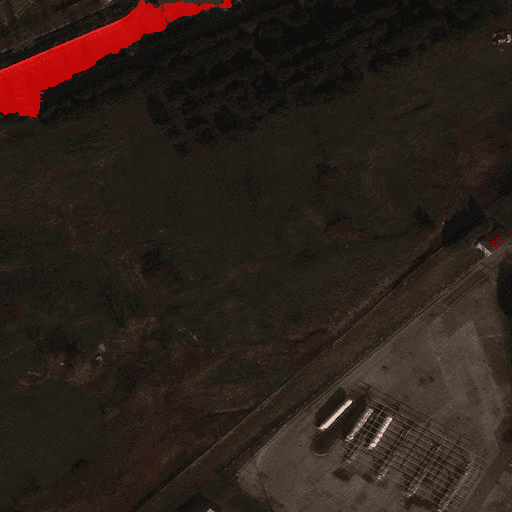} & 
\includegraphics[width=1.05\linewidth]{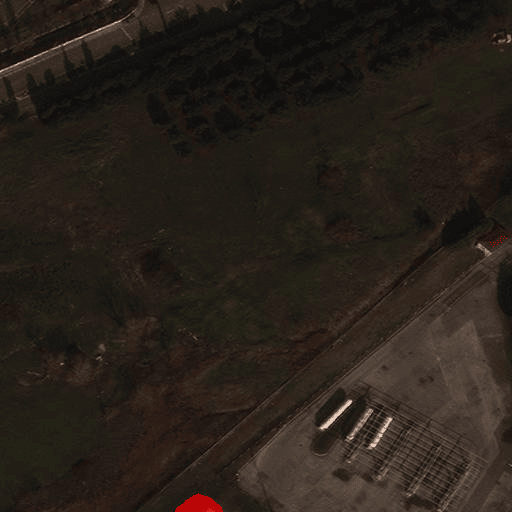} & 
\includegraphics[width=1.05\linewidth]{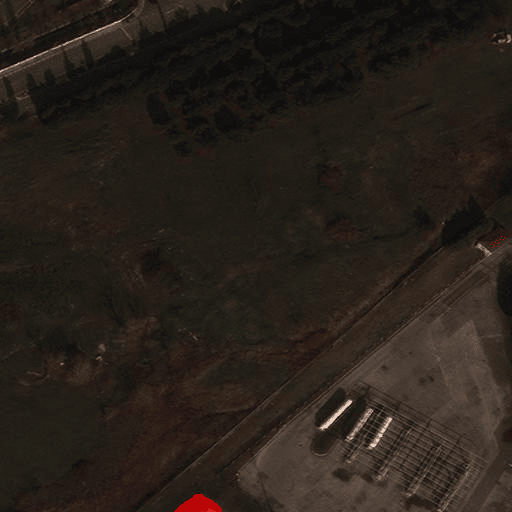} \\

\centering\arraybackslash\parbox{0.18\textwidth}{\centering\vspace{-9pt}Identify the large, rectangular building with a dark roof and multiple visible roof fixtures.
} & 
\includegraphics[width=1.05\linewidth]{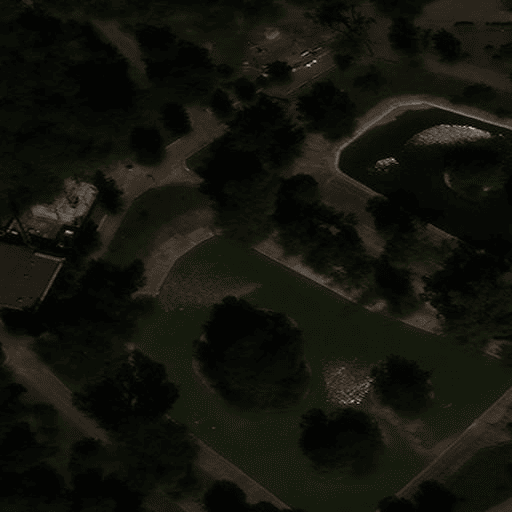} & 
\includegraphics[width=1.05\linewidth]{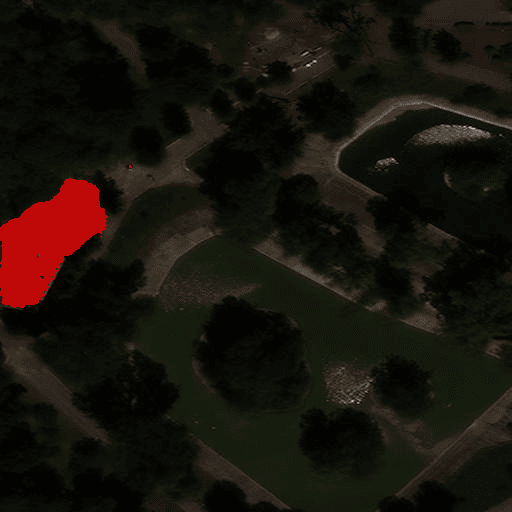} & 
\includegraphics[width=1.05\linewidth]{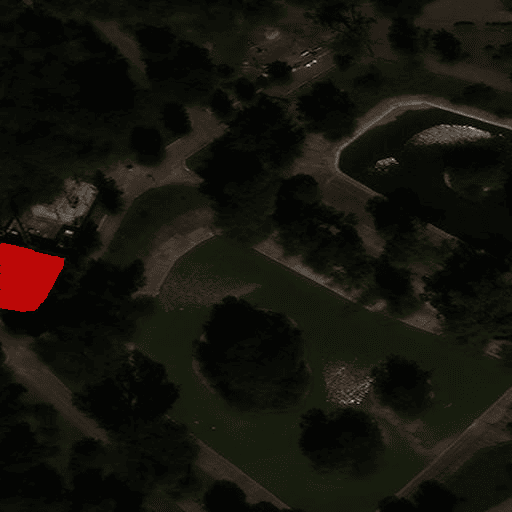} & 
\includegraphics[width=1.05\linewidth]{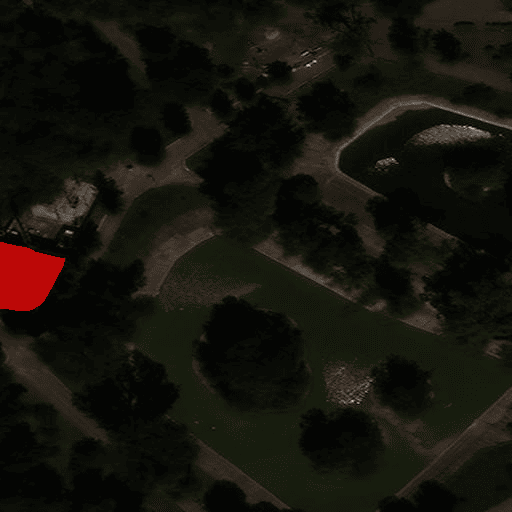} \\

\centering\arraybackslash\parbox{0.18\textwidth}{\centering\vspace{-9pt}Locate the trailer bed in the top-right of the image, characterized by a long rectangular shape with distinct wheels, typically metallic or painted in color, attached to a truck cab.
} & 
\includegraphics[width=1.05\linewidth]{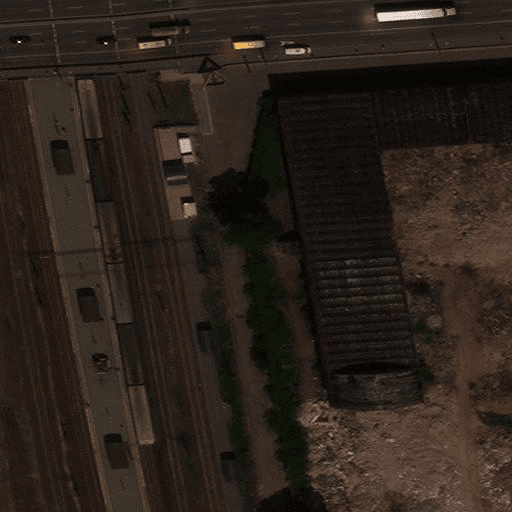} & 
\includegraphics[width=1.05\linewidth]{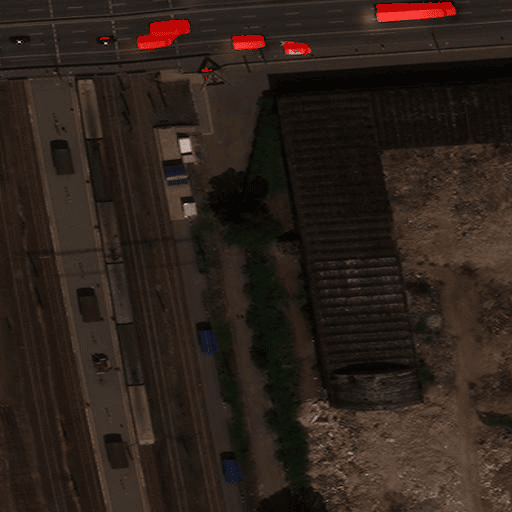} & 
\includegraphics[width=1.05\linewidth]{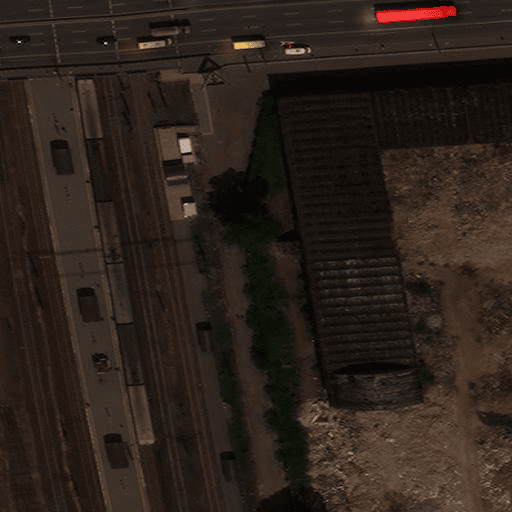} & 
\includegraphics[width=1.05\linewidth]{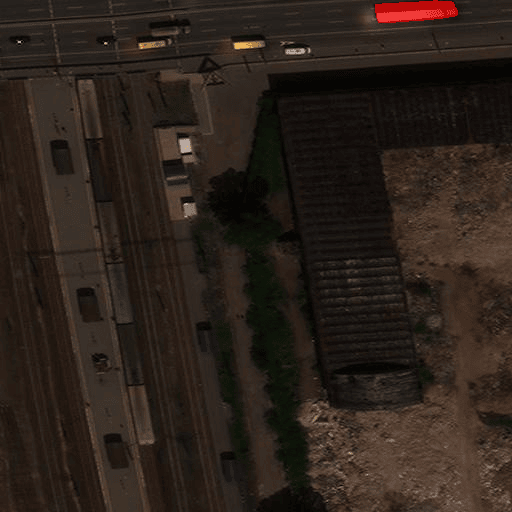} \\

\centering\arraybackslash\parbox{0.18\textwidth}{\centering\vspace{-9pt}Identify the liquid tank in the top-right of the image with a long rectangular shape connected to a truck cab at the front.
} & 
\includegraphics[width=1.05\linewidth]{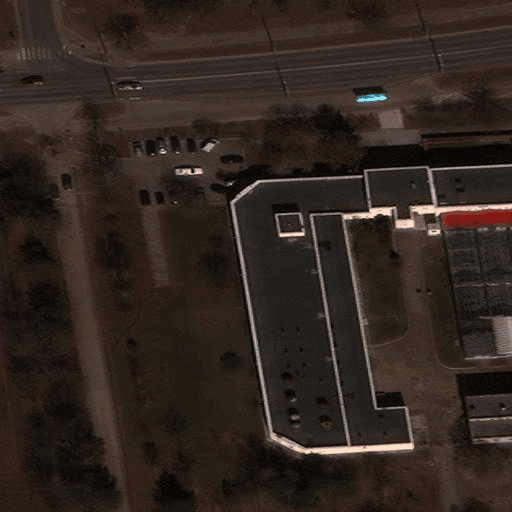} & 
\includegraphics[width=1.05\linewidth]{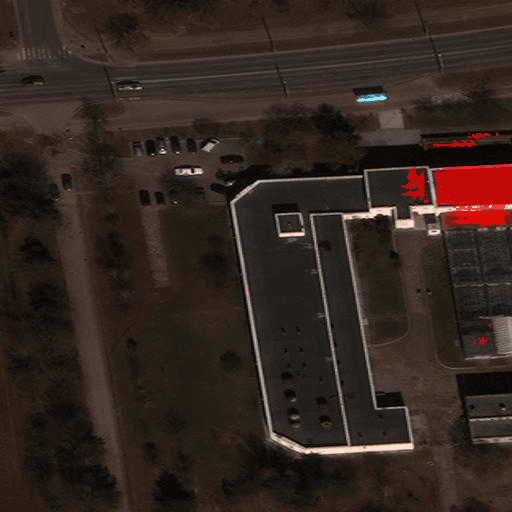} & 
\includegraphics[width=1.05\linewidth]{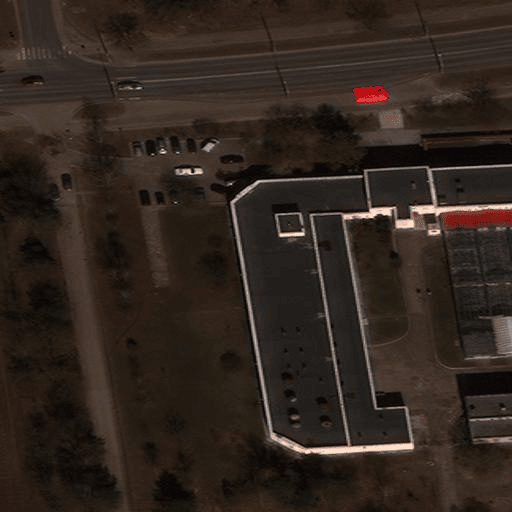} & 
\includegraphics[width=1.05\linewidth]{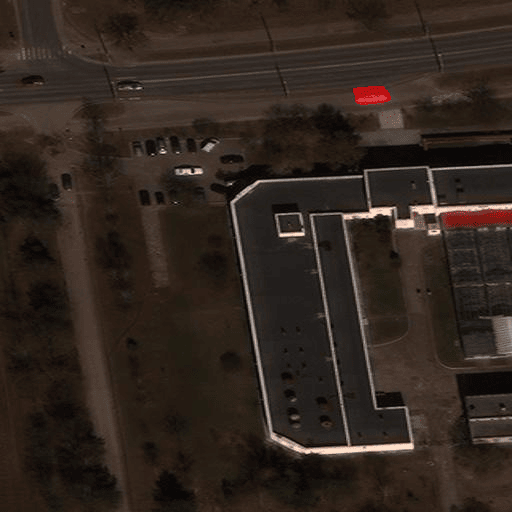} \\

\end{tabular}%
}
\label{prediction_comparison_3}
\end{table}


\begin{table}[h]
\centering
\caption{Comparison of Predictions and Ground Truth Across Models (Cont.)}
\resizebox{\textwidth}{!}{
\begin{tabular}{>{\centering\arraybackslash}p{0.18\textwidth}m{0.22\textwidth}m{0.22\textwidth}m{0.22\textwidth}m{0.22\textwidth}}
\hline
\multicolumn{1}{c}{\textbf{Queries}} & 
\multicolumn{1}{c}{\textbf{RGB}} & 
\multicolumn{1}{c}{\textbf{LISA}} & 
\multicolumn{1}{c}{\textbf{\lisat (Ours)}} & 
\multicolumn{1}{c}{\textbf{Ground Truth}} \\
\hline
\\
\centering\arraybackslash\parbox{0.18\textwidth}{\centering\vspace{-9pt}Locate the building with a reddish-brown roof next to a dark black structure in the image.
} & 
\includegraphics[width=1.05\linewidth]{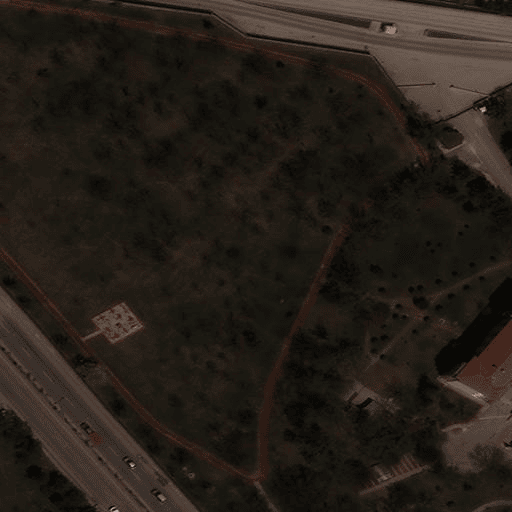} & 
\includegraphics[width=1.05\linewidth]{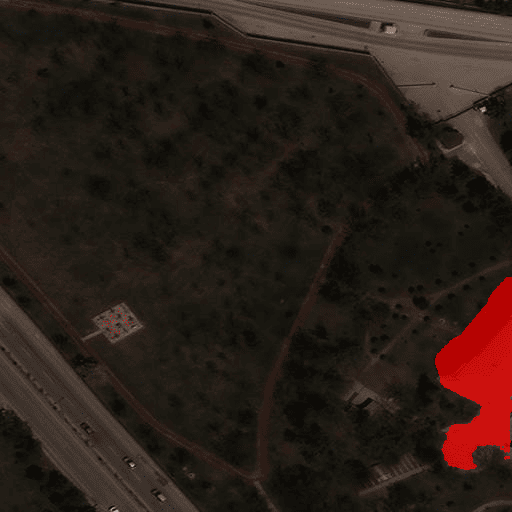} & 
\includegraphics[width=1.05\linewidth]{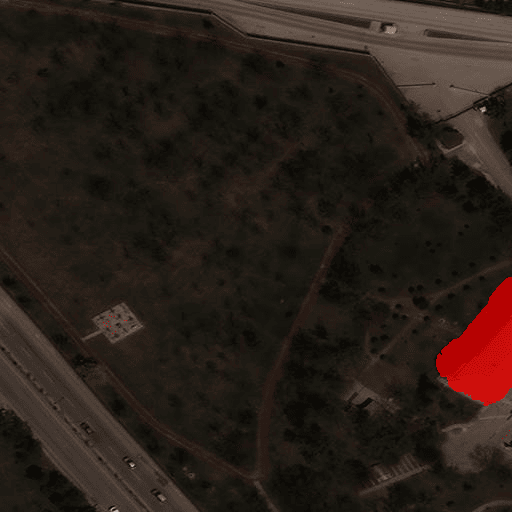} & 
\includegraphics[width=1.05\linewidth]{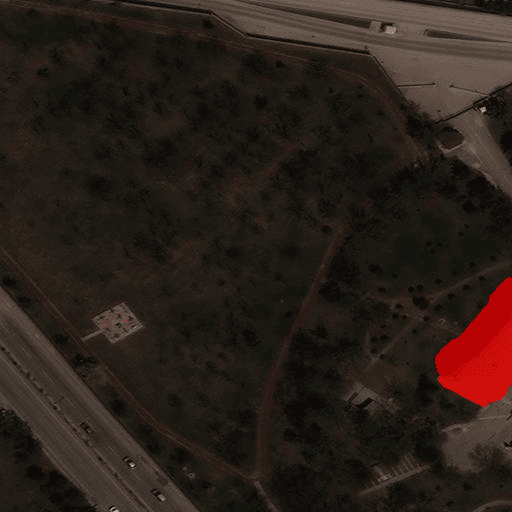} \\

\centering\arraybackslash\parbox{0.18\textwidth}{\centering\vspace{-9pt}Identify the damaged building in the center of the image.
} & 
\includegraphics[width=1.05\linewidth]{figures/success_images/COCO_train_000000013309/COCO_train_000000013309_original.png} & 
\includegraphics[width=1.05\linewidth]{figures/success_images/COCO_train_000000013309/COCO_train_000000013309_LISA.png} & 
\includegraphics[width=1.05\linewidth]{figures/success_images/COCO_train_000000013309/COCO_train_000000013309_LISAt.png} & 
\includegraphics[width=1.05\linewidth]{figures/success_images/COCO_train_000000013309/COCO_train_000000013309_GT.png} \\

\centering\arraybackslash\parbox{0.18\textwidth}{\centering\vspace{-9pt}Locate the maritime vessel in the bottom-right of the image.
} & 
\includegraphics[width=1.05\linewidth]{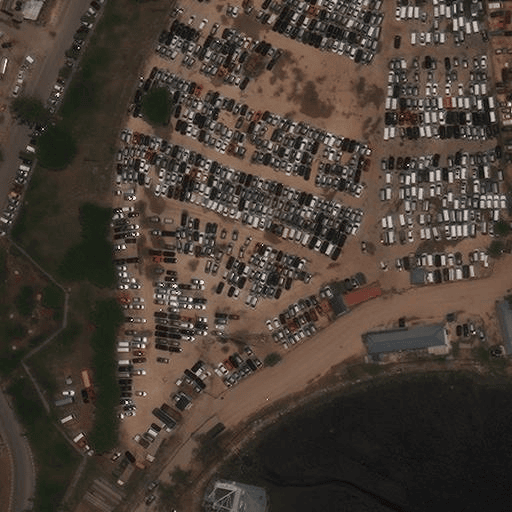} & 
\includegraphics[width=1.05\linewidth]{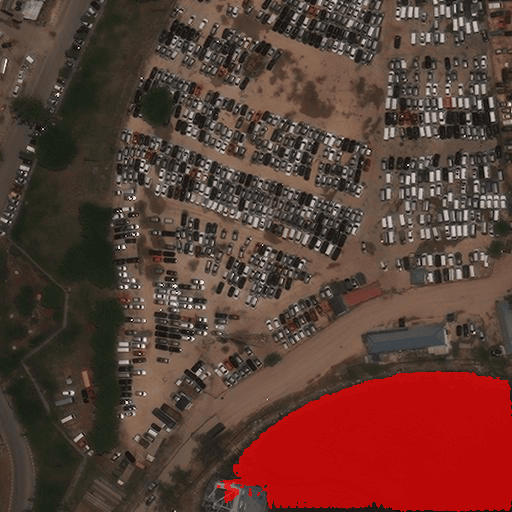} & 
\includegraphics[width=1.05\linewidth]{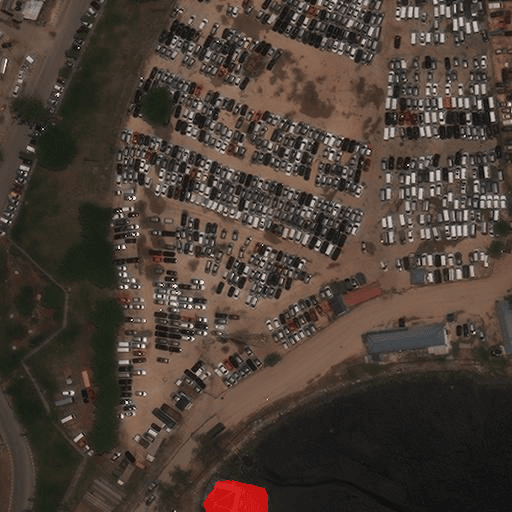} & 
\includegraphics[width=1.05\linewidth]{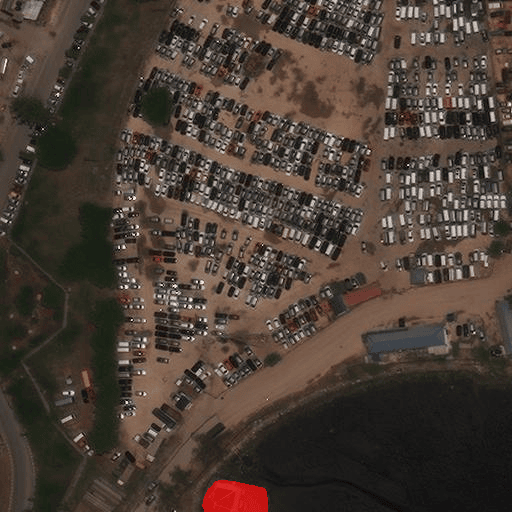} \\

\centering\arraybackslash\parbox{0.18\textwidth}{\centering\vspace{-9pt}Identify the building with a rectangular shape, dark roof, and noticeable white lines across its surface, set against a brownish background with green areas nearby.
} & 
\includegraphics[width=1.05\linewidth]{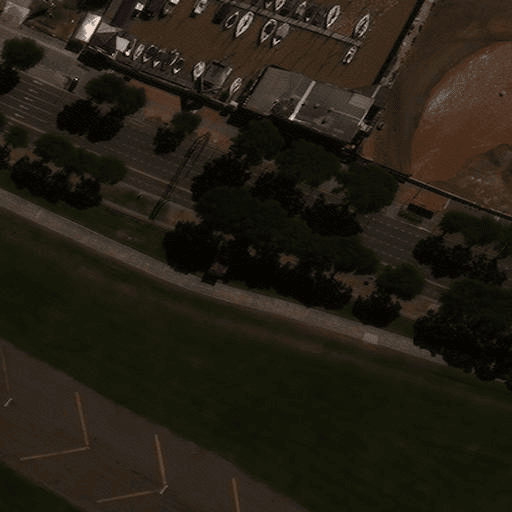} & 
\includegraphics[width=1.05\linewidth]{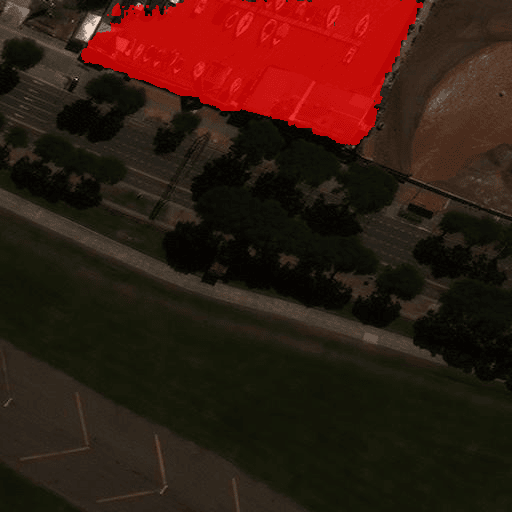} & 
\includegraphics[width=1.05\linewidth]{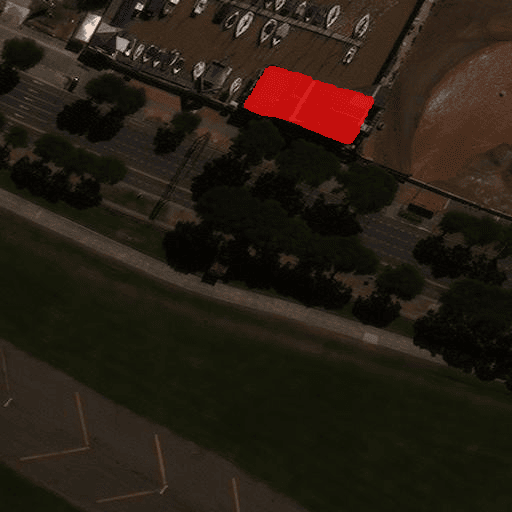} & 
\includegraphics[width=1.05\linewidth]{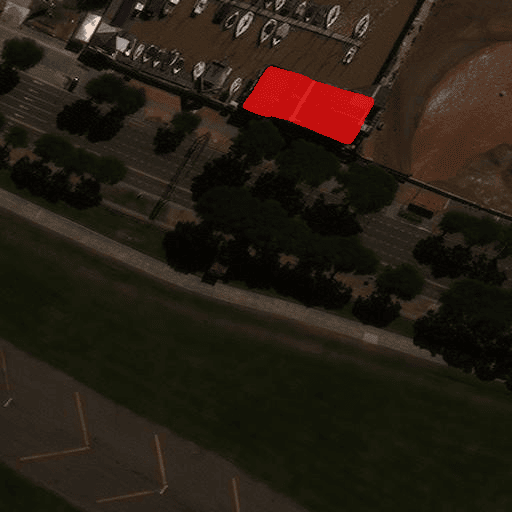} \\

\centering\arraybackslash\parbox{0.18\textwidth}{\centering\vspace{-9pt}Locate the barge in the bottom-left of the image.
} & 
\includegraphics[width=1.05\linewidth]{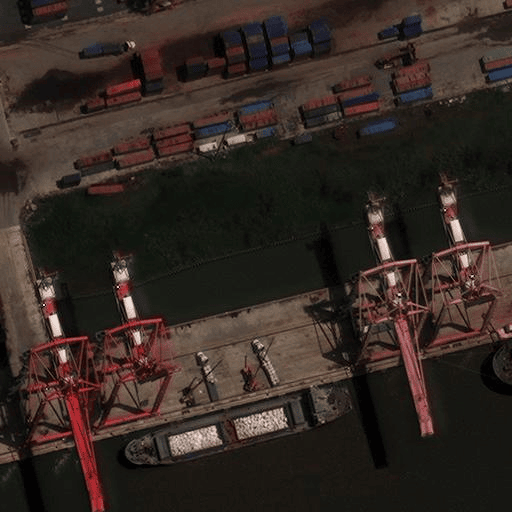} & 
\includegraphics[width=1.05\linewidth]{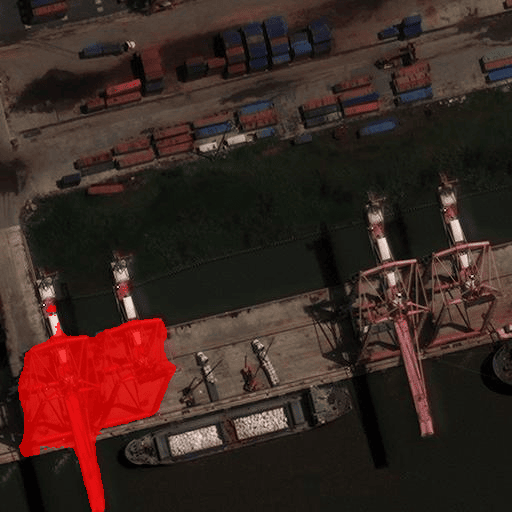} & 
\includegraphics[width=1.05\linewidth]{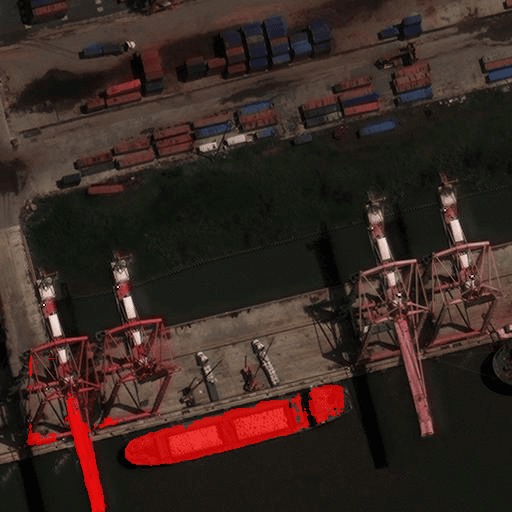} & 
\includegraphics[width=1.05\linewidth]{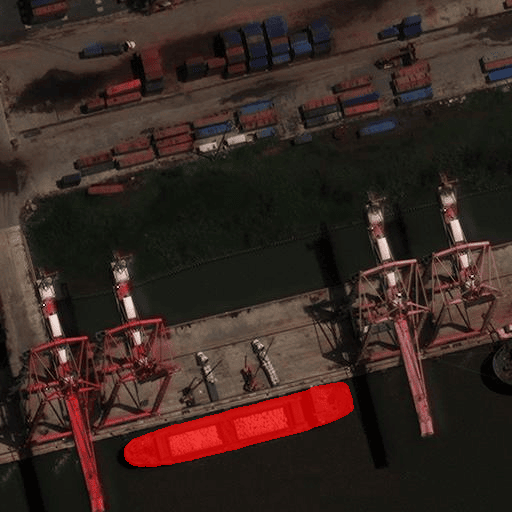} \\

\centering\arraybackslash\parbox{0.18\textwidth}{\centering\vspace{-9pt}Locate the engineering vehicle in the top-left of the image.
} & 
\includegraphics[width=1.05\linewidth]{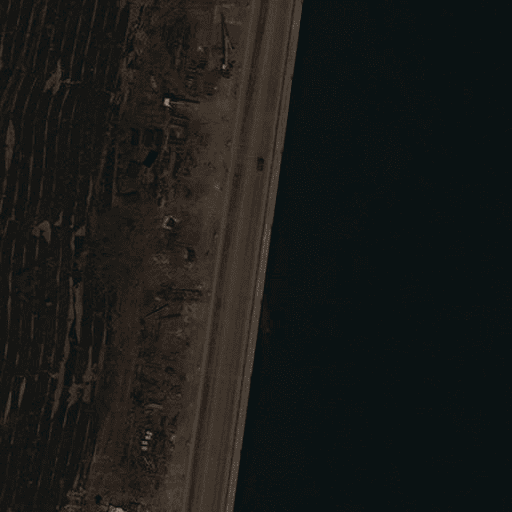} & 
\includegraphics[width=1.05\linewidth]{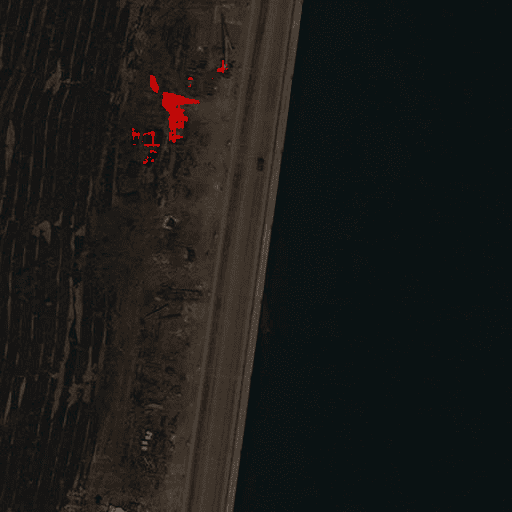} & 
\includegraphics[width=1.05\linewidth]{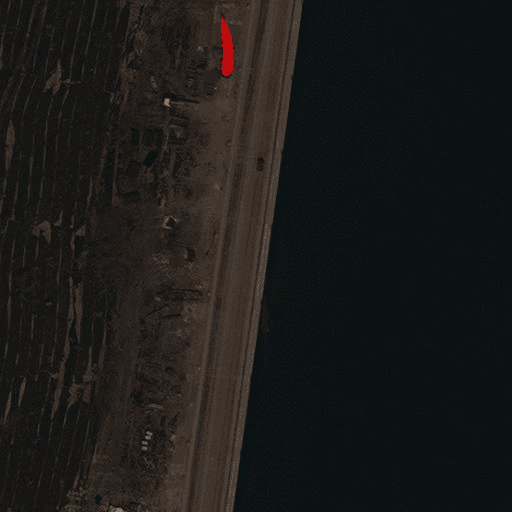} & 
\includegraphics[width=1.05\linewidth]{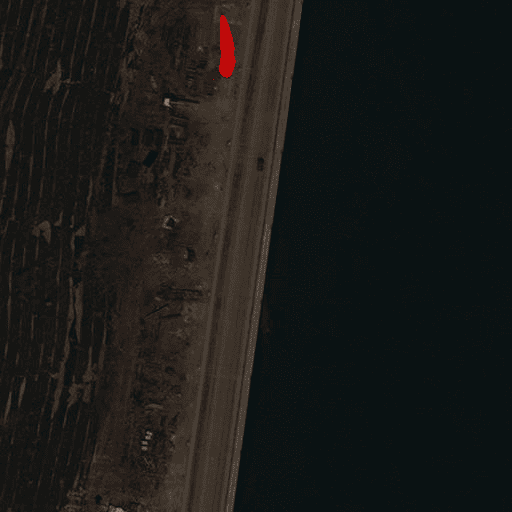} \\

\end{tabular}%
}
\label{prediction_comparison_4}
\end{table}


\begin{table}[h]
\centering
\caption{Comparison of Predictions and Ground Truth Across Models (Cont.)}
\resizebox{\textwidth}{!}{
\begin{tabular}{>{\centering\arraybackslash}p{0.18\textwidth}m{0.22\textwidth}m{0.22\textwidth}m{0.22\textwidth}m{0.22\textwidth}}
\hline
\multicolumn{1}{c}{\textbf{Queries}} & 
\multicolumn{1}{c}{\textbf{RGB}} & 
\multicolumn{1}{c}{\textbf{LISA}} & 
\multicolumn{1}{c}{\textbf{\lisat (Ours)}} & 
\multicolumn{1}{c}{\textbf{Ground Truth}} \\
\hline
\\
\centering\arraybackslash\parbox{0.18\textwidth}
{\centering\vspace{-9pt}Locate the large rectangular structure with stacked, multicolored containers floating on water as the container ship.
} &
\includegraphics[width=1.05\linewidth]{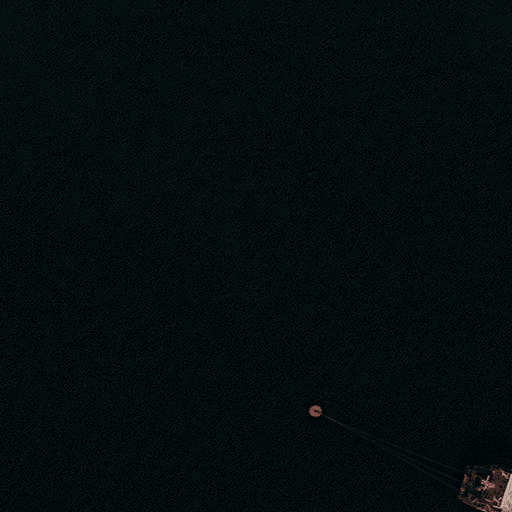} & 
\includegraphics[width=1.05\linewidth]{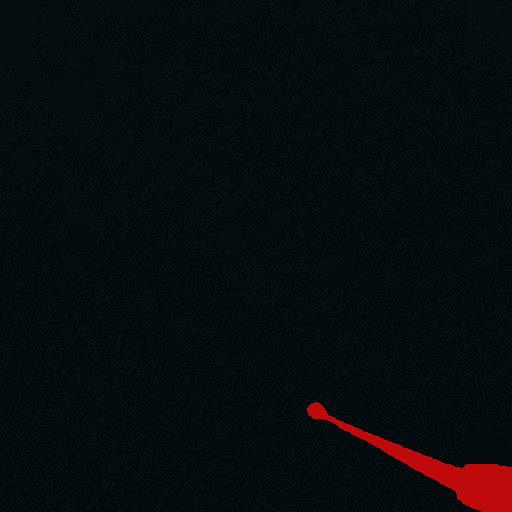} & 
\includegraphics[width=1.05\linewidth]{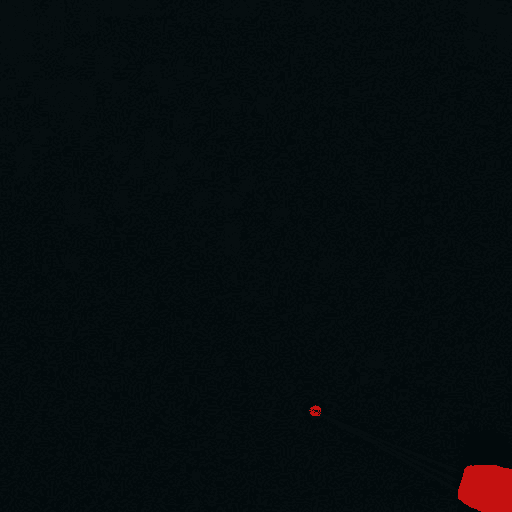} & 
\includegraphics[width=1.05\linewidth]{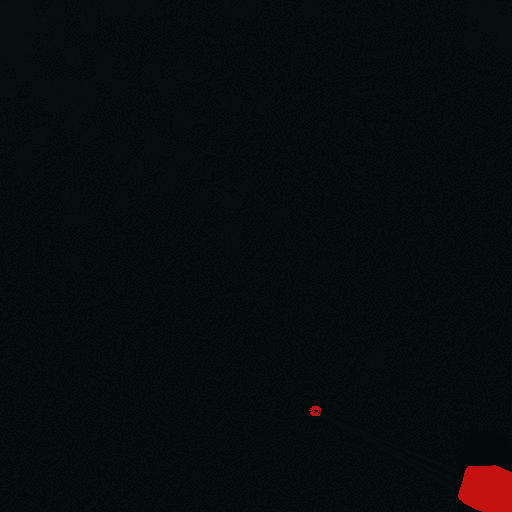} \\

\centering\arraybackslash\parbox{0.18\textwidth}
{\centering\vspace{-9pt}Locate the building in the top-left corner of the image.
} &
\includegraphics[width=1.05\linewidth]{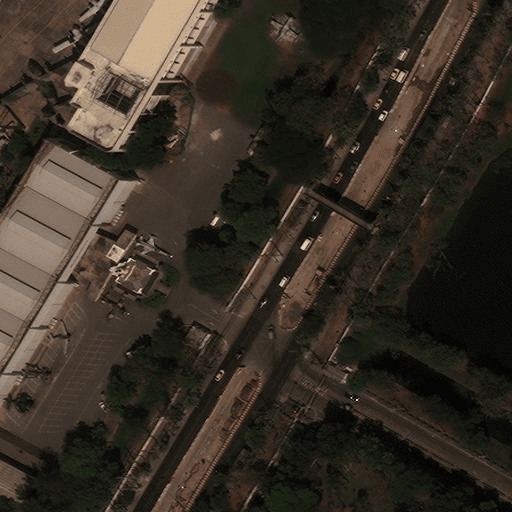} & 
\includegraphics[width=1.05\linewidth]{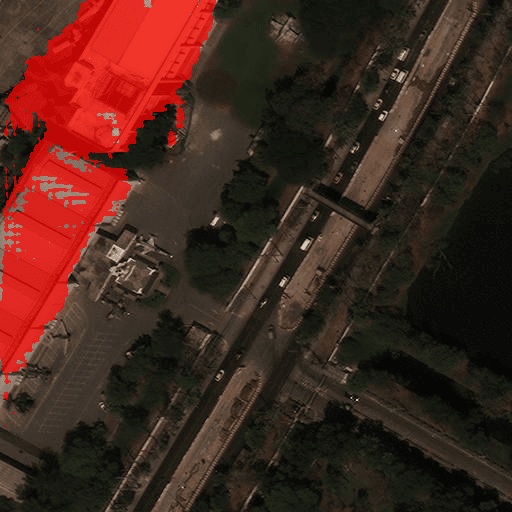} & 
\includegraphics[width=1.05\linewidth]{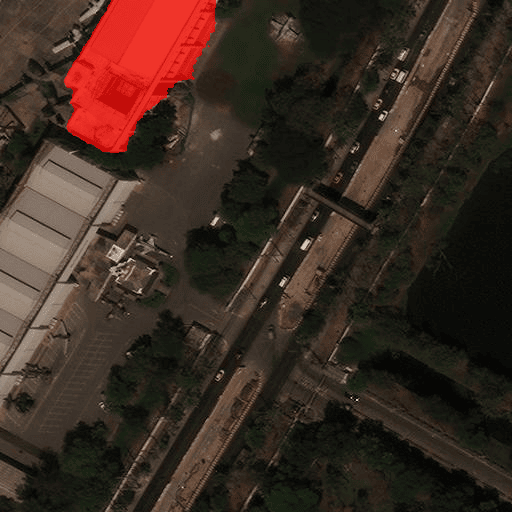} & 
\includegraphics[width=1.05\linewidth]{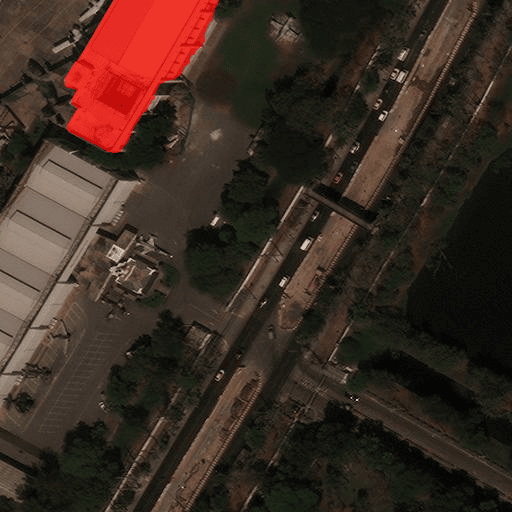} \\

\centering\arraybackslash\parbox{0.18\textwidth}
{\centering\vspace{-9pt}Identify the maritime vessel near the top-left corner of the image.
} &
\includegraphics[width=1.05\linewidth]{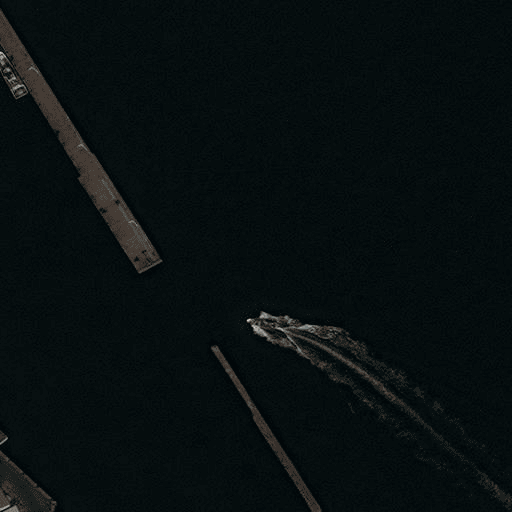} & 
\includegraphics[width=1.05\linewidth]{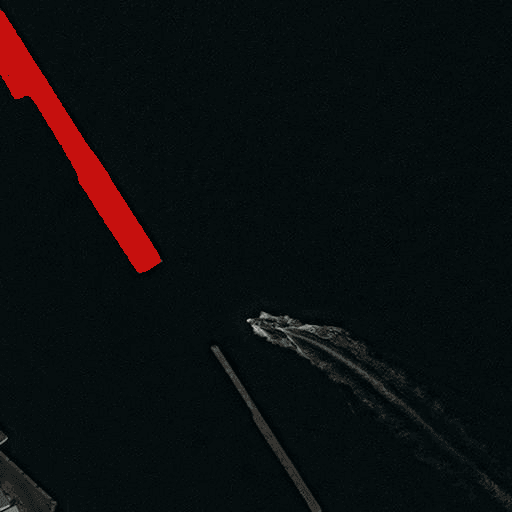} & 
\includegraphics[width=1.05\linewidth]{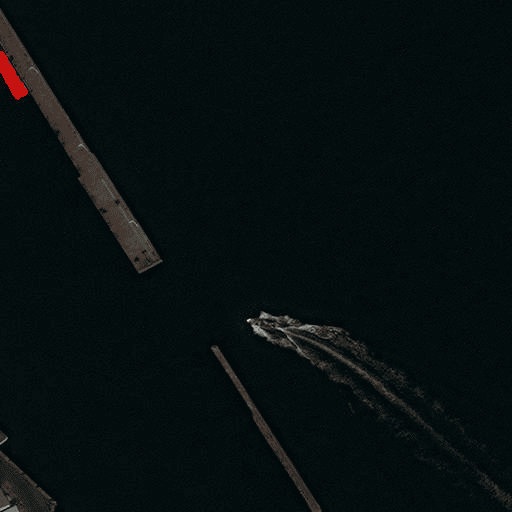} & 
\includegraphics[width=1.05\linewidth]{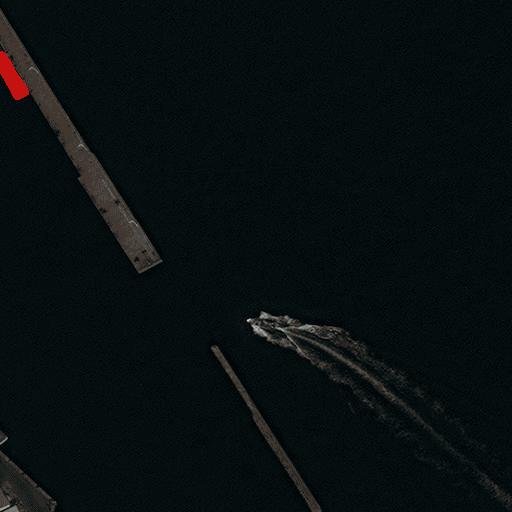} \\

\centering\arraybackslash\parbox{0.18\textwidth}
{\centering\vspace{-9pt}Identify the aircraft hangar with the large rectangular structure and curved roof, displaying a uniform beige coloration and surrounded by open areas.
} &
\includegraphics[width=1.05\linewidth]{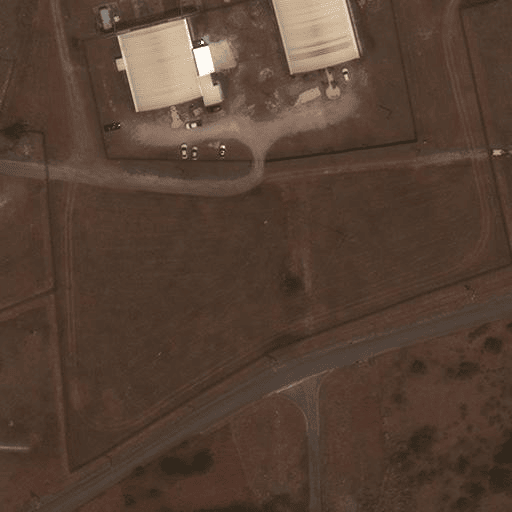} & 
\includegraphics[width=1.05\linewidth]{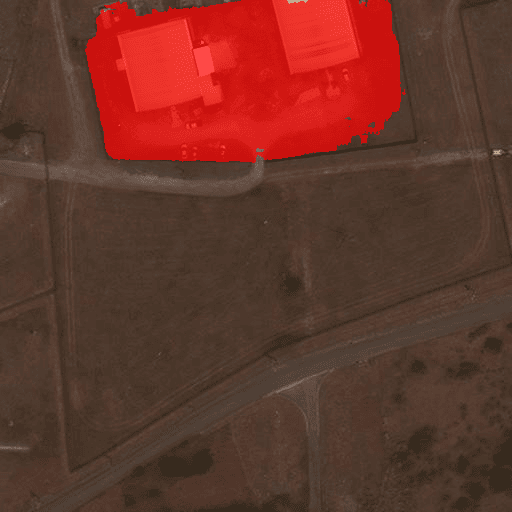} & 
\includegraphics[width=1.05\linewidth]{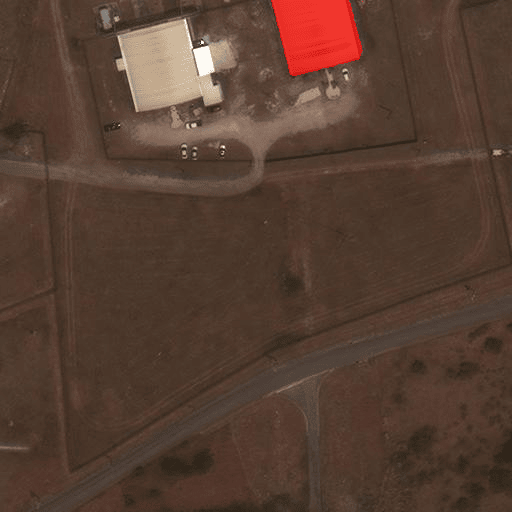} & 
\includegraphics[width=1.05\linewidth]{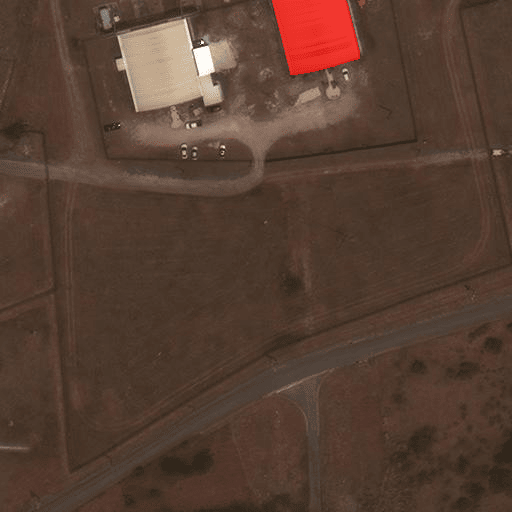} \\

\centering\arraybackslash\parbox{0.18\textwidth}
{\centering\vspace{-9pt}Identify the large rectangular brown building with a flat roof surrounded by vegetation.
} &
\includegraphics[width=1.05\linewidth]{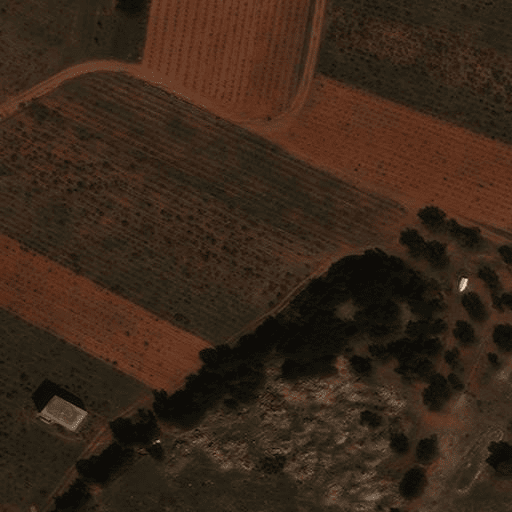} & 
\includegraphics[width=1.05\linewidth]{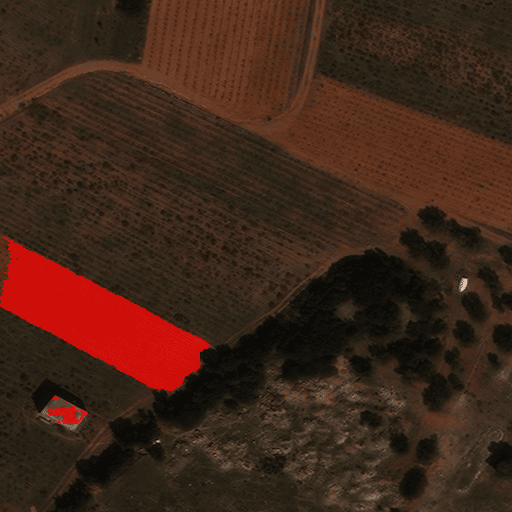} & 
\includegraphics[width=1.05\linewidth]{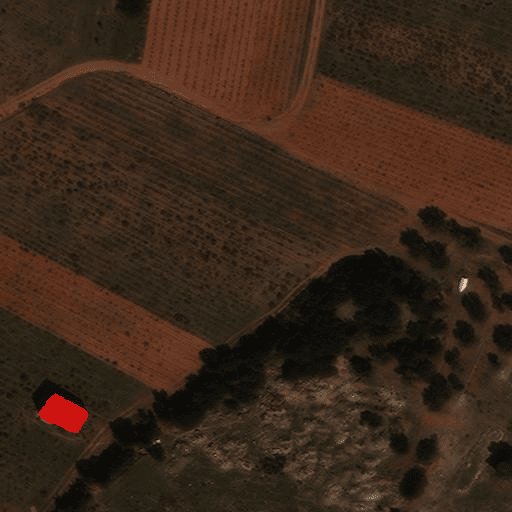} & 
\includegraphics[width=1.05\linewidth]{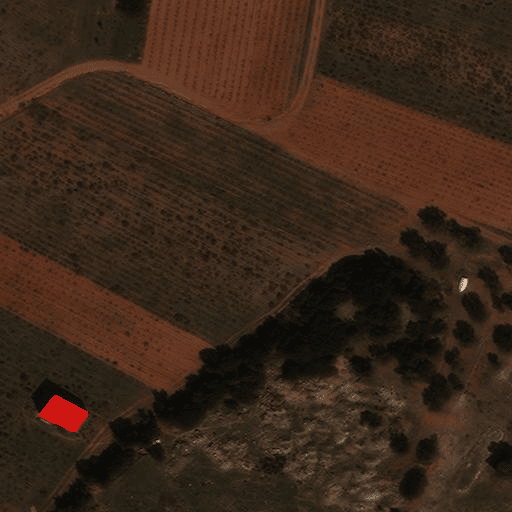} \\

\centering\arraybackslash\parbox{0.18\textwidth}
{\centering\vspace{-9pt}Identify the railway vehicle with an elongated, rectangular shape and a metallic texture contrasting against the dark background.
} & 
\includegraphics[width=1.05\linewidth]{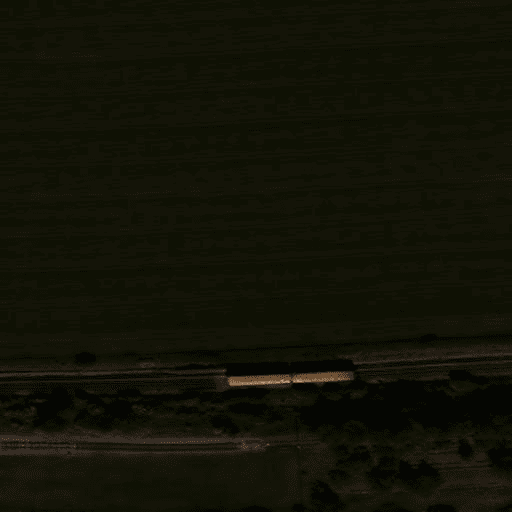} & 
\includegraphics[width=1.05\linewidth]{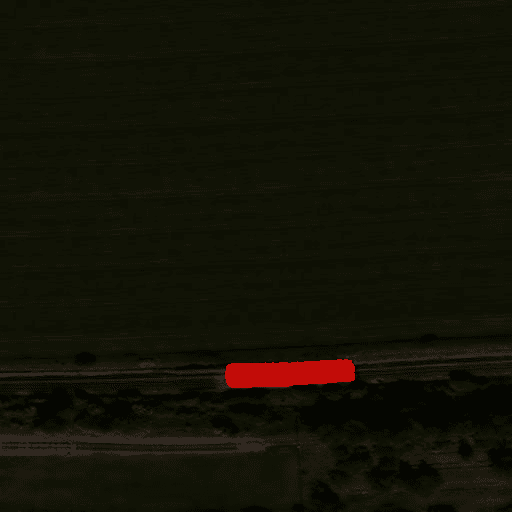} & 
\includegraphics[width=1.05\linewidth]{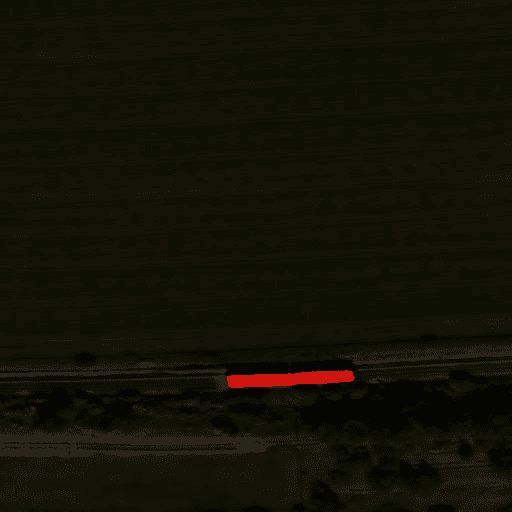} & 
\includegraphics[width=1.05\linewidth]{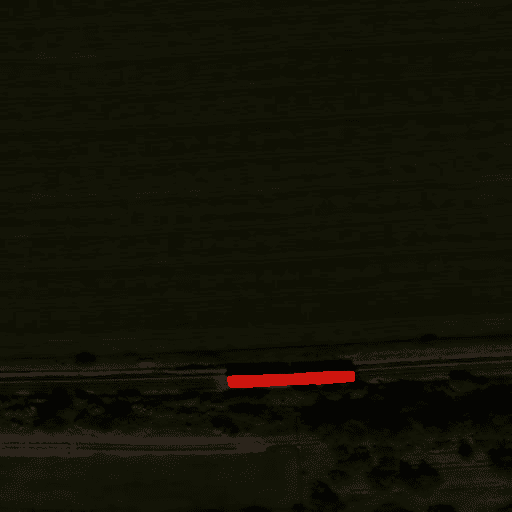} \\

\end{tabular}%
}
\label{prediction_comparison_5}
\end{table}

\subsection{Failure Cases of \lisat}
\label{Failure_Cases_of_LISAt}

We examined failure cases where \lisat struggled to make accurate predictions in \autoref{limitations}. Some of these instances, where the model's performance could be improved, highlight the challenges it faces under complex conditions, such as cloudy or ambiguous scenes as shown in \autoref{failure_prediction_comparison}.

\begin{table}[h]
\centering
\caption{Failure Cases}
\resizebox{\textwidth}{!}{
\begin{tabular}{>{\centering\arraybackslash}p{0.18\textwidth}m{0.22\textwidth}m{0.22\textwidth}m{0.22\textwidth}m{0.22\textwidth}}
\hline
\multicolumn{1}{c}{\textbf{Queries}} & 
\multicolumn{1}{c}{\textbf{RGB}} & 
\multicolumn{1}{c}{\textbf{LISA}} & 
\multicolumn{1}{c}{\textbf{\lisat (Ours)}} & 
\multicolumn{1}{c}{\textbf{Ground Truth}} \\
\hline
\\

\centering\arraybackslash\parbox{0.18\textwidth}
{\centering\vspace{-9pt}Locate the facility in the top-center of the image for identification.
} &
\includegraphics[width=1.05\linewidth]{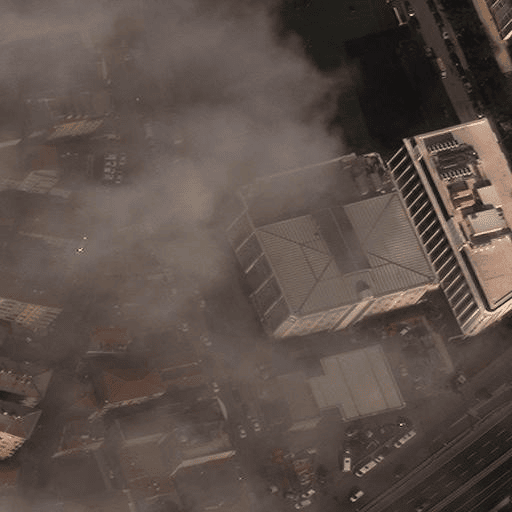} & 
\includegraphics[width=1.05\linewidth]{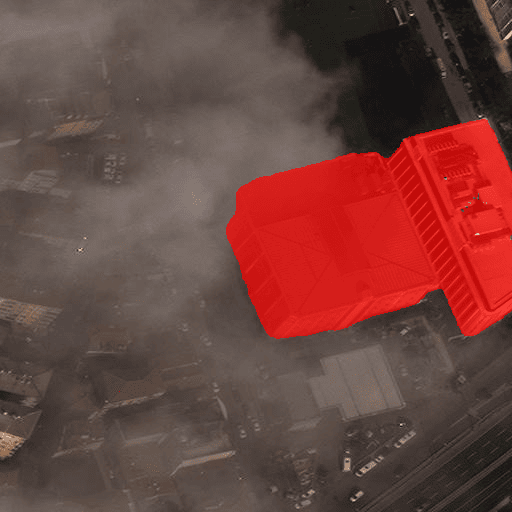} & 
\includegraphics[width=1.05\linewidth]{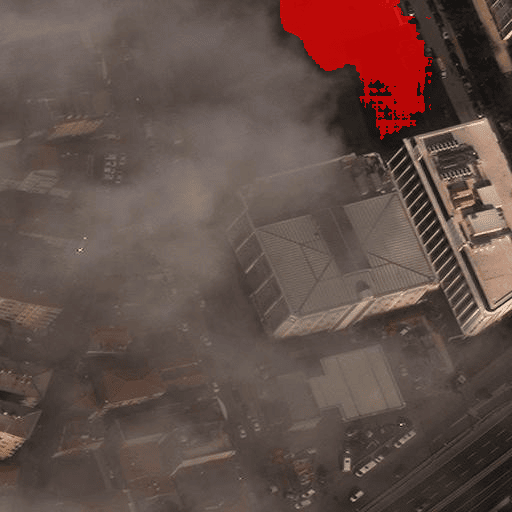} & 
\includegraphics[width=1.05\linewidth]{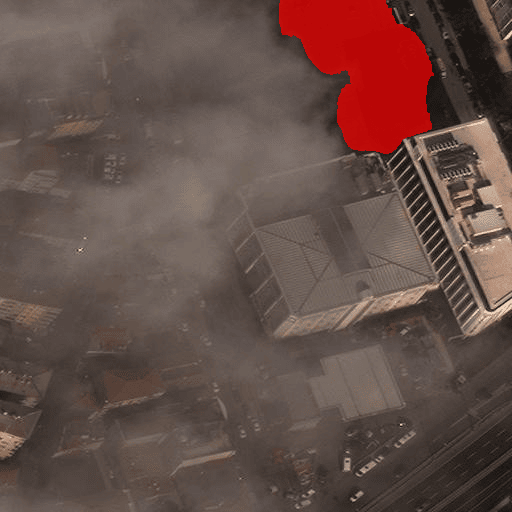} \\

\centering\arraybackslash\parbox{0.18\textwidth}
{\centering\vspace{-9pt}Find the facility in the bottom-left corner of the image.
} &
\includegraphics[width=1.05\linewidth]{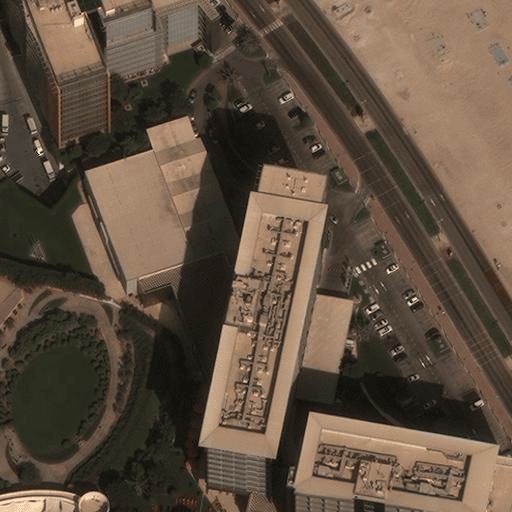} & 
\includegraphics[width=1.05\linewidth]{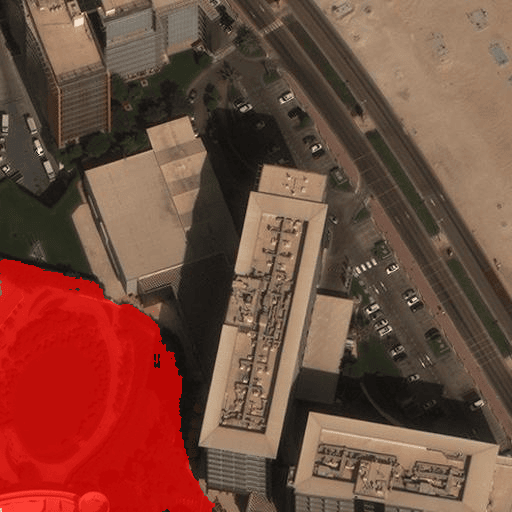} & 
\includegraphics[width=1.05\linewidth]{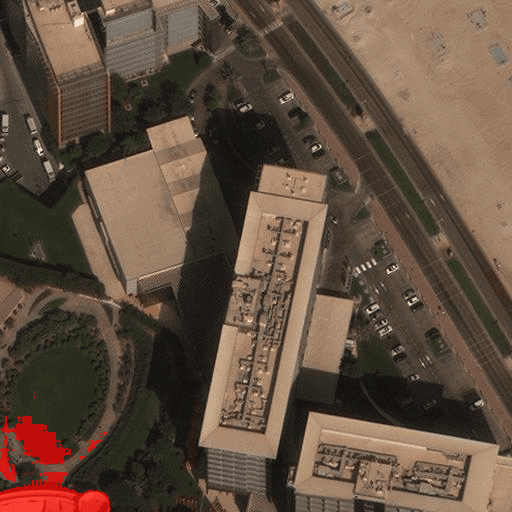} & 
\includegraphics[width=1.05\linewidth]{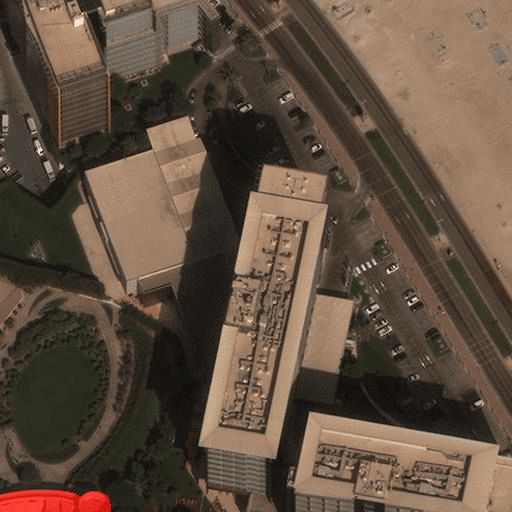} \\

\centering\arraybackslash\parbox{0.18\textwidth}
{\centering\vspace{-9pt}Identify the plane in the bottom-right of the image.
} &
\includegraphics[width=1.05\linewidth]{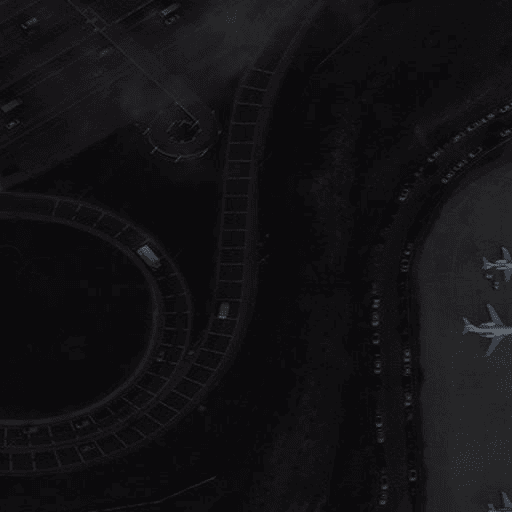} & 
\includegraphics[width=1.05\linewidth]{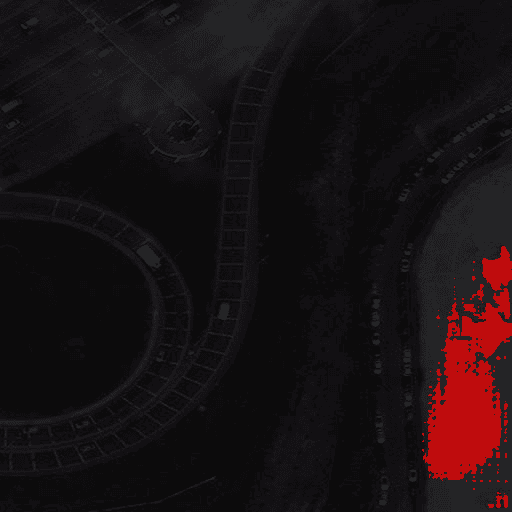} & 
\includegraphics[width=1.05\linewidth]{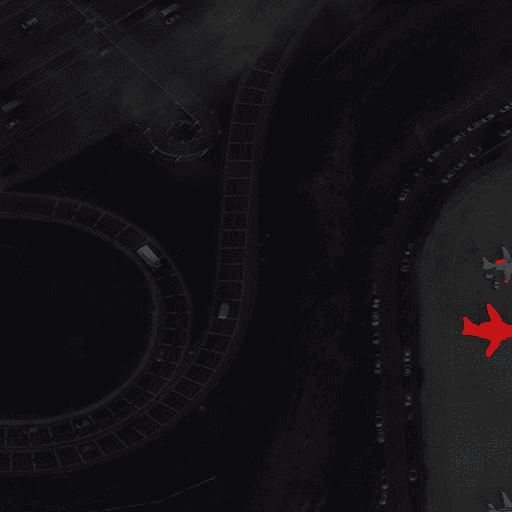} & 
\includegraphics[width=1.05\linewidth]{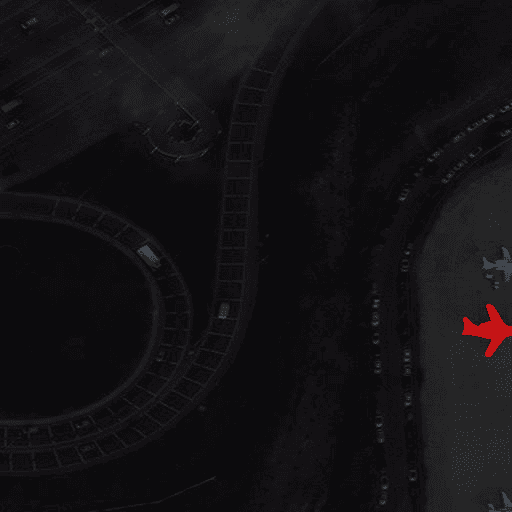} \\

\centering\arraybackslash\parbox{0.18\textwidth}
{\centering\vspace{-9pt}Locate the barge in the top-left of the image.
} &
\includegraphics[width=1.05\linewidth]{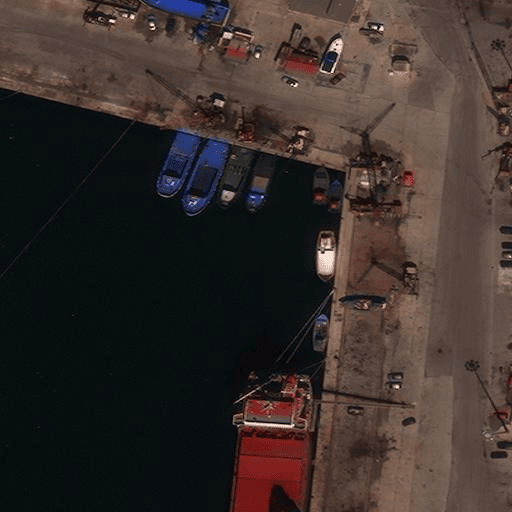} & 
\includegraphics[width=1.05\linewidth]{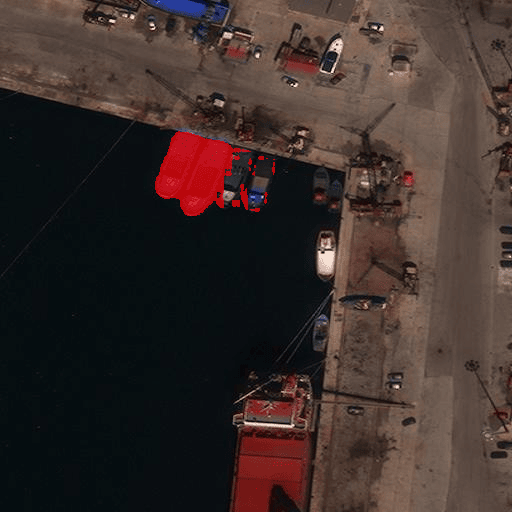} & 
\includegraphics[width=1.05\linewidth]{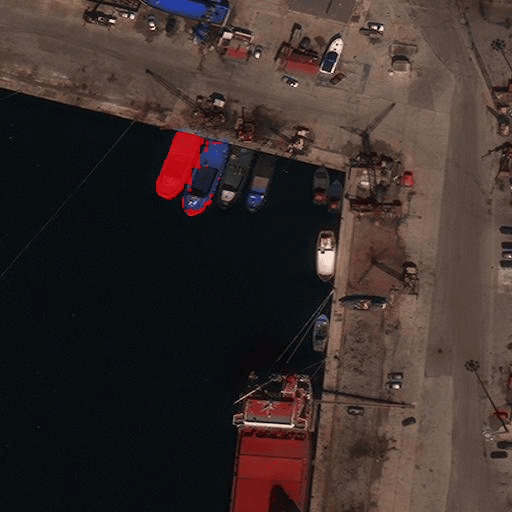} & 
\includegraphics[width=1.05\linewidth]{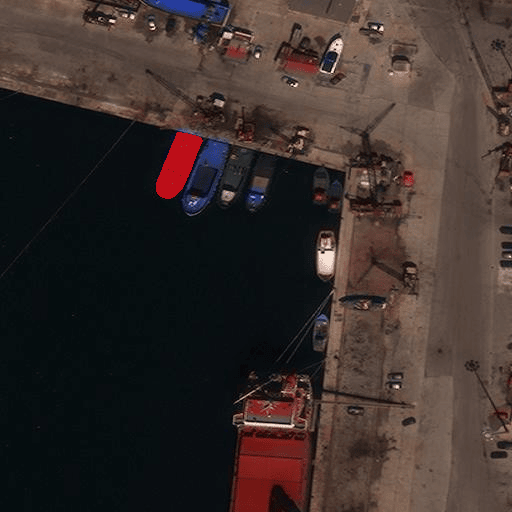} \\

\centering\arraybackslash\parbox{0.18\textwidth}
{\centering\vspace{-9pt}Locate the building with a distinctive light gray color and rectangular shape against the darker background.
} &
\includegraphics[width=1.05\linewidth]{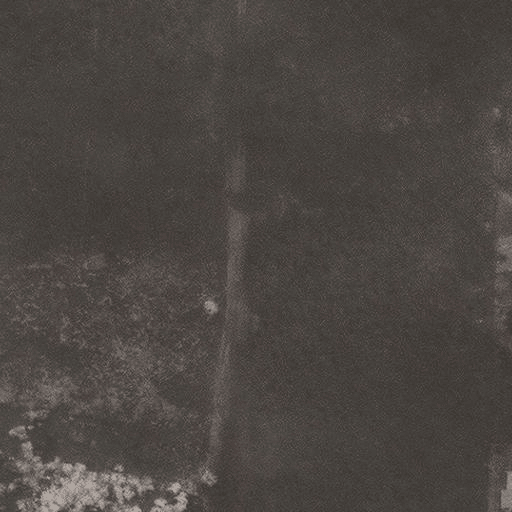} & 
\includegraphics[width=1.05\linewidth]{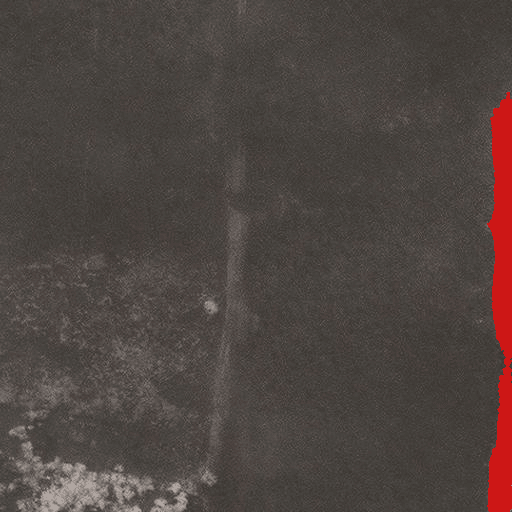} & 
\includegraphics[width=1.05\linewidth]{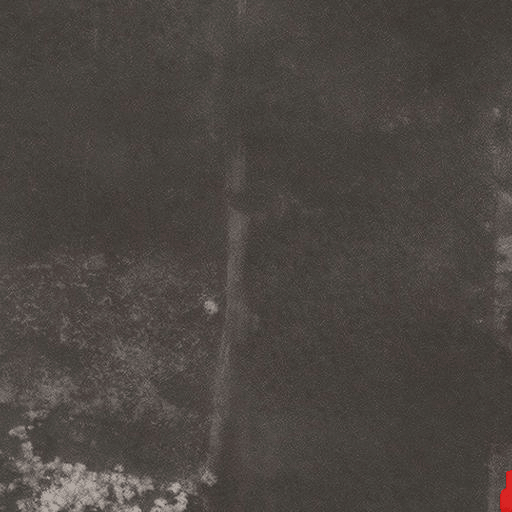} & 
\includegraphics[width=1.05\linewidth]{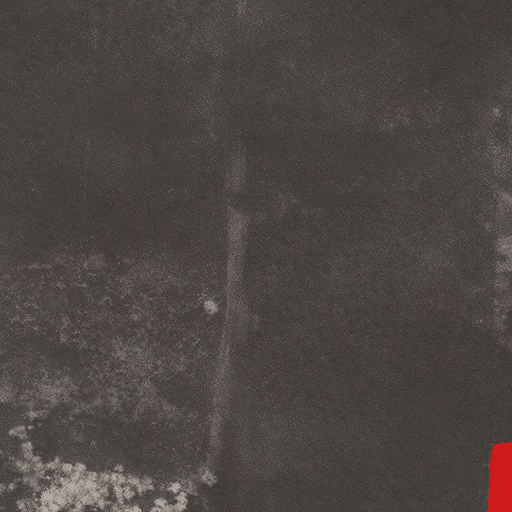} \\

\centering\arraybackslash\parbox{0.18\textwidth}
{\centering\vspace{-9pt}Identify the trailer in the bottom-right of the image with a distinct shape, typically metallic or painted, connected to a truck cab at the front.
} &
\includegraphics[width=1.05\linewidth]{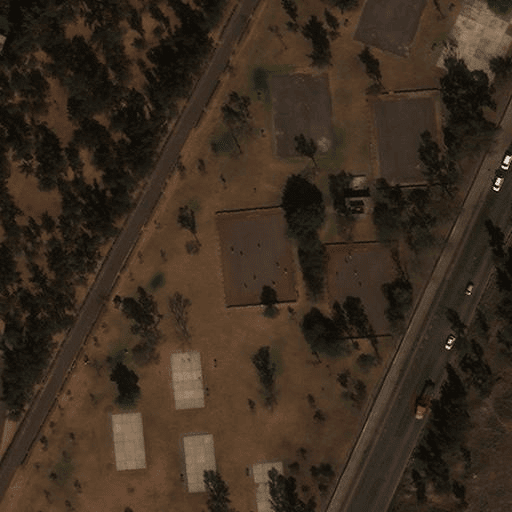} & 
\includegraphics[width=1.05\linewidth]{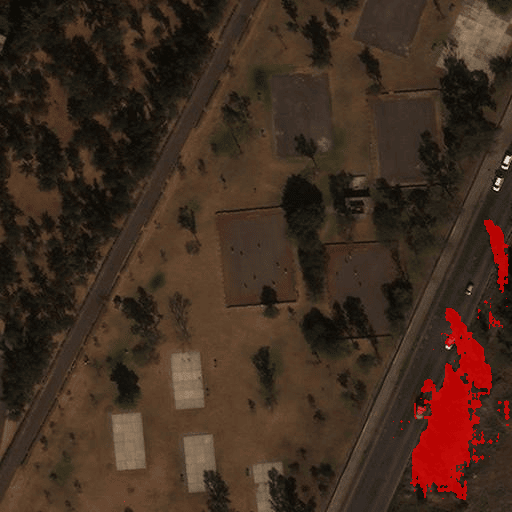} & 
\includegraphics[width=1.05\linewidth]{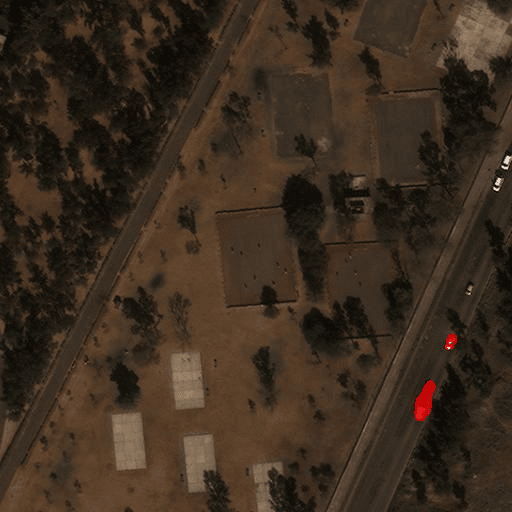} & 
\includegraphics[width=1.05\linewidth]{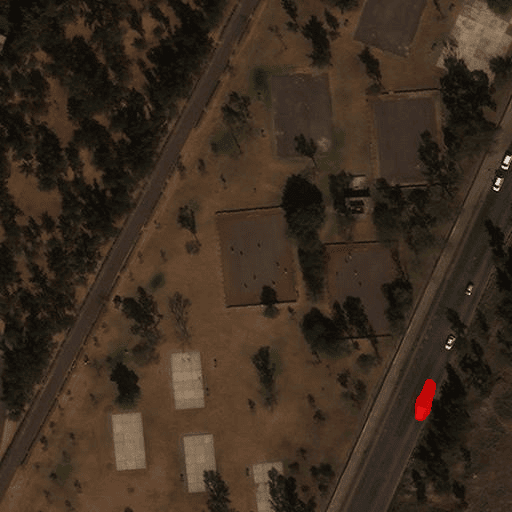} \\

\end{tabular}%
}
\label{failure_prediction_comparison}
\end{table}

\subsection{Ground Truth Error Cases}
\label{Ground_Truth_Error_Cases}
\autoref{gt_mistake_prediction_comparison} displays cases where the model's predictions are affected by errors in the ground truth data. These errors highlight discrepancies between the model's output and the labeled data, shedding light on limitations within the dataset and the potential impact on evaluation metrics.

\begin{table}[h]
\centering
\caption{GT Mistake Cases}
\resizebox{\textwidth}{!}{
\begin{tabular}{>{\centering\arraybackslash}p{0.18\textwidth}m{0.22\textwidth}m{0.22\textwidth}m{0.22\textwidth}m{0.22\textwidth}}
\hline
\multicolumn{1}{c}{\textbf{Queries}} & 
\multicolumn{1}{c}{\textbf{RGB}} & 
\multicolumn{1}{c}{\textbf{LISA}} & 
\multicolumn{1}{c}{\textbf{\lisat (Ours)}} & 
\multicolumn{1}{c}{\textbf{Ground Truth}} \\
\hline
\\

\centering\arraybackslash\parbox{0.18\textwidth}
{\centering\vspace{-9pt}Identify the pylon in the top-right area of the image.
} &
\includegraphics[width=1.05\linewidth]{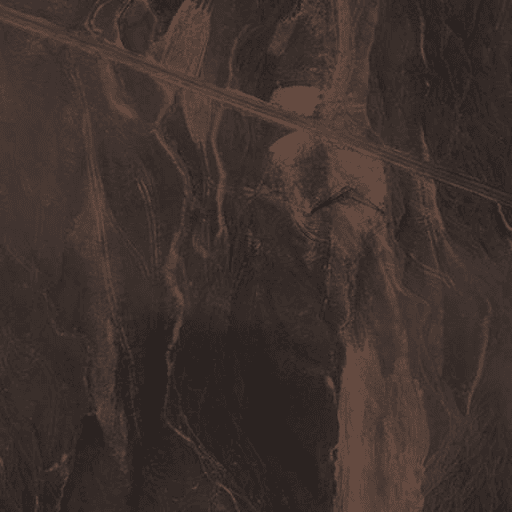} & 
\includegraphics[width=1.05\linewidth]{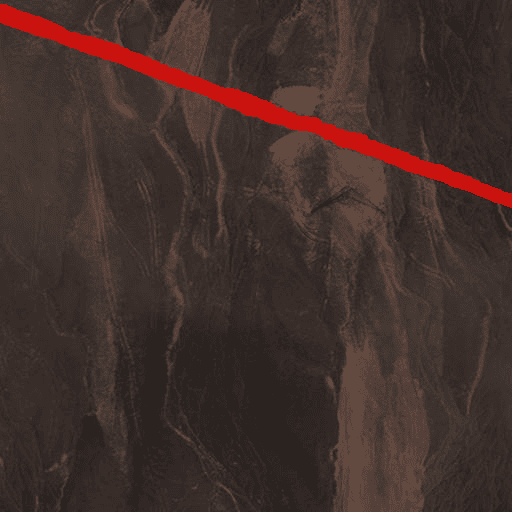} & 
\includegraphics[width=1.05\linewidth]{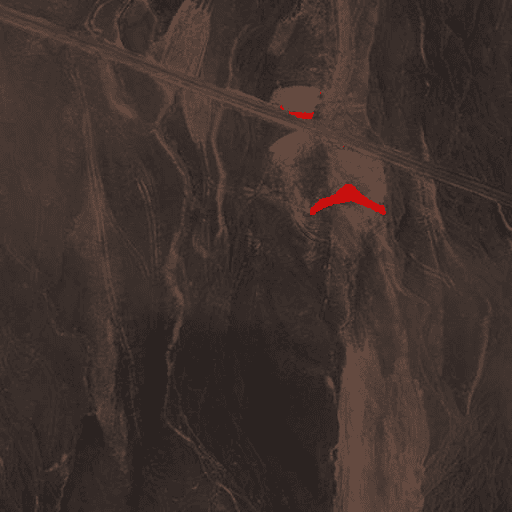} & 
\includegraphics[width=1.05\linewidth]{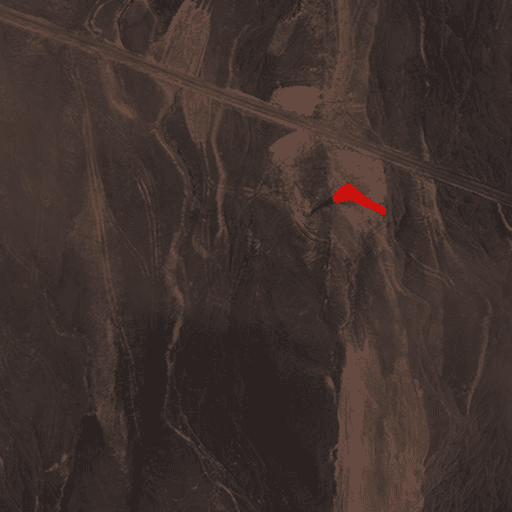} \\

\centering\arraybackslash\parbox{0.18\textwidth}
{\centering\vspace{-9pt}Identify the vertical, metallic structure with a lattice framework contrasting against the brown, earthy background.
} &
\includegraphics[width=1.05\linewidth]{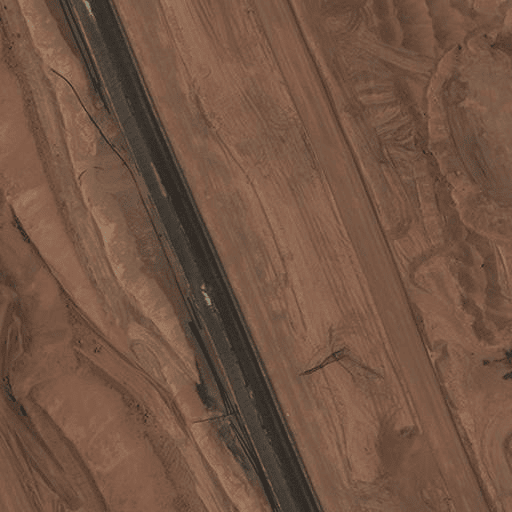} & 
\includegraphics[width=1.05\linewidth]{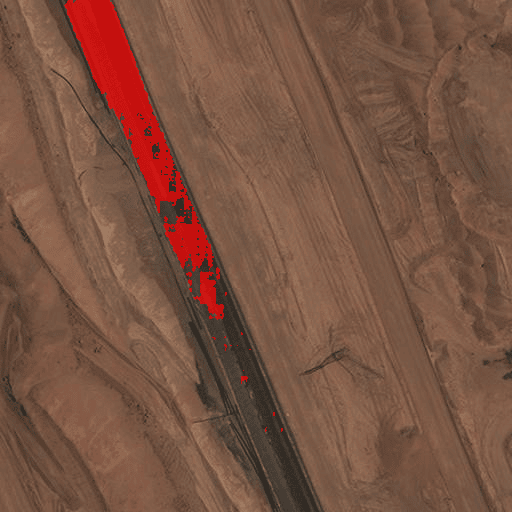} & 
\includegraphics[width=1.05\linewidth]{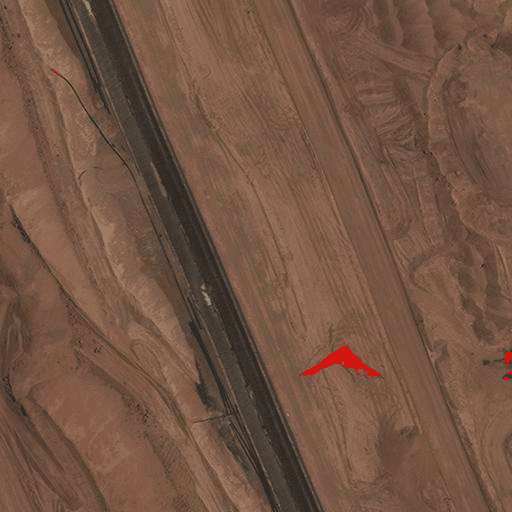} & 
\includegraphics[width=1.05\linewidth]{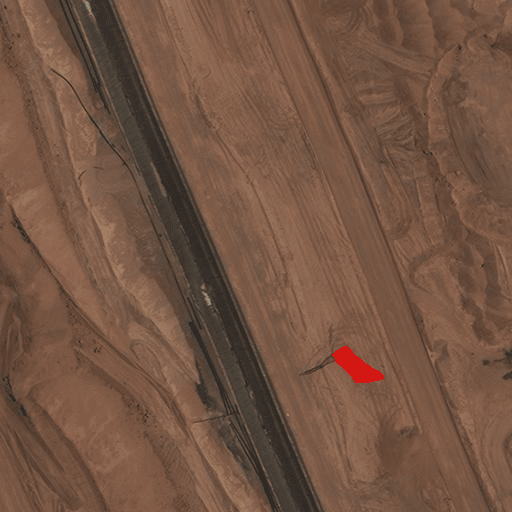} \\

\centering\arraybackslash\parbox{0.18\textwidth}
{\centering\vspace{-9pt}Identify the building with a large, rectangular structure and a distinct reddish-brown roof, surrounded by greenery.
} &
\includegraphics[width=1.05\linewidth]{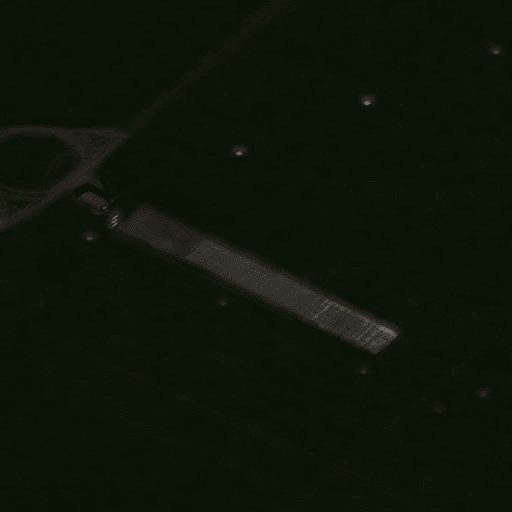} & 
\includegraphics[width=1.05\linewidth]{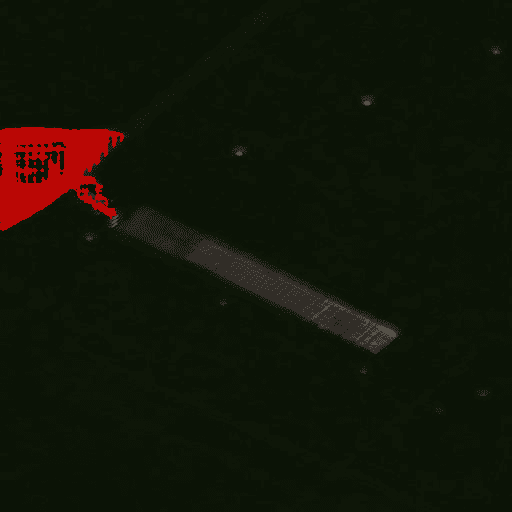} & 
\includegraphics[width=1.05\linewidth]{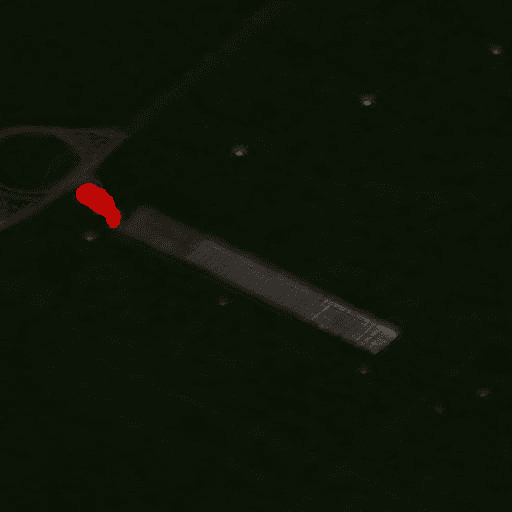} & 
\includegraphics[width=1.05\linewidth]{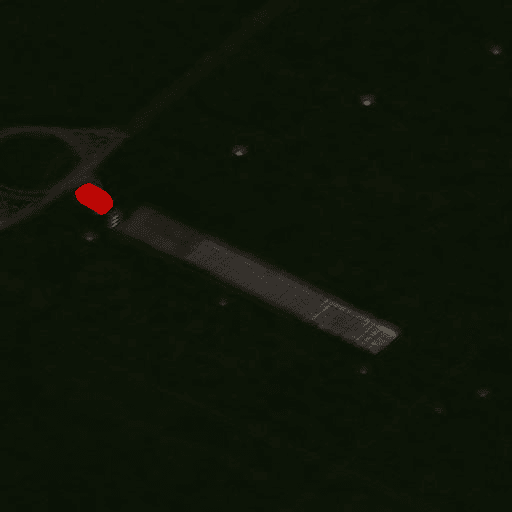} \\

\end{tabular}%
}
\label{gt_mistake_prediction_comparison}
\end{table}

\end{document}